\pdfoutput=1
\documentclass[preprint,authoryear]{elsarticle}
\usepackage{bm, amssymb, amsmath, ascmac, amsthm, mathtools, subcaption, url, xtab}
\usepackage[hidelinks, colorlinks=true, linkcolor=blue, urlcolor=blue]{hyperref}
\usepackage{tikz}
\usetikzlibrary{arrows.meta}
\mathtoolsset{showonlyrefs=true}
\allowdisplaybreaks[4]

\newtheorem{definition}{Definition}
\newtheorem{theorem}[definition]{Theorem}
\newtheorem{proposition}[definition]{Proposition}
\newtheorem{lemma}[definition]{Lemma}
\newtheorem{corollary}[definition]{Corollary}

\newtheorem{example}[definition]{Example}
\newtheorem{remark}[definition]{Remark}
\newtheorem{assumption}[definition]{Assumption}

\newcommand{\norm}[1]{\left\lVert#1\right\rVert}
\newcommand{\abs}[1]{\left|#1\right|}

\newcommand{\ceil}[1]{\left\lceil#1\right\rceil}

\newcommand{\VC}{\mathop{\mathrm{VC}}}
\newcommand{\Pdim}{\mathop{\mathrm{Pdim}}}

\begin{document}
\title{Universality of reservoir systems with recurrent neural networks}

\author[1]{Hiroki Yasumoto\corref{cor1}}
\ead{yasumoto@sys.i.kyoto-u.ac.jp}

\author[1]{Toshiyuki Tanaka}
\ead{tt@i.kyoto-u.ac.jp}

\cortext[cor1]{Corresponding author}

\affiliation[1]{
 organization={Graduate School of Informatics, Kyoto University}, 
 addressline={36-1},
 city={Yoshida Honmachi, Sakyo-ku, Kyoto}, 
 postcode={606-8501}, 
 country={Japan}}
\date{March 1, 2025}

\begin{abstract}
Approximation capability of reservoir systems whose reservoir is a recurrent neural network (RNN) is discussed.
We show what we call uniform strong universality of RNN reservoir systems for a certain class of dynamical systems.
This means that, given an approximation error to be achieved, one can construct an RNN reservoir system that approximates each target dynamical system in the class just via adjusting its linear readout.
To show the universality, we construct an RNN reservoir system via parallel concatenation that has an upper bound of approximation error independent of each target in the class.
\end{abstract}


\maketitle
\renewcommand*{\appendixname}{} 
\tableofcontents

\section{Introduction}
Reservoir computing \citep{Jaeger:2001:esn,Maass_et_al:2002:lsm} is a framework for processing sequential data \citep{Tanaka_et_al:2019:rc_review}.
Data are processed by a model that is a composition of a fixed reservoir, which maps data to a high-dimensional space, and a trainable readout \citep{Tanaka_et_al:2019:rc_review}.

In this paper, approximating ability of a family of reservoir systems whose reservoirs are recurrent neural networks (RNNs) is studied. 
We show what we call uniform strong universality (see below) of the family of RNN reservoir systems.
In general, given a family of functions, universality of the family 
implies that any function in another rich family of functions can be approximated with any precision by choosing a function out of the former family~\citep{Grigoryeva_and_Ortega:2018:esn_univ}.
Universality of families of reservoir systems have been shown in existing researches under various settings. 
Their settings can be classified into the following three categories. 
\begin{itemize}
 \item One is allowed to choose a reservoir according to each function to be approximated \citep{Grigoryeva_and_Ortega:2018:esn_univ,Grigoryeva_and_Ortega:2018:sas_univ,Gonon_and_Ortega:2020:rc_universality,Gonon_and_Ortega:2021:esn_univ_esp}. 
 In this paper, we call this type of results weak universality.
 \item Reservoir is randomly generated. Then, probabilistic approximation error bounds or expectation of approximation errors are discussed \citep{Gonon_et_al:2021:rc_app_ub,Li_and_Yang:2025:univ_esn_random_weights,Gonon_et_al:2024:inf_res_comp}.
 \item One is allowed to choose a reservoir, but it is fixed for a set of functions to be approximated. Then, only the readout is chosen according to each function to be approximated \citep{Cuchiero_et_al:2021:strong_univ}.
   In this paper, we call this type of results strong universality. 
\end{itemize}
This paper studies the approximation capability of the reservoir systems in the strong setting. 

Strong universality of a reservoir system means 
that one can obtain an upper bound of error in approximating any function
just by adjusting the linear readout of the reservoir system. 
Furthermore, such an upper bound of approximation errors can be made arbitrarily small as the scale of the reservoir becomes large, while allowing the upper bound to depend on functions to be approximated.
\cite{Cuchiero_et_al:2021:strong_univ} showed strong universality of a reservoir system whose reservoir implements a basis of a truncated Volterra series.
In their paper, the class of functions to be approximated is a certain class of filters, which are functions that map infinite sequences to infinite sequences.
By adjusting the linear readout, such a reservoir system can construct a truncated Volterra series of any filter in the class, and therefore an upper bound of approximation errors is obtained.

As mentioned above, we discuss what we call uniform strong universality of the family of RNN reservoir systems for a class of target functions, which is a special version of the strong universality.
This property means that, for any positive number $\varepsilon\in(0,\infty)$ and for the target class, we can construct a sufficiently large reservoir system out of the family, such that the approximation error for each target in the class is uniformly bounded from above by $\varepsilon$.
The term ``uniform'' indicates that the error bound $\varepsilon$ is dependent on the target class but is independent of each target function in the class.
As a target class, we assume a class of contracting dynamical systems whose maps are characterized in terms of the Barron class \citep{Barron:1993:nn_univ}.

It should be noted that the above notions of universality are
on representation ability of reservoir systems,
and not on learnability (e.g., how one can
find a reservoir system that approximates a specified function)
nor on generalization ability (e.g., how well the trained reservoir system
behaves on inputs that are not in the training set).
We do not discuss learnability in this paper\footnote{%
When one uses a linear readout, its training should be straightforward.},
while briefly touch on generalization in Section~\ref{ssec:complexity}
in terms of complexity measures. 

This paper is organized as follows.
Section~\ref{sec:preliminaries} introduces notations and fundamental notions that are to be used in the remainder of this paper.
Sections~\ref{sec:approx_ub_fnn}--\ref{sec:usu_lf} show the uniform strong universality step by step.
An approach that we take is parallel concatenation of RNN reservoirs on the basis of covering.
While such construction leads to impractically large RNN reservoirs, it helps to upgrade the weak approximation result to the uniform strong approximation result in an intuitive way.
It should be noted that the idea of parallel concatenation has also been employed with the aim of improving practical efficiency or performance of reservoir systems in empirical studies \citep{Schwenker_and_Labib:2009:esn_ensemble,Grigoryeva_el_al:2014:pr_res_univ}.
For taking this approach, Section~\ref{sec:approx_ub_fnn} introduces the existing approximation bounds of FNNs \citep{Caragea_et_al:2023:barron_unif_improved,Sreekumar_et_al:2021:barron_unif_sig1,Sreekumar_and_Goldfeld:2022:barron_unif_sig2} with properties suitable for taking covering, such as explicit dependence on the number of hidden nodes of an FNN and boundedness of parameters of an FNN.
On the basis of these approximation bounds, we show weak universality of the family of RNN reservoir systems for a case of finite-length inputs in Section~\ref{sec:weak_univ}.
Section~\ref{sec:unif_strong_univ} upgrades the weak universality of RNN reservoir systems to the strong one via parallel concatenation of RNN reservoirs and on the basis of covering.
In Section~\ref{sec:usu_lf}, the uniform strong universality of RNN reservoir systems is extended to a case of left-infinite inputs via a technique that we call cascading.
Section~\ref{sec:discussion} discusses the results and methods in this paper.
Section~\ref{sec:conclusion} concludes this paper.
Proofs are in Appendices.

The contributions of this paper are summarized as follows.
\begin{itemize} 
 \item We show uniform strong universality of RNN reservoir systems for contracting dynamical systems characterized in terms of the Barron class (Sections~\ref{sec:unif_strong_univ} and \ref{sec:usu_lf}). 
 \item We show that pseudo-dimension (see, e.g., \cite{Mohri_et_al:2018:fml}), a complexity measure commonly used in statistical learning theory, of the concatenated RNN reservoir systems is much smaller than the dimension of its reservoir state (Section~\ref{ssec:complexity}).
 \item We show that construction of RNN reservoir systems based on FNN approximation in uniform norm cannot guarantee uniqueness of states of the RNN reservoir systems given a left-infinite input sequence (Section~\ref{ssec:lim_ex_approach}).
\end{itemize}

\section{Preliminaries}\label{sec:preliminaries}
\subsection{Overview}
Before discussing universality from Section~\ref{sec:approx_ub_fnn}, we define necessary concepts in this section.
Section~\ref{ssec:notations} defines concepts used throughout this paper.
In order to define universality, we need to define target functions to be approximated, reservoir systems used to approximate targets, and approximation error between them.
Sections~\ref{ssec:dyn_sys}, \ref{ssec:approx_res_sys} and~\ref{ssec:approx_err_dyn_sys_by_res_sys} define these concepts, respectively.

\subsection{Notations}\label{ssec:notations}
For $D_1,D_2\in\mathbb{N}_+$, let $O_{D_1\times D_2}\in\mathbb{R}^{D_1\times D_2}$ be the zero matrix.
For $D\in\mathbb{N}_+$, let $I_{D\times D}\in\mathbb{R}^{D\times D}$ be the identity matrix.

For the $D$-dimensional Euclidean space, $\|\cdot\|_p$ denotes its $p$-norm, which is defined as $\norm{\bm{z}}_p:=(\sum_{d=1}^D\abs{z_d}^p)^{1/p}$ when $p\in[1,\infty)$ and $\norm{\bm{z}}_p:=\max_{d\in\{1,\ldots,D\}}\abs{z_d}$ when $p=\infty$.
Given $p\in[1,\infty]$, let $q\in[1,\infty]$ be
the H\"{o}lder conjugate of $p$, that is, $q$ satisfies $1/q+1/p=1$.

To represent a vector-valued function, we use a boldface italic symbol.
Otherwise, we use an italic symbol.

For a metric space $(\mathcal{A},d_{\mathcal{A}})$
and a normed vector space $(\mathcal{V},\|\cdot\|_{\mathcal{V}})$,
let $\mathcal{C}(\mathcal{A},\mathcal{V})$ denote
the collection of all continuous mappings from $\mathcal{A}$ to $\mathcal{V}$.
Note that $\mathcal{C}(\mathcal{A},\mathcal{V})$ itself is a normed vector
space with the supremum norm
$\norm{f}_{\mathcal{C}(\mathcal{A},\mathcal{V})}:=\sup_{x\in\mathcal{A}}\|f(x)\|_{\mathcal{V}}$. 
Especially, when $(\mathcal{V},\|\cdot\|_{\mathcal{V}})=(\mathbb{R}^y,\|\cdot\|_p)$, 
we use the symbol $\norm{\bm{f}}_{\mathcal{A},p}:=\norm{\bm{f}}_{\mathcal{C}(\mathcal{A},\mathbb{R}^y)}=\sup_{\bm{x}\in\mathcal{A}}\norm{\bm{f}(\bm{x})}_p$ to represent the norm of continuous functions.

For a function $f:\mathbb{R}^d\to\mathbb{R}$, let $\tilde{f}:\mathbb{R}^d\to\mathbb{C}$, $\tilde{f}(\bm{\omega}):=(2\pi)^{-d}\int_{\mathbb{R}^d}e^{i\bm{\omega}^\top\bm{x}}f(\bm{x})d\bm{x}$ be the Fourier transform of $f$.

For a normed vector space $(\mathcal{V},\|\cdot\|_{\mathcal{V}})$,
let $\bar{\mathcal{B}}_{\mathrm{unit},\|\cdot\|_{\mathcal{V}}}
:=\{\bm{x}\mid\norm{\bm{x}}_{\mathcal{V}}\le1,\bm{x}\in\mathcal{V}\}$
be the unit-norm ball in $\mathcal{V}$.
For $\mathcal{S}\subset\mathcal{V}$ and $a>0$,
let $a\mathcal{S}:=\{a\bm{x}\mid\bm{x}\in\mathcal{S}\}$.

Given a proposition $P$, $1_P$ denotes an indicator function that takes the value $1$ when $P$ is true and $0$ when $P$ is false.

Appendix~\ref{app:glossary} provides a glossary of the mathematical symbols that are introduced here and in the following parts of the paper.

\subsection{Target dynamical systems}\label{ssec:dyn_sys}
\subsubsection{Dynamical systems}\label{sssec:dyn_sys}
Here we provide a definition of generic
discrete-time dynamical systems in terms of (possibly nonlinear)
state space models as follows:
\begin{align}
 \bm{x}(t)&=\bm{g}(\bm{x}(t-1),\bm{u}(t)),\label{eq:ds}\\
 \bm{y}(t)&=\bm{h}(\bm{x}(t)),\label{eq:ds_ro}
\end{align}
where $\bm{x}(t)\in\mathbb{R}^D$ is a state in a $D$-dimensional
state space $\mathbb{R}^D$ at time step $t$, where $\bm{u}(t)\in\mathbb{R}^E$ is an $E$-dimensional input at time step $t$, where $\bm{y}(t)\in\mathbb{R}^{D_\mathrm{out}}$ is an output in a $D_\mathrm{out}$-dimensional space $\mathbb{R}^{D_\mathrm{out}}$ at time step $t$, where $\bm{g}:\mathbb{R}^D\times\mathbb{R}^E\to\mathbb{R}^D$ is a map that we call a state map, and where $\bm{h}:\mathbb{R}^D\to\mathbb{R}^{D_\mathrm{out}}$ is an output map.
By abuse of terminology, we may also identify
a pair of the maps $(\bm{g},\bm{h})$ as
the dynamical system that is defined by the maps $\bm{g}$ and $\bm{h}$ via~\eqref{eq:ds} and \eqref{eq:ds_ro},
upon which we may simply call $(\bm{g},\bm{h})$ a dynamical system. 

When one considers finite-length inputs,
one has to specify an initial state $\bm{x}_{\mathrm{init}}\in\mathbb{R}^D$
in order for the dynamical systems to be well defined.
In this paper, we assume without loss of generality
that all the dynamical systems
share the same initial state $\bm{x}_{\mathrm{init}}$. 
This is because, given a dynamical system with a certain initial state $\hat{\bm{x}}_\mathrm{init}\in\mathbb{R}^D$,
one may find a dynamical system with the initial state $\bm{x}_{\mathrm{init}}$
that behaves identically in terms of the input-output relation
via a system isomorphism \citep{Grigoryeva_and_Ortega:2018:esn_univ}:
Suppose that a dynamical system $(\hat{\bm{g}},\hat{\bm{x}}_\mathrm{init},\hat{\bm{h}})$ is given, where $\hat{\bm{g}}:\mathbb{R}^D\times\mathbb{R}^E\to\mathbb{R}^D$ is a state map and where $\hat{\bm{h}}:\mathbb{R}^D\to\mathbb{R}^{D_\mathrm{out}}$ is an output map.
Let $\bm{\phi}:\mathbb{R}^D\to\mathbb{R}^D$ be a homeomorphism such that $\bm{\phi}(\hat{\bm{x}}_\mathrm{init})=\bm{x}_\mathrm{init}$.
Then, a dynamical system $(\bm{g},\bm{x}_\mathrm{init},\bm{h})$, where $\bm{g}(\bm{x},\bm{u}):=\bm{\phi}(\hat{\bm{g}}(\bm{\phi}^{-1}(\bm{x}),\bm{u}))$ and where $\bm{h}:=\hat{\bm{h}}\circ\bm{\phi}^{-1}$, has the input-output relation identical to that of $(\hat{\bm{g}},\hat{\bm{x}}_\mathrm{init},\hat{\bm{h}})$.

Furthermore, in the reminder of this paper,  
without loss of generality we restrict our attention to 
dynamical systems with identity output maps, 
that is, $\bm{y}(t)=\mathrm{id}(\bm{x}(t))=\bm{x}(t)$,
where $\mathrm{id}$ denotes the identity map.
Indeed, given a dynamical system $(\bm{g},\bm{h})$,
one can define another dynamical system $(\bar{\bm{g}},\bar{\bm{h}})$ as
\begin{align}
  \bar{\bm{x}}(t)&=\bar{\bm{g}}(\bar{\bm{x}}(t-1),\bm{u}(t)),
  \nonumber\\
  \bm{y}(t)&=\bar{\bm{h}}(\bar{\bm{x}}(t)):=[O_{D_\mathrm{out}\times D},I_{D_\mathrm{out}\times D_\mathrm{out}}]\bar{\bm{x}}(t),
\end{align}
where
\begin{equation}
  \bar{\bm{x}}(t)=\left(\begin{array}{c}
    \bm{x}(t)\\
    \bm{y}(t)
  \end{array}\right),\quad
  \bar{\bm{g}}(\bar{\bm{x}}(t-1),\bm{u}(t))
  :=\left(\begin{array}{c}
    \bm{g}(\bm{x}(t-1),\bm{u}(t))\\
    \bm{h}\bigl(\bm{g}(\bm{x}(t-1),\bm{u}(t))\bigr)
  \end{array}\right).\label{eq:dynsys_redefined}
\end{equation}
Since the dynamical systems $(\bm{g},\bm{h})$ and $(\bar{\bm{g}},\bar{\bm{h}})$ are equivalent in terms of their input-output relations,
approximation of $(\bm{g},\bm{h})$ is the same as that of $(\bar{\bm{g}},\bar{\bm{h}})$, the latter of which will immediately be available
once approximation of $(\bar{\bm{g}},\mathrm{id})$ is obtained. 
For a dynamical system $(\bm{g},\mathrm{id})$ with the identity output map $\mathrm{id}$, by abuse of terminology, we call $\bm{g}$ itself a dynamical system.

\subsubsection{Assumptions for target dynamical systems}\label{sssec:cls_dyn_sys}
In this paper, we consider approximating dynamical systems in a certain class
by a reservoir system.
We call the dynamical systems to be approximated target dynamical systems. 
In order to make the following argument concrete,
first of all we have to specify a class $\mathcal{G}$
of target dynamical systems to be considered in this paper.

We mainly consider finite-horizon dynamical systems
with bounded input and state sequences in Sections~\ref{sec:weak_univ} and \ref{sec:unif_strong_univ}, while Section~\ref{sec:usu_lf}
discusses the infinite-horizon case. 
Let $\mathcal{T}_T:=\{-T+1,\ldots,1,0\}$ be a set of time indices of length $T$.
We consider finite-length bounded input sequences $\{\bm{u}(t)\}_{t\in\mathcal{T}_T}\in(\bar{\mathcal{B}}_I)^T$ with length $T$, 
where $\bar{\mathcal{B}}_I:=I\bar{\mathcal{B}}_{\mathrm{unit},\|\cdot\|_p}=\{\bm{u}\mid\norm{\bm{u}}_p\le I,\bm{u}\in\mathbb{R}^E\}$ with $I\in(0,\infty)$ denotes a space of inputs at every time step. 
Let $\bar{\mathcal{B}}_S:=S\bar{\mathcal{B}}_{\mathrm{unit},\|\cdot\|_p}=\{\bm{x}\mid\norm{\bm{x}}_p\le S,\bm{x}\in\mathbb{R}^D\}$ with $S\in(0,\infty)$.
We assume that, given $p\in[1,\infty]$ and $I,S\in(0,\infty)$,
for any function $\bm{g}$ in $\mathcal{G}$
the domain of $\bm{g}$ contains 
$\bar{\mathcal{B}}_{S,I}:=\bar{\mathcal{B}}_S\times\bar{\mathcal{B}}_I$.
We furthermore assume, in view of the argument in Section~\ref{sssec:dyn_sys},
that all $\bm{g}\in\mathcal{G}$ have a common initial state $\bm{x}_\mathrm{init}\in\bar{\mathcal{B}}_S$.
We assume that 
the sequence $\{\bm{x}(t)\}_{t\in\mathcal{T}_T}$ of states generated
by any dynamical system $\bm{g}\in\mathcal{G}$ with any input sequence
in $(\bar{\mathcal{B}}_I)^T$ resides in 
the set $(\bar{\mathcal{B}}_S)^T$.

In order to state the assumptions on $\mathcal{G}$, a relevant notion is introduced.
Our approximation is based on approximating $\bm{g}$ with FNNs.
We use the following class of functions, which is one of classes of real-valued functions that can be approximated with FNNs \citep{Barron:1993:nn_univ,Caragea_et_al:2023:barron_unif_improved}.
We use the following simplified version of the definition of \cite{Caragea_et_al:2023:barron_unif_improved}.
\begin{definition}[Barron class \citep{Caragea_et_al:2023:barron_unif_improved}]\label{def:M_Barron}
 Let $\mathcal{B}$ be a bounded subset of $\mathbb{R}^d$ that includes the origin $\bm{0}$.
 Let $M\in(0,\infty)$.
 Let $\norm{\bm{\omega}}_\mathcal{B}:=\sup_{\bm{x}\in\mathcal{B}}\abs{\bm{\omega}^\top\bm{x}}$. 
 Let $\mathcal{Z}_{M,\mathbb{R}^d}$ be the set of functions $g:\mathbb{R}^d\to\mathbb{R}$ such that $g(\bm{x})=g(\bm{0})+\int(e^{i\bm{\omega}^\top\bm{x}}-1)\tilde{g}(\bm{\omega})d\bm{\omega}$ on $\mathcal{B}$ and such that $\abs{g(\bm{0})}\le M$.
Let $\mathcal{Z}_{M,\mathcal{B}}$ be the set of functions on $\mathcal{B}$ each of which has an extension $g\in\mathcal{Z}_{M,\mathbb{R}^d}$ satisfying $\int\norm{\bm{\omega}}_\mathcal{B}\abs{\tilde{g}(\bm{\omega})}d\bm{\omega}\le M$.
 We call $\mathcal{Z}_{M,\mathcal{B}}$ the Barron class with constant $M$.
\end{definition}

Now the assumptions on $\mathcal{G}$ are stated.
\begin{assumption}[assumptions on target dynamical systems]\label{asm:ds_asm_fin}
Fix $p\in[1,\infty]$.
Let $M,S,I,P$, and $L$ be positive constants that are dependent on the family $\mathcal{G}$ and independent of each element $\bm{g}\in\mathcal{G}$.
We assume the following properties on all of the target dynamical systems $\bm{g}=(g_1,\ldots,g_D)\in\mathcal{G}$.
\begin{enumerate}\leftskip=4mm
  \renewcommand{\labelenumi}{A\ref*{asm:ds_asm_fin}-\theenumi.}
 \item $g_d\in\mathcal{Z}_{M,\bar{\mathcal{B}}_{S,I}}$, $d=1,\ldots,D$.\label{enm:can_approximate_w_nn}
 \item $\forall\bm{u}\in\bar{\mathcal{B}}_I$, $\bm{g}(\cdot,\bm{u}):(\bar{\mathcal{B}}_S,d_p)\to(\bar{\mathcal{B}}_S,d_p)$ is $L$-Lipschitz, where $d_p$ is the metric induced by $l^p$-norm of $\mathbb{R}^D$.\label{enm:lipschitz_wrt_input}
 \item $\norm{\bm{g}}_{\bar{\mathcal{B}}_{S,I},p}\le P<S$. \label{enm:uniformly_bounded}
\end{enumerate}
\end{assumption}

Some remarks on Assumption~\ref{asm:ds_asm_fin} are in order.
\begin{remark}
Assumption A\ref*{asm:ds_asm_fin}-\ref{enm:can_approximate_w_nn} is needed for approximating a target dynamical system by an FNN in a componentwise manner on the basis of results in \cite{Caragea_et_al:2023:barron_unif_improved}.
It should be noted that $M$ has to be universal in the sense that
Assumption A\ref*{asm:ds_asm_fin}-\ref{enm:can_approximate_w_nn}
should hold for any $\bm{g}\in\mathcal{G}$ with the same value of $M$. 
This property is important for the uniform strong universality to hold, which requires the upper bound of approximation error to be independent of each $\bm{g}\in\mathcal{G}$. 
Assumptions A\ref*{asm:ds_asm_fin}-\ref{enm:lipschitz_wrt_input} and A\ref*{asm:ds_asm_fin}-\ref{enm:uniformly_bounded} are needed for applying the framework of \cite{Grigoryeva_and_Ortega:2018:esn_univ} for proving universality.
The constant $P$ in Assumption A\ref*{asm:ds_asm_fin}-\ref{enm:uniformly_bounded} should be smaller than $S$.
$S-P>0$ represents a margin between the range of $\bm{g}$ and the domain of the state of $\bm{g}$.
This margin allows FNNs that approximate $\bm{g}$ to be again domain-preserving.
The domain-preserving property of the approximating FNNs enables us to evaluate norm of filter defined later (see Section~\ref{sssec:norm_filter} and Lemma~\ref{lem:filter_nn_diff_ub}).
\end{remark}

\begin{remark}\label{rm:ex_barron_class}
 As discussed in Section~IX of \cite{Barron:1993:nn_univ},
 a function whose partial derivatives are bounded up to a sufficiently high order is of the Barron class. 
 Thus, a sufficient condition 
 for a set of functions $g_1,\ldots,g_D:\bar{\mathcal{B}}_{S,I}\to\mathbb{R}$ to satisfy Assumption A\ref*{asm:ds_asm_fin}-\ref{enm:can_approximate_w_nn}
 is that these functions have extensions such that the partial derivatives up to a sufficiently high order are bounded by a universal constant on a domain slightly larger than $\bar{\mathcal{B}}_{S,I}$.
 Then, by following the argument of smooth extension discussed in 15th item in Section~IX of \cite{Barron:1993:nn_univ}, one can show that there exists a universal constant $M>0$ such that all of the functions in the set are of the Barron class with the constant $M$.
\end{remark}

\begin{remark}\label{rm:lim_dom_prsv}
 The domain-preserving property (A\ref*{asm:ds_asm_fin}-\ref{enm:uniformly_bounded}) may limit the variety of the dynamical systems in $\mathcal{G}$.
 For example, it excludes dynamical systems whose states grow over time without bound.
 We argue, however, that this assumption is inevitable
 in proving strong universality of RNN reservoir systems in this paper,
 because our argument is based on approximation bounds of FNNs, 
 where approximation of a function on a compact domain with an FNN
 is discussed.
\end{remark}

\begin{remark}\label{rm:comp_assumptions}
 The existing strong universality result on reservoir systems~\citep{Cuchiero_et_al:2021:strong_univ} assumes analyticity and boundedness of target filters, the latter of which is similar to the domain-preserving property.
 Compared with their assumptions, one can argue that our assumptions are weaker in that we do not assume analyticity, and stronger in that we explicitly assume state-space expressions of target dynamical systems whose state dimension is assumed known, which affects the scales of approximating reservoirs and the approximation error bounds.
\end{remark}

\subsection{Approximating reservoir systems}\label{ssec:approx_res_sys}
In this subsection, we define a family of RNN reservoir systems, whose weak and uniform strong universality against $\mathcal{G}$ is proved later in this paper.
We define reservoir systems, RNN reservoir systems and the family of RNN reservoir systems in order.

We consider the following discrete-time reservoir system:
\begin{align}
 \bm{s}(t)&=\bm{r}(\bm{s}(t-1),\bm{u}(t)),\label{eq:res}
 \\
 \bm{y}(t)&=W\bm{s}(t),\label{eq:readout}
\end{align}
where $\bm{s}(t)\in\mathbb{R}^{N_\mathrm{res}}$ and $\bm{u}(t)\in\mathbb{R}^E$ denote
a state and an input, respectively, at time step $t$, 
and where 
$\bm{r}:\mathbb{R}^{N_\mathrm{res}}\times\mathbb{R}^E\to\mathbb{R}^{N_\mathrm{res}}$ is a reservoir map.
We call the dynamical system \eqref{eq:res} defined by $\bm{r}$ a reservoir made with $\bm{r}$.
$\bm{y}(t)\in\mathbb{R}^{D_\mathrm{out}}$ is an output at time step $t$.
The matrix $W\in\mathbb{R}^{D_\mathrm{out}\times N_\mathrm{res}}$ defines an adjustable linear readout.
The initial state $\bm{s}_\mathrm{init}\in\mathbb{R}^{N_\mathrm{res}}$ is fixed.
We call the combination of the reservoir \eqref{eq:res} and the readout \eqref{eq:readout} a reservoir system.

We specifically consider use of the following FNN $\bm{f}_{\rho,\theta}:\mathbb{R}^{N_\mathrm{res}}\times\mathbb{R}^E\to\mathbb{R}^{N_\mathrm{res}}$ as a reservoir map $\bm{r}$ in~\eqref{eq:res}, where $\rho(x):=\max\{0,x\}$, $\rho:\mathbb{R}\to\mathbb{R}$ is the rectified linear unit (ReLU) activation function and where $\theta:=\{A,B,C,\bm{d},\bm{e}\}$ is a set of parameters.
\begin{align}
 \bm{s}(t)&=\bm{f}_{\rho,\theta}(\bm{s}(t-1),\bm{u}(t)):=A\bm{\rho}(B \bm{s}(t-1)+C\bm{u}(t)+\bm{d})+\bm{e},\label{eq:fnn}
\end{align}
where $A\in\mathbb{R}^{N_\mathrm{res}\times N}$, $B\in\mathbb{R}^{N\times N_\mathrm{res}}$, $C\in\mathbb{R}^{N\times E}$, $\bm{d}\in\mathbb{R}^N$ and $\bm{e}\in\mathbb{R}^{N_\mathrm{res}}$ are parameters 
and where $\bm{\rho}:\mathbb{R}^N\to\mathbb{R}^N$ is a map that applies the ReLU activation function $\rho$ elementwise to its vector argument.
When a reservoir and a reservoir system is made with an FNN reservoir map $\bm{f}_{\rho,\theta}$, we call it an RNN reservoir and an RNN reservoir system, respectively.

We consider the family of RNN reservoir systems with all sizes and parameters.
Formally, we consider RNN reservoir systems whose FNN reservoir maps are in a family $\mathcal{F}:=\{\bm{f}_{\rho,\theta}\mid \theta=(A,B,C,\bm{d},\bm{e}),A\in\mathbb{R}^{N_\mathrm{res}\times N},B\in\mathbb{R}^{N\times N_\mathrm{res}},C\in\mathbb{R}^{N\times E},\bm{d}\in\mathbb{R}^N,\bm{e}\in\mathbb{R}^{N_\mathrm{res}},N\in\mathbb{N}_+,N_\mathrm{res}\in\mathbb{N}_+\}$.

\subsection{Approximation error of a reservoir system for a dynamical system}\label{ssec:approx_err_dyn_sys_by_res_sys}
\subsubsection{Overview}\label{ssec:approx_err_overview}
In order to define universality, we have to define approximation error of a reservoir system for a target dynamical system.
In Section~\ref{sssec:filter}, dynamical systems and reservoir systems are regarded as functions that output a state sequence and an output sequence given an finite-length input sequence, respectively, which we call finite-length filter.
Section~\ref{sssec:norm_filter} defines a norm for such entities.
Section~\ref{sssec:approx_err} defines the approximation error of the two systems in terms of the norm.

In this section, we only formulate the case of finite-length inputs, which will be extended to the case of left-infinite inputs in Section~\ref{ssec:def_usu_lf}.
Although finite-length filters are just functions of finite-dimensional vectors, we formulate their norm and approximation error in a way that is consistent with the left-infinite analogues \citep{Grigoryeva_and_Ortega:2018:esn_univ}.
This enables straightforward adaptation of the technique developed in \cite{Grigoryeva_and_Ortega:2018:esn_univ} as in the proof of Lemma~\ref{lem:filter_nn_diff_ub}.
Also, this makes our migration to the left-infinite case easy.

\subsubsection{Finite-length filters}\label{sssec:filter}
We call an operator $V$ that maps a sequence indexed with $\mathcal{T}_T$ to another sequence indexed with $\mathcal{T}_T$ a finite-length filter.
This name is based on the term ``filter'', which is defined as a map from infinite-length inputs to infinite-length outputs \citep{Grigoryeva_and_Ortega:2018:sas_univ}.

Let $\mathcal{A}$ be a set.
For any finite-length filter $V:\mathcal{A}^T\to(\mathbb{R}^D)^T$, let $V(\bm{a})_t\in\mathbb{R}^D$, $t\in\mathcal{T}_T$ denote the output of the finite-length filter at time step $t$ given an input sequence $\bm{a}\in\mathcal{A}^T$.

A dynamical system induces a finite-length filter. 
\begin{definition}\label{def:dyn_sys_filter}
A state map $\bm{g}$ in \eqref{eq:ds} and its initial state $\bm{x}_\mathrm{init}$ induce a map from the space $(\bar{\mathcal{B}}_I)^T$ of input sequences to the space $(\mathbb{R}^D)^T$ of state sequences.
We call this map a \emph{finite-length state filter} and write it as $V_{\bm{g},\bm{x}_\mathrm{init}}$.
Furthermore, by composing an output map $\bm{h}$ \eqref{eq:ds_ro} with $V_{\bm{g},\bm{x}_\mathrm{init}}$, one can define a map $V_{\bm{g},\bm{x}_\mathrm{init}}^{\bm{h}}:(\bar{\mathcal{B}}_I)^T\to(\mathbb{R}^{D_\mathrm{out}})^T$, $V_{\bm{g},\bm{x}_\mathrm{init}}^{\bm{h}}(\bm{u})_t=\bm{h}(V_{\bm{g},\bm{x}_\mathrm{init}}(\bm{u})_t)$, $\forall\bm{u}\in(\bar{\mathcal{B}}_I)^T$, $\forall t\in\mathcal{T}_T$.
We call this map a \emph{finite-length dynamical system filter}.
\end{definition}

Since a reservoir system is itself a dynamical system, we can define its filters in the same manner.
\begin{definition}\label{def:res_sys_filter}
We call a state filter $V_{\bm{r},\bm{s}_\mathrm{init}}$ and a dynamical system filter $V_{\bm{r},\bm{s}_\mathrm{init}}^W$ of a reservoir system $(\bm{r},\bm{s}_\mathrm{init},W)$ a \emph{finite-length reservoir state filter} and a \emph{finite-length reservoir system filter}, respectively.
\end{definition}

The notations of filters are based on those in \cite{Grigoryeva_and_Ortega:2018:esn_univ}.

\subsubsection{Norm of finite-length filters}\label{sssec:norm_filter}
We use a finite-length-input version of the norm of filter that is used in \cite{Grigoryeva_and_Ortega:2018:esn_univ} for left-infinite inputs.
Let $(\mathbb{R}^a,\|\cdot\|_p)$ be the Euclidean space equipped
with the $p$-norm. 
For the class
$\mathcal{C}((\bar{\mathcal{B}}_I)^T,(\mathbb{R}^a)^T)$ of continuous filters,
the following supremum norm is naturally defined. 
\begin{definition}[norm of finite-length filters]
\label{def:filter_norm}
We define the supremum norm on
$\mathcal{C}((\bar{\mathcal{B}}_I)^T,(\mathbb{R}^a)^T)$ as
$\norm{V}_{\mathcal{T}_T,p}:=\sup_{\bm{u}\in(\bar{\mathcal{B}}_I)^T}\sup_{t\in\mathcal{T}_T}
\|V(\bm{u})_t\|_p$ for
$V\in\mathcal{C}((\bar{\mathcal{B}}_I)^T,(\mathbb{R}^a)^T)$.
\end{definition}

Because a length-$T$ sequence in $(\mathbb{R}^a)^T$
can alternatively be viewed as a mapping from $\mathcal{T}_T$ to $\mathbb{R}^a$,
a function from $(\bar{\mathcal{B}}_I)^T$
to $(\mathbb{R}^a)^T$ can alternatively be regarded as
that from $(\bar{\mathcal{B}}_I)^T\times\mathcal{T}_T$
to $\mathbb{R}^a$.
The above definition of $\|\cdot\|_{\mathcal{T}_T,p}$
can then be regarded as the supremum norm
in the class $\mathcal{C}((\bar{\mathcal{B}}_I)^T\times\mathcal{T}_T,\mathbb{R}^a)$ of continuous functions
from the metric space $(\bar{\mathcal{B}}_I)^T\times\mathcal{T}_T$
with the sup metric to the normed space $(\mathbb{R}^a,\|\cdot\|_p)$.

\subsubsection{Approximation error}\label{sssec:approx_err}
We define the approximation error between a dynamical system $\bm{g}$ and a reservoir system $(\bm{r},\bm{s}_\mathrm{init},W)$ in terms of their filters as follows.
\begin{definition}[approximation error]\label{def:f_and_res_approx_err}
The approximation error between a dynamical system $\bm{g}:\bar{\mathcal{B}}_{S,I}\to\bar{\mathcal{B}}_S$ with an initial state $\bm{x}_\mathrm{init}\in\bar{\mathcal{B}}_S$ and a reservoir system $(\bm{r},\bm{s}_\mathrm{init},W)\in\{\mathbb{R}^{N_\mathrm{res}}\times\mathbb{R}^E\to\mathbb{R}^{N_\mathrm{res}}\}\times\mathbb{R}^{N_\mathrm{res}}\times\mathbb{R}^{D\times N_\mathrm{res}}$ is defined as 
\begin{align}
 \mathrm{err}(\bm{g},(\bm{r},\bm{s}_\mathrm{init},W)):=\norm{V_{\bm{g},\bm{x}_\mathrm{init}}-V_{\bm{r},\bm{s}_\mathrm{init}}^W}_{\mathcal{T}_T,p}.
\end{align}
\end{definition}

\section{Approximation bounds of feedforward neural networks}\label{sec:approx_ub_fnn}
\subsection{A case of ReLU activation function}\label{ssec:nn_approx_relu}
We derive uniform strong universality results for a family of RNN reservoir systems with ReLU activation function on the basis of the following existing FNN approximation bound in \cite{Caragea_et_al:2023:barron_unif_improved} with minor revisions.
See Appendix~\ref{aapp:proof_prop_2_2} for revised points and derivation,
and Appendix~\ref{aapp:remark_prop_2_2} for technical remarks.
\begin{proposition}[corrected and modified version of Proposition~2.2 in \cite{Caragea_et_al:2023:barron_unif_improved}]\label{pp:prop_2_2}
 There exists a universal constant $\kappa>0$ that has the following properties.
 Let $\mathcal{B}\subset\mathbb{R}^Q$ be a bounded set including the origin $\bm{0}$ and with non-empty interior.
 For any $M>0$, $g\in\mathcal{Z}_{M,\mathcal{B}}$, $N\in\mathbb{N}_+$, there exists an FNN with $4N$ hidden nodes
 \begin{align}
  f_\rho(\bm{x}):=\sum_{i=1}^{4N}a_i\rho(\bm{b}_i^\top\bm{x}+c_i)+e,
 \end{align}
 such that
 \begin{align}
  \norm{g-f_\rho}_{\mathcal{B},\infty}\le\kappa\sqrt{Q}MN^{-1/2},\label{eq:ub_desired}
 \end{align}
 and such that
 \begin{align}
  \abs{a_i}\le2\sqrt{M/N},\ \norm{\bm{b}_i}_\mathcal{B}\le\sqrt{M},\ \abs{c_i}\le2\sqrt{M},\ \abs{e}\le M,\quad i=1,\ldots,4N.\label{eq:coeff_ub}
 \end{align}
\end{proposition}

It should be noted that Barron's result~\citep{Barron:1993:nn_univ}
is not applicable here, as we will argue approximation errors
in terms of uniform norm, whereas the result in~\cite{Barron:1993:nn_univ} is on approximation errors
in terms of the $L^2$ norm.

\subsection{A case of monotone sigmoid activation function}\label{ssec:nn_approx_sig}
Although the approximation bound in Section~\ref{ssec:nn_approx_relu} will lead to uniform strong universality of RNN reservoir systems with the ReLU, 
one may argue that a smooth sigmoid function would be
more appropriate when one is interested in approximating dynamical systems.
Therefore, we introduce an alternative approximation bound for FNNs, which assumes a monotone sigmoid activation function.
One can repeat almost the same discussion on the basis of this bound for proving uniform strong universality of RNN reservoir systems with a monotonically increasing sigmoid function that is Lipschitz continuous.
We discuss this case in detail in Section~\ref{ssec:sig_func}.

Although one may regard that our result introduced in this section is qualitatively covered by existing work \citep{Barron:1992:barron_unif,Sreekumar_et_al:2021:barron_unif_sig1,Sreekumar_and_Goldfeld:2022:barron_unif_sig2}, ours gives an alternative proof and slight improvement over the existing ones, making it of independent interest. 
See Appendix~\ref{aapp:proof_prop_2_2_sig} for derivation, and
Appendix~\ref{aapp:remark_prop_2_2_sig} for relations to the existing work and a technical remark.
In this paper, we define a sigmoid function as a function $\sigma:\mathbb{R}\to[0,1]$ such that $\lim_{x\to\infty}\sigma(x)=1$ and $\lim_{x\to-\infty}\sigma(x)=0$.
\begin{proposition}[a complementary result of \cite{Barron:1992:barron_unif,Sreekumar_et_al:2021:barron_unif_sig1,Sreekumar_and_Goldfeld:2022:barron_unif_sig2}]\label{pp:prop_2_2_sig}
 Let $\sigma$ be a monotonically increasing continuous sigmoid function.
 Let $\delta:[0,\infty)\to[0,\infty)$, $\delta(\Lambda):=\int_{-1}^1\abs{1_{x>0}-\sigma(\Lambda x)}dx$ be an approximation error of a step function by a scaled sigmoid function on the interval $[-1,1]$.
 There exists a universal constant $\kappa'>0$ that has the following properties.
 Let $\mathcal{B}\subset\mathbb{R}^Q$ be a bounded set including the origin $\bm{0}$ and with non-empty interior.
 For any $M>0$, $g\in\mathcal{Z}_{M,\mathcal{B}}$, $N\in\mathbb{N}_+$ and $\Lambda>0$, there exists an FNN with $2N$ hidden nodes
 \begin{align}
  f_\sigma(\bm{x}):=\sum_{i=1}^{2N}a_i\sigma(\bm{b}_i^\top\bm{x}+c_i)+e,
 \end{align}
 such that
 \begin{align}
  \norm{g-f_\sigma}_{\mathcal{B},\infty}\le M(4\delta(\Lambda)+\kappa'\sqrt{Q}N^{-1/2}),\label{eq:ub_desired_sig}
 \end{align}
 and such that
 \begin{align}
  \abs{a_i}\le2M/N,\ \norm{\bm{b}_i}_\mathcal{B}\le\Lambda,\ \abs{c_i}\le\Lambda,\ \abs{e}\le M,\quad i=1,\ldots,2N.\label{eq:coeff_ub_sig}
 \end{align}
\end{proposition}

We are inspired by Theorem~3 in \cite{Barron:1993:nn_univ} to use the approximation error $\delta(\Lambda)$ of a step function and a scaled sigmoid function in Proposition~\ref{pp:prop_2_2_sig}.
It should be noted that $\delta(\Lambda)\to0$ as $\Lambda\to\infty$.

\section{Weak universality}\label{sec:weak_univ}
\subsection{Overview}
In this section, we prove weak universality of the family of RNN reservoir systems for a case of finite-length inputs on the basis of the approximation bound of FNNs discussed in Section~\ref{ssec:nn_approx_relu}.
The purpose is to introduce ideas that will be used to show the uniform strong universality for a case of finite-length inputs in Section~\ref{sec:unif_strong_univ}.

It should be noted that we do not claim the result of weak universality to be presented in this section as our novel contribution because similar weak universality results have already been shown for RNN reservoir systems \citep{Grigoryeva_and_Ortega:2018:esn_univ,Gonon_and_Ortega:2020:rc_universality,Gonon_and_Ortega:2021:esn_univ_esp} for left-infinite inputs and for general target functions.
In a broader context beyond reservoir computing, weak universality results can also be found in the literature of studying approximation of dynamical systems with RNNs (see, e.g., \cite{Matthews:1993:non_fading_mem,Funahashi_and_Nakamura:1993:approx_dynsys_rnn,Chow_and_Li:2000:approx_dynsys_rnn_input,Hanson_and_Raginsky:2020:sim_dynsys_rnn}).

We do not discuss the weak universality for the case of left-infinite inputs in this paper, 
because the uniform strong universality for the left-infinite case will be proved in Section~\ref{sec:usu_lf} on the basis of the uniform strong universality result for the finite-length case in Section~\ref{sec:unif_strong_univ}. 

Section~\ref{ssec:def_wu} defines the weak universality.
Section~\ref{ssec:result_wu} states the main result.
Section~\ref{ssec:outline_proof_wu} provides outline of the proof of the main result.

\subsection{Definition}\label{ssec:def_wu}
We define weak universality of a family of reservoir systems for a class $\mathcal{G}$ of target dynamical systems.
It is a property such that, for any target dynamical system $\bm{g}$ in the family $\mathcal{G}$, one can choose a reservoir system with an arbitrarily small approximation error.
Note that in this case one is allowed to choose a reservoir map depending on the individual target dynamical system $\bm{g}$.
The formal definition is given as follows.
\begin{definition}[weak universality]\label{def:weak_univ}
 Let $\mathcal{R}\subset\{\mathbb{R}^{N_\mathrm{res}}\times\mathbb{R}^E\to\mathbb{R}^{N_\mathrm{res}},N_\mathrm{res}\in\mathbb{N}_+\}$ be a family of reservoir maps with finite state dimensions.
 We say that the family of reservoir systems made with $\mathcal{R}$ has weak universality for $\mathcal{G}$ if and only if, for all $\varepsilon>0$ and for all $\bm{g}\in\mathcal{G}$, there exist $N_\mathrm{res}\in\mathbb{N}_+$, an $N_\mathrm{res}$-dimensional reservoir map $\bm{r}\in\mathcal{R}$, its initial state $\bm{s}_\mathrm{init}\in\mathbb{R}^{N_\mathrm{res}}$ and a readout $W\in\mathbb{R}^{D\times N_\mathrm{res}}$ such that $\mathrm{err}(\bm{g},(\bm{r},\bm{s}_\mathrm{init},W))\le\varepsilon$.
\end{definition}

\subsection{Result}\label{ssec:result_wu}
The weak universality of the family of RNN reservoir systems for the target dynamical systems $\mathcal{G}$ holds.
\begin{theorem}[weak universality]\label{th:rnn_res_sys_weak_univ}
 When inputs are finite-length, under Assumption~\ref{asm:ds_asm_fin}, the family of RNN reservoir systems has weak universality for $\mathcal{G}$.
\end{theorem}

\subsection{Outline of proof}\label{ssec:outline_proof_wu}
Theorem~\ref{th:rnn_res_sys_weak_univ} is proved via combining two results.
Firstly, we use Proposition~\ref{pp:prop_2_2} for approximating the map of a target dynamical system with a vector-valued FNN reservoir map.
Then, we upgrade this approximation bound to that in terms of filter norm via a technique in \cite{Grigoryeva_and_Ortega:2018:esn_univ}.

We approximate a target dynamical system $\bm{g}\in\mathcal{G}$ with a vector-valued FNN.
Since each component function of $\bm{g}$ is of the Barron class, $\bm{g}$ can be approximated by a vector-valued FNN made via bundling $D$ real-valued bounded-parameter FNNs on the basis of Proposition~\ref{pp:prop_2_2}.
For stating the approximation bound, we specify the set of vector-valued FNNs from which FNNs for approximating $\bm{g}$ are chosen.
\begin{definition}[bounded-parameter FNNs]\label{def:cb_approx_fnn}\ 
\begin{enumerate}
 \item Let $\mathcal{F}_{N,M}^1:=\{f:\mathbb{R}^D\times\mathbb{R}^E\to\mathbb{R}\mid 
 f(\bm{s},\bm{u})=\sum_{n=1}^{4N}a_n\rho(\bm{b}_n^\top\bm{s}+\bm{c}_n^\top\bm{u}+d_n)+e,
  a_n\in\mathcal{Q}_{\mathrm{a},M},
  \bm{b}_n\in\mathcal{Q}_{\mathrm{b},M},
  \bm{c}_n\in\mathcal{Q}_{\mathrm{c},M},
  d_n\in\mathcal{Q}_{\mathrm{d},M},n=1,\ldots,4N,
  e\in\mathcal{Q}_{\mathrm{e},M}\}$,
  where $\mathcal{Q}_{\mathrm{a},M}:=[-2\sqrt{M},2\sqrt{M}]$, $\mathcal{Q}_{\mathrm{b},M}:=\{\bm{b}_i\mid\norm{\bm{b}_i}_{\bar{\mathcal{B}}_S}\le\sqrt{M}\}$, $\mathcal{Q}_{\mathrm{c},M}:=\{\bm{c}_i\mid\norm{\bm{c}_i}_{\bar{\mathcal{B}}_I}\le\sqrt{M}\}$, $\mathcal{Q}_{\mathrm{d},M}:=[-2\sqrt{M},2\sqrt{M}]$, $\mathcal{Q}_{\mathrm{e},M}:=[-M,M]$ be a family of FNNs for approximating component $g_d\in\mathcal{Z}_{M,\bar{\mathcal{B}}_{S,I}}$, $d=1,\ldots,D$ of $\bm{g}$ on the basis of Proposition~\ref{pp:prop_2_2}.
  We call $\mathcal{F}_{N,M}^1$ a family of real-valued bounded-parameter FNNs.
 \item Let $\mathcal{F}_{N,M}^D:=\{\bm{f}:\mathbb{R}^D\times\mathbb{R}^E\to\mathbb{R}^D\mid\bm{f}(\bm{s},\bm{u})=(f_1(\bm{s},\bm{u}),\ldots,f_D(\bm{s},\bm{u})),f_1,\ldots,f_D\in\mathcal{F}_{N,M}^1\}$.
We call $\mathcal{F}_{N,M}^D\subset\mathcal{F}$ a family of bounded-parameter FNNs.
\end{enumerate}
\end{definition}

Then, the following upper bound of approximation error holds trivially.
\begin{lemma}[corollary of Proposition~\ref{pp:prop_2_2}]\label{lem:f_approx_ub}
 Let $p\in[1,\infty]$.
 For any $N\in\mathbb{N}_+$ and $\bm{g}\in\mathcal{G}$, there exists a bounded-parameter FNN $\bm{f}_{N,M}\in\mathcal{F}_{N,M}^D$ satisfying $\norm{\bm{g}-\bm{f}_{N,M}}_{\bar{\mathcal{B}}_{S,I},p}\le D^{1/p}\kappa\sqrt{D+E}MN^{-1/2}$, where $\kappa>0$ is the universal constant appearing in Proposition~\ref{pp:prop_2_2}.
\end{lemma}
We call the FNN $\bm{f}_{N,M}$ in Lemma~\ref{lem:f_approx_ub} an approximating FNN of $\bm{g}$.

\begin{remark}
 For simplicity of the discussion, $\mathcal{Q}_{\mathrm{a},M}$ is made larger than necessary.
 Also note that it would be more natural to work with an upper bound of $\norm{(\bm{b}_i,\bm{c}_i)}_{\bar{\mathcal{B}}_{S,I}}$ in view of Propositions~\ref{pp:prop_2_2} and \ref{pp:prop_2_2_sig}.
For simplicity of the discussion, however, we work with the upper bounds of $\norm{\bm{b}_i}_{\bar{\mathcal{B}}_S}$ and $\norm{\bm{c}_i}_{\bar{\mathcal{B}}_I}$ rather than the upper bound of $\norm{(\bm{b}_i,\bm{c}_i)}_{\bar{\mathcal{B}}_{S,I}}$,
although it might result in a looser evaluation. 
\end{remark}

Next, we use a technique called internal approximation \citep{Matthews:1993:non_fading_mem,Grigoryeva_and_Ortega:2018:esn_univ} to bound $\norm{V_{\bm{g},\bm{x}_\mathrm{init}}-V_{\bm{f}_{N,M},\bm{x}_\mathrm{init}}}_{\mathcal{T}_T,p}$ from above by $\norm{\bm{g}-\bm{f}_{N,M}}_{\bar{\mathcal{B}}_{S,I},p}$.
This technique approximates a filter of a dynamical system via constructing another dynamical system whose state-variable representation, which is also called internal representation, is close to the given one in terms of function.
We adopt this technique in Theorem~3.1 (iii) in \cite{Grigoryeva_and_Ortega:2018:esn_univ} to our case and obtain the following bound.
\begin{lemma}\label{lem:filter_nn_diff_ub}
 Let $\bm{g}\in\mathcal{G}$.
 Suppose that $\bm{f}:\bar{\mathcal{B}}_{S,I}\to\mathbb{R}^D$ is a map such that $\norm{\bm{g}-\bm{f}}_{\bar{\mathcal{B}}_{S,I},p}\le S-P$, where $P\in(0,S)$ is one of the parameters of $\mathcal{G}$ defined in Assumption A\ref*{asm:ds_asm_fin}-\ref{enm:uniformly_bounded}.
  Then, for any $\bm{x}_\mathrm{init}\in\bar{\mathcal{B}}_S$, one has $\norm{V_{\bm{g},\bm{x}_\mathrm{init}}-V_{\bm{f},\bm{x}_\mathrm{init}}}_{\mathcal{T}_T,p}\le\norm{\bm{g}-\bm{f}}_{\bar{\mathcal{B}}_{S,I},p}\sum_{i=0}^{T-1}L^i$,
  where $L$ is the Lipschitz constant of any $\bm{g}\in\mathcal{G}$ when given an input (see Assumption A\ref*{asm:ds_asm_fin}-\ref{enm:lipschitz_wrt_input}).
\end{lemma}
\begin{remark}
 This result is essentially specialization of contributions of the existing research and we do not claim it as our novel contribution.
 The result and the proof are provided only for the sake of completeness.
 In the above lemma, we have specialized Theorem~3.1 (iii) in \cite{Grigoryeva_and_Ortega:2018:esn_univ} in the case of left-infinite inputs to the case of finite-length inputs.
 Furthermore, the domain-preserving assumption on $\bm{f}$ in the original result is replaced by a sufficient condition $\norm{\bm{g}-\bm{f}}_{\bar{\mathcal{B}}_{S,I},p}\le S-P$ with the additional assumption that the range of $\bm{g}$ is smaller than its domain (Assumption A\ref*{asm:ds_asm_fin}-\ref{enm:uniformly_bounded}).
 This type of sufficient condition is used in the proof of Theorem~4.1 of \cite{Grigoryeva_and_Ortega:2018:esn_univ}.
\end{remark}

The weak universality of RNN reservoir systems against $\mathcal{G}$ is a straightforward consequence of Lemma~\ref{lem:f_approx_ub} and Lemma~\ref{lem:filter_nn_diff_ub}.

\section{Uniform strong universality: Finite-length inputs}\label{sec:unif_strong_univ}
\subsection{Overview}\label{ssec:overview_usu}
The weak universality discussed in Section~\ref{sec:weak_univ} allows one to choose reservoirs according to a target dynamical system.
In this section, we study universal strong universality for the case of finite-length inputs,
which is more demanding than weak universality. 
More concretely, under the setting of uniform strong universality,
given a prescribed approximation error bound and the fixed class of target dynamical systems, one chooses a reservoir, but it is fixed once it is chosen.
Then, one is only allowed to adjust readout according to each target dynamical system in the class to approximate the target within the prescribed approximation error bound. 
The tools developed in this section will then be used to prove the uniform strong universality for a case of left-infinite inputs in Section~\ref{sec:usu_lf}.

We define uniform strong universality formally in Section~\ref{ssec:unif_str_univ}.
We state that the family of RNN reservoir systems has uniform strong universality in Section~\ref{sec:result_rnn_res_sys}.
An outline of proof is provided in Section~\ref{ssec:outline_proof_usu}.

\subsection{Definition}\label{ssec:unif_str_univ}
In order to define uniform strong universality, we define the worst approximation error.
This is the worst case of approximation error of a reservoir system when its reservoir map is fixed and when only its linear readout can be adjusted according to each target dynamical system in the class $\mathcal{G}$.
Given an $N_\mathrm{res}$-dimensional reservoir map, we assume that the readout for each target $\bm{g}\in\mathcal{G}$ is determined by what we call a readout-choosing rule $k:\mathcal{G}\to\mathbb{R}^{D\times N_\mathrm{res}}$.

\begin{definition}[worst approximation error]\label{def:approx_error}
Worst approximation error for $\mathcal{G}$ of a pair of an $N_\mathrm{res}$-dimensional reservoir map $\bm{r}$, its initial state $\bm{s}_\mathrm{init}\in\mathbb{R}^{N_\mathrm{res}}$ and its readout-choosing rule $k\in\{\mathcal{G}\to\mathbb{R}^{D\times N_\mathrm{res}}\}$ is defined as
\begin{align}
 \mathrm{werr}(\mathcal{G},(\bm{r},\bm{s}_\mathrm{init},k)):=\sup_{\bm{g}\in\mathcal{G}}\mathrm{err}(\bm{g},(\bm{r},\bm{s}_\mathrm{init},k(\bm{g})))=\sup_{\bm{g}\in\mathcal{G}}\norm{V_{\bm{g},\bm{x}_\mathrm{init}}-V_{\bm{r},\bm{s}_\mathrm{init}}^{k(\bm{g})}}_{\mathcal{T}_T,p}.
\end{align}
\end{definition}

We define the uniform strong universality of a family of reservoir systems for a target class of dynamical systems.
This is the property such that we can choose a reservoir system out of the family whose worst approximation error against the target class is arbitrarily small.
\begin{definition}[uniform strong universality]\label{def:us_universality}
 We say that the family of reservoir systems made with a family $\mathcal{R}\subset\{\mathbb{R}^{N_\mathrm{res}}\times\mathbb{R}^E\to\mathbb{R}^{N_\mathrm{res}},N_\mathrm{res}\in\mathbb{N}_+\}$ of reservoir maps has uniform strong universality for a class $\mathcal{G}$ of dynamical systems if and only if, for all $\varepsilon>0$, there exist $N_\mathrm{res}\in\mathbb{N}_+$, $N_\mathrm{res}$-dimensional reservoir map $\bm{r}\in\mathcal{R}$, its initial state $\bm{s}_\mathrm{init}\in\mathbb{R}^{N_\mathrm{res}}$ and its readout-choosing rule $k\in\{\mathcal{G}\to\mathbb{R}^{D\times N_\mathrm{res}}\}$ such that $\mathrm{werr}(\mathcal{G},(\bm{r},\bm{s}_\mathrm{init},k)):=\sup_{\bm{g}\in\mathcal{G}}\norm{V_{\bm{g},\bm{x}_\mathrm{init}}-V_{\bm{r},\bm{s}_\mathrm{init}}^{k(\bm{g})}}_{\mathcal{T}_T,p}\le\varepsilon$.
\end{definition}

\subsection{Result}\label{sec:result_rnn_res_sys}
When inputs are finite-length, the family of RNN reservoir systems has uniform strong universality for $\mathcal{G}$.
\begin{theorem}[uniform strong universality]\label{th:usu_fnn}\ 
 When inputs are finite-length, under Assumption~\ref{asm:ds_asm_fin}, the family of RNN reservoir systems has uniform strong universality for $\mathcal{G}$.
\end{theorem}

\subsection{Outline of proof}\label{ssec:outline_proof_usu}
\subsubsection{Overview}\label{sssec:overview_proof_usu}

The main idea for proving the uniform strong universality of the family of RNN reservoir systems for $\mathcal{G}$ is
to upgrade the weak universality described in Section~\ref{sec:weak_univ} 
via a combination of parallel concatenation (e.g., \cite{Schwenker_and_Labib:2009:esn_ensemble,Grigoryeva_el_al:2014:pr_res_univ}) and covering.
The basic idea is to concatenate many but finite RNN reservoir systems used for proving the weak universality.
Such RNN reservoir systems are chosen on the basis of covering.
Then, for any target dynamical system $\bm{g}\in\mathcal{G}$, we can adjust the readout of the concatenated RNN reservoir system so that it passes the output of the component RNN reservoir that approximates $\bm{g}$ the best.
Thus, the worst approximation error is bounded from above by the summation of filter approximation error and covering error.
Since this upper bound of worst approximation error can be made arbitrarily small by choosing a sufficiently large concatenated RNN reservoir system, the uniform strong universality holds.

\begin{figure}
  \centering
  \def\IsCovering#1{\mathrel{\xrightarrow{\hbox to26mm{\hss$\scriptstyle#1$\hss}}}}
  \def\CoverLayout#1#2#3{\setbox0\hbox{$#1$}\hbox to0pt{\kern-\wd0%
    $#1\IsCovering{#2}#3$\hss}}
  \begin{tikzpicture}
    \node at (0,13) [right] {Parameters:};
    \node at (3.2,13) {\CoverLayout{\bar{\mathcal{Q}}_{\mathrm{a},M,\Gamma}}{|\cdot|,\frac{\Gamma}{4\sqrt{M}\cdot20ND^{1/p}}}{\mathcal{Q}_{\mathrm{a},M}}};
    \node at (3.2,12) {\CoverLayout{\bar{\mathcal{Q}}_{\mathrm{b},M,\Gamma}}{\|\cdot\|_q,\frac{\Gamma}{2S\sqrt{M}\cdot20ND^{1/p}}}{\mathcal{Q}_{\mathrm{b},M}}};
    \node at (3.2,11) {\CoverLayout{\bar{\mathcal{Q}}_{\mathrm{c},M,\Gamma}}{\|\cdot\|_q,\frac{\Gamma}{2I\sqrt{M}\cdot20ND^{1/p}}}{\mathcal{Q}_{\mathrm{c},M}}};
    \node at (3.2,10) {\CoverLayout{\bar{\mathcal{Q}}_{\mathrm{d},M,\Gamma}}{|\cdot|,\frac{\Gamma}{2\sqrt{M}\cdot20ND^{1/p}}}{\mathcal{Q}_{\mathrm{d},M}}};
    \node at (3.2,9) {\CoverLayout{\bar{\mathcal{Q}}_{\mathrm{e},M,\Gamma}}{|\cdot|,\frac{\Gamma}{5D^{1/p}}}{\mathcal{Q}_{\mathrm{e},M}}};
    \draw[-] (0,8)--(12,8);
    \node at (4.7,8) {\Large$\Downarrow$};
    \node at (5.8,7.8) {Lemma~\ref{lem:approx_rv_funcs_covering_wrt_map_norm}};
    \node at (0,7.2) [right] {Real-valued};
    \node at (0,6.7) [right] {FNNs:};
    \node at (3.2,7) {\CoverLayout{\bar{\mathcal{F}}_{N,M,\Gamma}^1}{\|\cdot\|_{\bar{\mathcal{B}}_{S,I},\infty},\Gamma/D^{1/p}}{\mathcal{F}_{N,M}^1}};
    \draw[-] (0,6)--(12,6);
    \node at (4.7,6) {\Large$\Downarrow$};
    \node at (5.8,5.8) {Lemma~\ref{lem:approx_funcs_covering_wrt_map_norm}};
    \node at (10,5.8) {Lemma~\ref{lem:f_approx_ub}};
    \node at (0,5.4) [right] {Vector-valued};
    \node at (0,4.9) [right] {FNNs \&};
    \node at (0,4.4) [right] {dynamical};
    \node at (0,3.9) [right] {systems:};
    \node at (3.2,5) {\CoverLayout{\bar{\mathcal{F}}_{N,M,\Gamma}^D}{\|\cdot\|_{\bar{\mathcal{B}}_{S,I},p},\Gamma}{\mathcal{F}_{N,M}^D\xrightarrow{\|\cdot\|_{\bar{\mathcal{B}}_{S,I},p},D^{1/p}\kappa\sqrt{D+E}MN^{-1/2}}\mathcal{G}}};
    \draw [thin,-{Classical TikZ Rightarrow[length=0.9mm]}] (3.3,4.5) [out=285] to [in=180] (4.2,3)--(10.4,3) [out=0] to [in=265] (11.4,4.5);
    \node at (7.5,3.25) {$\scriptstyle\|\cdot\|_{\bar{\mathcal{B}}_{S,I},p},D^{1/p}\kappa\sqrt{D+E}MN^{-1/2}+\Gamma$};
    \node at (6.8,4.1) {{\Large$\Downarrow$}\quad Lemma~\ref{lem:approx_dyn_sys_by_cov_fnn}\quad{\Large$\Downarrow$}};
    \draw[-] (0,2)--(12,2);
    \node at (4.7,2) {\Large$\Downarrow$};
    \node at (5.8,1.8) {Lemma~\ref{lem:cov_fnn_fils_cov_tar_dyn_sys_fils}};
    \node at (0,1) [right] {Filters:};
    \node at (3.2,1) {\setbox0\hbox{$\bar{\mathcal{V}}_{N,M,\Gamma}$}
      \hbox to0pt{\kern-\wd0%
    $\bar{\mathcal{V}}_{N,M,\Gamma}\xrightarrow{\hbox to78mm{\hss$\scriptstyle\|\cdot\|_{\mathcal{T}_T,p},p(D,E,M,N,\Gamma)\sum_{i=0}^{T-1}L^i$\hss}}\mathcal{V}_{\mathcal{G}}$\hss}};
  \end{tikzpicture}
  \caption{Relations of coverings used in the proof.}
  \label{fig:covs}
\end{figure}
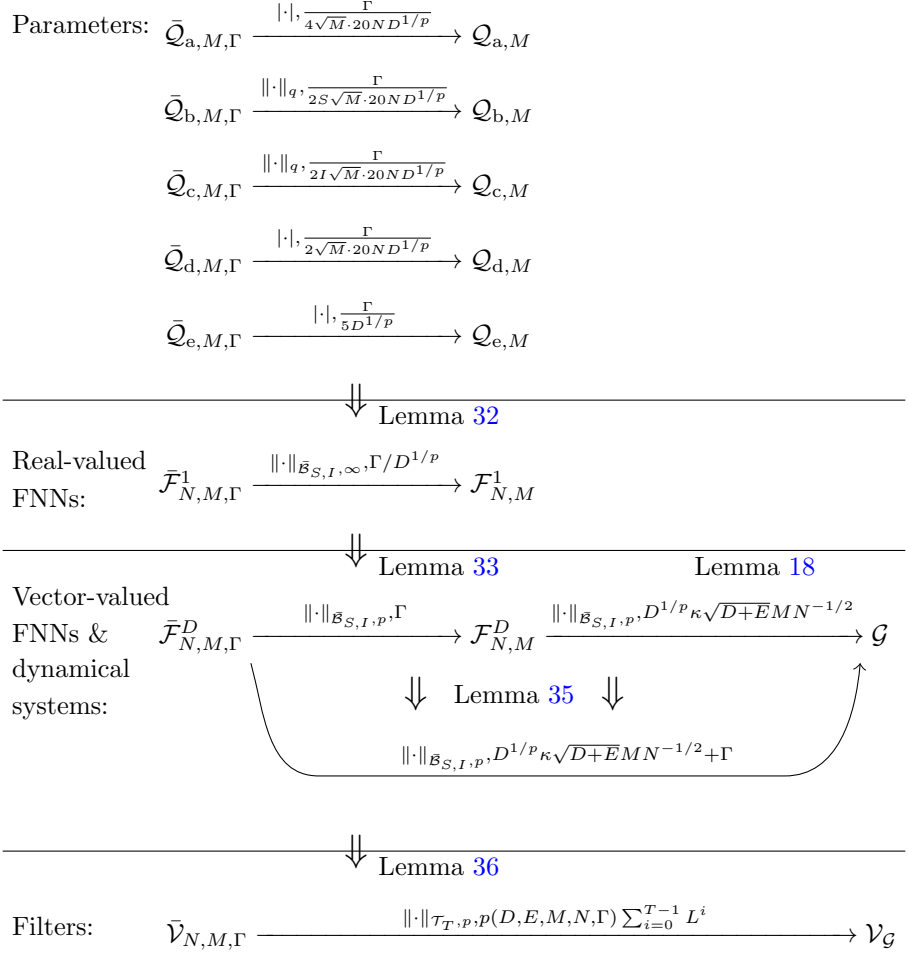

The following sections construct the concatenated RNN reservoir system step by step.
Section~\ref{sssec:cov_param_rv_fnn} constructs coverings of parameters of the real-valued bounded-parameter FNNs $\mathcal{F}_{N,M}^1$.
Section~\ref{sssec:cov_rv_fnn} constructs a covering $\bar{\mathcal{F}}_{N,M,\Gamma}^1$ of the real-valued bounded-parameter FNNs $\mathcal{F}_{N,M}^1$ on the basis of the coverings of their parameters via techniques in \cite{Chen_et_al:2020:rnn_gen_ub}.
Section~\ref{sssec:cov_fnn} constructs the covering $\bar{\mathcal{F}}_{N,M,\Gamma}^D$ of the bounded-parameter FNNs $\mathcal{F}_{N,M}^D$.
Section~\ref{sssec:approx_dyn_sys_by_cov_fnn} confirms that the covering $\bar{\mathcal{F}}_{N,M,\Gamma}^D$ is a covering of $\mathcal{G}$ not only in terms of functions but also in terms of filters.
These steps are depicted in Figure~\ref{fig:covs}.
Section~\ref{sssec:pl_con_fnn} concatenates the FNNs in the covering $\bar{\mathcal{F}}_{N,M,\Gamma}^D$ to construct a large RNN reservoir system.
Lastly, Section~\ref{sssec:usu_fin_conc} derives the upper bound of worst approximation error of the concatenated RNN reservoir system on the basis of the covering result developed so far.

\subsubsection{Covering of parameters of real-valued feedforward neural networks}\label{sssec:cov_param_rv_fnn}
We construct coverings of parameters of real-valued bounded-parameter FNNs. 
Before constructing the coverings of parameters of FNNs, we define the concept of covering as follows.
\begin{definition}[covering (see, e.g., Section~10.4 in \cite{Anthony_and_Bartlett:1999:nnl} or \cite{Chen_et_al:2020:rnn_gen_ub})]\label{def:covering}
Given a normed vector space $(\mathcal{V},\|\cdot\|)$
and its subset $\mathcal{S}\subset\mathcal{V}$,
as well as $\Gamma>0$, 
we say that a subset $\bar{\mathcal{S}}_\Gamma\subseteq\mathcal{S}$ is a $(\|\cdot\|,\Gamma)$-covering of $\mathcal{S}$
if and only if for any $s\in\mathcal{S}$ there exists $\bar{q}_s\in\bar{\mathcal{S}}_\Gamma$ such that $\norm{s-\bar{q}_s}\le \Gamma$.
In the above case, we say that $\bar{q}_s$ covers $s$.
We also write $\bar{\mathcal{S}}_\Gamma\xrightarrow{\|\cdot\|,\Gamma}\mathcal{S}$
to represent that $\bar{\mathcal{S}}_\Gamma$
is a $(\|\cdot\|,\Gamma)$-covering of $\mathcal{S}$. 
\end{definition}

The covering relation is transitive in the sense of the following
proposition, which can be proved straightfowardly via the triangle inequality.
\begin{proposition}
\label{pp:covtrans}
Given a normed vector space $(\mathcal{V},\|\cdot\|)$, 
assume that its subsets
$\mathcal{S}_1,\mathcal{S}_2,\mathcal{S}_3\subseteq\mathcal{V}$
satisfy $\mathcal{S}_1\xrightarrow{\|\cdot\|,\Gamma_1}\mathcal{S}_2$ 
and $\mathcal{S}_2\xrightarrow{\|\cdot\|,\Gamma_2}\mathcal{S}_3$
with $\Gamma_1,\Gamma_2>0$. 
One then has
$\mathcal{S}_1\xrightarrow{\|\cdot\|,\Gamma_1+\Gamma_2}\mathcal{S}_3$.
\end{proposition}

We construct coverings of parameters of $\mathcal{F}_{N,M}^1$ and confirm that they are finite.
\begin{definition}[coverings of parameters of FNNs]\label{def:cov_param_fnn}
For any $\Gamma>0$, let $\bar{\mathcal{Q}}_{\mathrm{a},M,\Gamma}$, $\bar{\mathcal{Q}}_{\mathrm{b},M,\Gamma}$, $\bar{\mathcal{Q}}_{\mathrm{c},M,\Gamma}$, $\bar{\mathcal{Q}}_{\mathrm{d},M,\Gamma}$ and $\bar{\mathcal{Q}}_{\mathrm{e},M,\Gamma}$ be the smallest $(|\cdot|,\frac{\Gamma}{4\sqrt{M}\cdot20ND^{1/p}})$-, $(\|\cdot\|_q,\frac{\Gamma}{2S\sqrt{M}\cdot20ND^{1/p}})$-, $(\|\cdot\|_q,\frac{\Gamma}{2I\sqrt{M}\cdot20ND^{1/p}})$-, $(|\cdot|,\frac{\Gamma}{2\sqrt{M}\cdot20ND^{1/p}})$- and $(|\cdot|,\frac{\Gamma}{5D^{1/p}})$-coverings of $\mathcal{Q}_{\mathrm{a},M}$, $\mathcal{Q}_{\mathrm{b},M}$, $\mathcal{Q}_{\mathrm{c},M}$, $\mathcal{Q}_{\mathrm{d},M}$ and $\mathcal{Q}_{\mathrm{e},M}$, respectively.
\end{definition}
\begin{lemma}[finiteness of coverings of parameters]\label{lem:fin_cov_params}
  For any $\Gamma>0$, one has
  \begin{align}
    \abs{\bar{\mathcal{Q}}_{\mathrm{a},M,\Gamma}}&\le8M\cdot20ND^{1/p}/\Gamma+1,
    \\
    \abs{\bar{\mathcal{Q}}_{\mathrm{b},M,\Gamma}}&\le(4M\cdot20ND^{1/p}/\Gamma+1)^D,
    \\
    \abs{\bar{\mathcal{Q}}_{\mathrm{c},M,\Gamma}}&\le(4M\cdot20ND^{1/p}/\Gamma+1)^E,
    \\
    \abs{\bar{\mathcal{Q}}_{\mathrm{d},M,\Gamma}}&\le4M\cdot20ND^{1/p}/\Gamma+1,
    \\
    \abs{\bar{\mathcal{Q}}_{\mathrm{e},M,\Gamma}}&\le M\cdot5D^{1/p}/\Gamma+1.
  \end{align}
\end{lemma}

\subsubsection{Covering of real-valued feedforward neural networks}\label{sssec:cov_rv_fnn}
We construct a covering of the family of real-valued bounded-parameter FNNs $\mathcal{F}_{N,M}^1$.
This is constructed via the following inequality and the coverings of the sets of parameters defined in Definition~\ref{def:cov_param_fnn}, as in~\cite{Chen_et_al:2020:rnn_gen_ub,Sreekumar_and_Goldfeld:2022:barron_unif_sig2}.
\begin{lemma}\label{lem:diff_approx_funcs_ub_wrt_params}
For any $f_{N,M},f'_{N,M}\in\mathcal{F}_{N,M}^1$ with parameters $(\{(a_n,\bm{b}_n,\bm{c}_n,d_n)\}_{n=1}^{4N},e)$, $(\{(a'_n,\bm{b}'_n,\bm{c}'_n,d'_n)\}_{n=1}^{4N},e')$, respectively, it holds that 
\begin{align}
 &\norm{f_{N,M}-f'_{N,M}}_{\bar{\mathcal{B}}_{S,I},\infty}\nonumber
 \\
 &\le4\sqrt{M}\sum_{n=1}^{4N}\abs{a_n-a_n'}
 +2S\sqrt{M}\sum_{n=1}^{4N}\norm{\bm{b}_n-\bm{b}_n'}_q\nonumber
 \\
 &+2I\sqrt{M}\sum_{n=1}^{4N}\norm{\bm{c}_n-\bm{c}_n'}_q
 +2\sqrt{M}\sum_{n=1}^{4N}\abs{d_n-d_n'}+\abs{e-e'}.\label{eq:g_diff_ub_for_cov}
\end{align}
\end{lemma}

\begin{lemma}[real-valued covering FNNs]\label{lem:approx_rv_funcs_covering_wrt_map_norm}
 Let $\bar{\mathcal{F}}_{N,M,\Gamma}^1:=\{f:\mathbb{R}^D\times\mathbb{R}^E\to\mathbb{R}\mid f(\bm{s},\bm{u}):=\sum_{n=1}^{4N}a_n\rho (\bm{b}_n^\top\bm{s}+\bm{c}_n^\top\bm{u}+d_n)+e,a_n\in\bar{\mathcal{Q}}_{\mathrm{a},M,\Gamma},\bm{b}_n\in\bar{\mathcal{Q}}_{\mathrm{b},M,\Gamma},\bm{c}_n\in\bar{\mathcal{Q}}_{\mathrm{c},M,\Gamma},d_n\in\bar{\mathcal{Q}}_{\mathrm{d},M,\Gamma},n=1,\ldots,4N,e\in\bar{\mathcal{Q}}_{\mathrm{e},M,\Gamma}\}$.
 Then, $\bar{\mathcal{F}}_{N,M,\Gamma}^1$ is a finite $(\|\cdot\|_{\bar{\mathcal{B}}_{S,I},\infty},\Gamma/D^{1/p})$-covering of $\mathcal{F}_{N,M}^1$.
\end{lemma}

\subsubsection{Covering of feedforward neural networks}\label{sssec:cov_fnn}
We bundle these FNNs to construct a family of FNN reservoir maps that is a covering of the bounded-parameter FNNs $\mathcal{F}_{N,M}^D$.
\begin{lemma}[covering FNNs]\label{lem:approx_funcs_covering_wrt_map_norm}
Let $\bar{\mathcal{F}}_{N,M,\Gamma}^D:=\{\bar{\bm{f}}_{N,M,\Gamma}:\mathbb{R}^D\times\mathbb{R}^E\to\mathbb{R}^D\mid\bar{\bm{f}}_{N,M,\Gamma}(\bm{x}):=(f_1(\bm{x}),\ldots,f_D(\bm{x})),f_1,\ldots,f_D\in\bar{\mathcal{F}}_{N,M,\Gamma}^1\}$. 
Then $\bar{\mathcal{F}}_{N,M,\Gamma}^D$ is a finite $(\|\cdot\|_{\bar{\mathcal{B}}_{S,I},p},\Gamma)$-covering of $\mathcal{F}_{N,M}^D$ and it holds that
\begin{align}
 &N_{N,M,\Gamma}:=\abs{\bar{\mathcal{F}}_{N,M,\Gamma}^D}
 \\
 &\le(8M\cdot20ND^{1/p}/\Gamma+1)^{4D(D+E+2)N+D}.
\end{align}
\end{lemma}
We call FNNs in $\bar{\mathcal{F}}_{N,M,\Gamma}^D$ covering FNNs.

\subsubsection{Approximation by covering feedforward neural networks}\label{sssec:approx_dyn_sys_by_cov_fnn}
We can approximate any target dynamical system $\bm{g}\in\mathcal{G}$ by a covering FNN $\bar{\bm{f}}_{N,M,\Gamma}\in\bar{\mathcal{F}}_{N,M,\Gamma}^D$ via approximating FNN  $\bm{f}_{N,M}\in\mathcal{F}_{N,M}^D$ of $\bm{g}$.
Lemma~\ref{lem:f_approx_ub} implies that 
$\mathcal{F}_{N,M}^D$ is a $(\|\cdot\|_{\bar{\mathcal{B}}_{S,I},p},D^{1/p}\kappa\sqrt{D+E}MN^{-1/2})$-covering of $\mathcal{G}$.
Combining it with Lemma~\ref{lem:approx_funcs_covering_wrt_map_norm}
via Proposition~\ref{pp:covtrans}, we have the following lemma.

\begin{lemma}[function approximation by covering]\label{lem:approx_dyn_sys_by_cov_fnn}
 Let $p(D,E,M,N,\Gamma):=D^{1/p}\kappa\sqrt{D+E}MN^{-1/2}+\Gamma$.
 The set $\bar{\mathcal{F}}_{N,M,\Gamma}^D$ of covering FNNs is a $(\|\cdot\|_{\bar{\mathcal{B}}_{S,I},p},p(D,E,M,N,\Gamma))$-covering of $\mathcal{G}$.
\end{lemma}

Furthermore, the covering FNNs induces a covering of the set of target dynamical systems in terms of norm of filter.
By Lemma~\ref{lem:approx_dyn_sys_by_cov_fnn} and Lemma~\ref{lem:filter_nn_diff_ub}, the following lemma holds trivially.
\begin{lemma}[filter approximation by covering]\label{lem:cov_fnn_fils_cov_tar_dyn_sys_fils}
 Fix $\bm{x}_\mathrm{init}\in\bar{\mathcal{B}}_S$.
 Let $\mathcal{V}_{\mathcal{G}}:=\{V_{\bm{g},\bm{x}_\mathrm{init}}\mid\bm{g}\in\mathcal{G}\}$
 and $\bar{\mathcal{V}}_{N,M,\Gamma}:=\{V_{\bar{\bm{f}}_{N,M,\Gamma},\bm{x}_\mathrm{init}}\mid\bar{\bm{f}}_{N,M,\Gamma}\in\bar{\mathcal{F}}_{N,M,\Gamma}^D\}$. 
 If $p(D,E,M,N,\Gamma)<S-P$, $\bar{\mathcal{V}}_{N,M,\Gamma}$ is a $(\|\cdot\|_{\mathcal{T}_T,p},p(D,E,M,N,\Gamma)\sum_{i=0}^{T-1}L^i)$-covering of $\mathcal{V}_{\mathcal{G}}$.
\end{lemma}

\def\fnn(#1,#2)#3{\begin{scope}[shift={(#1,#2)}]
    \node at (0,0) (u#3) {$\bm{u}(t)$};
    \node at (1.1,0) [rectangle,draw] (c#3) {$C^{[{#3}]}$};
    \node at (2.1,0) [circle,draw,inner sep=1pt] (pl1#3) {$+$};
    \node at (2.1,1) (d#3) {$\bm{d}^{[{#3}]}$};
    \node at (3.9,-0.7) [rectangle,draw] (B#3) {$B^{[{#3}]}$};
    \node at (2.9,0) [rectangle,draw] (sigma#3) {$\bm{\rho}$};
    \node at (3.9,0) [rectangle,draw] (A#3) {$A^{[{#3}]}$};
    \node at (4.9,0) [circle,draw,inner sep=1pt] (pl2#3) {$+$};
    \node at (4.9,1) (e#3) {$\bm{e}^{[{#3}]}$};
    \node at (6.2,0) (s#3) {$\bm{s}^{[{#3}]}(t)$};
    \draw[->] (u#3)--(c#3);
    \draw[->] (c#3)--(pl1#3);
    \draw[->] (d#3)--(pl1#3);
    \draw[->] (pl1#3)--(sigma#3);
    \draw[->] (sigma#3)--(A#3);
    \draw[->] (A#3)--(pl2#3);
    \draw[->] (pl2#3)--(s#3);
    \draw[->] (e#3)--(pl2#3);
    \draw[->] (5.3,0)--(5.3,-0.7)--(B#3);
    \draw[->] (B#3)--(2.1,-0.7)--(pl1#3);
  \end{scope}}

\def\plfnn(#1,#2)#3{\begin{scope}[shift={(#1,#2)}]
    \node at (0,2) (u) {$\bm{u}(t)$};
    \node at (1.5,2) [rectangle,draw] (c) {{\scriptsize$\left(
        \begin{array}{c}C^{[1]}\\C^{[2]}\\\vdots\\C^{[J]}
        \end{array}\right)$}};
    \node at (2.8,2) [circle,draw,inner sep=1pt] (pl1) {$+$};
    \node at (2.8,4) (d) {{\scriptsize$\left(
        \begin{array}{c}\bm{d}^{[1]}\\\bm{d}^{[2]}\\\vdots\\\bm{d}^{[J]}
        \end{array}\right)$}};
    \node at (6.3,0) [rectangle,draw] (B) {{\scriptsize$\left(\begin{array}{cccc}
          B^{[1]}&O&\cdots&O\\
          O&B^{[2]}&\cdots&O\\
          \vdots&\vdots&\ddots&\vdots\\
          O&O&\cdots&B^{[J]}
        \end{array}\right)$}};
    \node at (3.6,2) [rectangle,draw] (sigma) {$\bm{\rho}$};
    \node at (6.3,2) [rectangle,draw] (A) {{\scriptsize$\left(\begin{array}{cccc}
          A^{[1]}&O&\cdots&O\\
          O&A^{[2]}&\cdots&O\\
          \vdots&\vdots&\ddots&\vdots\\
          O&O&\cdots&A^{[J]}
        \end{array}\right)$}};
    \node at (9.0,2) [circle,draw,inner sep=1pt] (pl2) {$+$};
    \node at (9.0,4) (e) {{\scriptsize$\left(
        \begin{array}{c}\bm{e}^{[1]}\\\bm{e}^{[2]}\\\vdots\\\bm{e}^{[J]}
        \end{array}\right)$}};
    \node at (10.7,2) (s) {{\scriptsize$\left(
        \begin{array}{c}\bm{s}^{[1]}(t)\\\bm{s}^{[2]}(t)\\\vdots\\\bm{s}^{[J]}(t)
        \end{array}\right)$}};
    \draw[->] (u)--(c);
    \draw[->] (c)--(pl1);
    \draw[->] (d)--(pl1);
    \draw[->] (pl1)--(sigma);
    \draw[->] (sigma)--(A);
    \draw[->] (A)--(pl2);
    \draw[->] (pl2)--(s);;
    \draw[->] (e)--(pl2);
    \draw[->] (B)--(2.8,0)--(pl1);
    \draw[->] (9.4,2)--(9.4,0)--(B);
  \end{scope}}
  
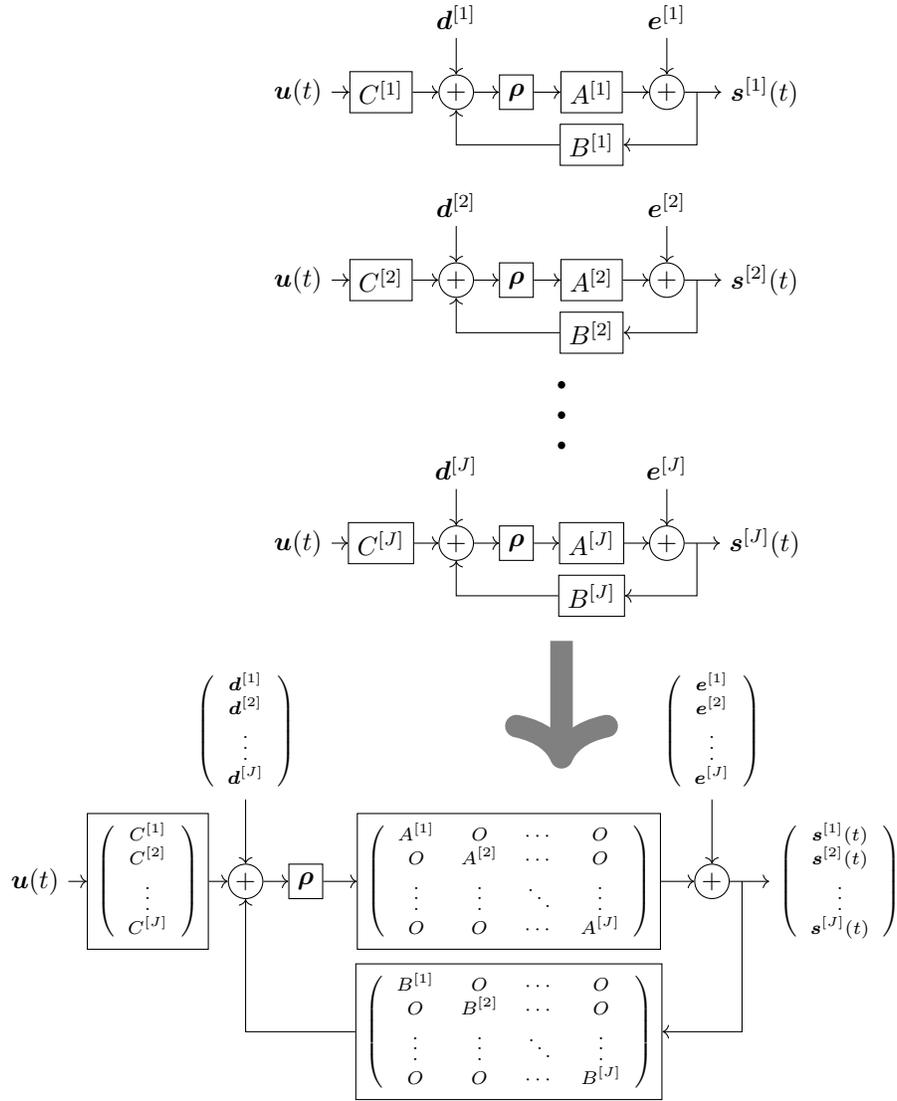
\begin{figure}
\centering
\begin{tikzpicture}
  \fnn(0.5,6){1}
  \fnn(0.5,3.5){2}
  \fnn(0.5,0){J}
  \draw[circle dotted/.style={dash pattern=on .025mm off 4mm,line cap=round},line width = 1mm,circle dotted] (4,1.3) -- (4,2.3);
  \draw[->,line width=3mm,gray] (4,-1.3)--(4,-3);
  \plfnn(-3,-6.5);
\end{tikzpicture}
\caption{Parallel concatenation of $J$ RNN reservoirs.}
\label{fig:pc}
\end{figure}

\subsubsection{Parallel concatenation of feedforward neural networks}\label{sssec:pl_con_fnn}
By concatenating in parallel all the covering FNNs $\bar{\mathcal{F}}_{N,M,\Gamma}^D$, we construct a concatenated FNN $\bm{F}_{N,M,\Gamma}$.
The RNN reservoir system made with $\bm{F}_{N,M,\Gamma}$ can approximate any target dynamical system by choosing readout appropriately.
Parallel concatenation is done as follows.
\begin{definition}[concatenated FNN]\label{def:parallel_nns}
Let $\{\bm{f}^{[i]}\}_{i=1}^{N_{N,M,\Gamma}}$ be all of the FNNs contained in $\bar{\mathcal{F}}_{N,M,\Gamma}^D$.
Assume that $\bm{f}^{[i]}:\mathbb{R}^D\times\mathbb{R}^E\to\mathbb{R}^D$ has parameters $A^{[i]}\in\mathbb{R}^{D\times4ND}$, $B^{[i]}\in\mathbb{R}^{4ND\times D}$, $C^{[i]}\in\mathbb{R}^{4ND\times E}$, $\bm{d}^{[i]}\in\mathbb{R}^{4ND}$ and $\bm{e}^{[i]}\in\mathbb{R}^D$, $i=1,\ldots,N_{N,M,\Gamma}$.
The concatenated FNN $\bm{F}_{N,M,\Gamma}:\mathbb{R}^{N_{N,M,\Gamma}D}\times\mathbb{R}^E\to\mathbb{R}^{N_{N,M,\Gamma}D}$ is defined by the following parallel concatenation.
\begin{align}
 \left(
\begin{array}{c}
 \bm{s}^{[1]}(t)\\
 \vdots\\
 \bm{s}^{[N_{N,M,\Gamma}]}(t) 
\end{array}
\right)
&=\bm{F}_{N,M,\Gamma}\left(
\left(
\begin{array}{c}
 \bm{s}^{[1]}(t-1)\\
 \vdots\\
 \bm{s}^{[N_{N,M,\Gamma}]}(t-1) 
\end{array}
\right)
 ,\bm{u}(t)
\right)
\\
 &:=
\left(
\begin{array}{c}
 \bm{f}^{[1]}(\bm{s}^{[1]}(t-1),\bm{u}(t))\\
 \vdots\\
 \bm{f}^{[N_{N,M,\Gamma}]}(\bm{s}^{[N_{N,M,\Gamma}]}(t-1),\bm{u}(t))
\end{array}
\right)\label{eq:pl_nn}
\\
&=\left(
   \begin{array}{cccc}
    A^{[1]}&O&\cdots&O\\
    O&A^{[2]}&\cdots&O\\
    \vdots&\vdots&\ddots&\vdots\\
    O&O&\cdots&A^{[N_{N,M,\Gamma}]}
   \end{array}
  \right)\nonumber
  \\
   &\cdot\bm{\rho}\left(
   \left(
   \begin{array}{cccc}
    B^{[1]}&O&\cdots&O\\
    O&B^{[2]}&\cdots&O\\
    \vdots&\vdots&\ddots&\vdots\\
    O&O&\cdots&B^{[N_{N,M,\Gamma}]}
   \end{array}
  \right)
   \left(
   \begin{array}{c}
    \bm{s}^{[1]}(t-1)\\
    \bm{s}^{[2]}(t-1)\\
    \vdots\\
    \bm{s}^{[N_{N,M,\Gamma}]}(t-1)
   \end{array}
  \right)\right.
  \\
   &\left.+
   \left(
   \begin{array}{c}
    C^{[1]}\\
    C^{[2]}\\
    \vdots\\
    C^{[N_{N,M,\Gamma}]}
   \end{array}
  \right)\bm{u}(t)
   +
   \left(
   \begin{array}{c}
    \bm{d}^{[1]}\\
    \bm{d}^{[2]}\\
    \vdots\\
    \bm{d}^{[N_{N,M,\Gamma}]}
   \end{array}
  \right)
 \right)
   +
   \left(
   \begin{array}{c}
    \bm{e}^{[1]}\\
    \bm{e}^{[2]}\\
    \vdots\\
    \bm{e}^{[N_{N,M,\Gamma}]}
   \end{array}
  \right).
\end{align}
We call the RNN reservoir made with $\bm{F}_{N,M,\Gamma}$ a concatenated RNN reservoir.
We call an RNN reservoir made with $\bm{f}^{[i]},i=1,\ldots,N_{N,M,\Gamma}$ an $i$th component RNN reservoir.
\end{definition}
Figure~\ref{fig:pc} illustrates parallel concatenation of RNN reservoirs.

Note that $\bm{F}_{N,M,\Gamma}$ is in the family $\mathcal{F}$ of FNN reservoir maps defined in Section~\ref{ssec:approx_res_sys}.

\subsubsection{Proof of Theorem~\ref{th:usu_fnn}}\label{sssec:usu_fin_conc}
We prove the uniform strong universality via studying an upper bound of the worst approximation error of the concatenated RNN reservoir system.
Lemma~\ref{lem:cov_fnn_fils_cov_tar_dyn_sys_fils} assures that the concatenated RNN reservoir system has the desired approximation capability.
Indeed, for any $\bm{g}\in\mathcal{G}$, we can choose a linear readout $W_{\bm{g}}$ such that it passes output of a component RNN reservoir that best approximates $\bm{g}$ as the final output.
Let $k_{N,M,\Gamma}:\mathcal{G}\to\mathbb{R}^{D\times N_{N,M,\Gamma}D}$ be such a readout-choosing rule, which we call a covering RNN selector in the following.
Recall that the approximation error in terms of filters in this case is bounded from above by $p(D,E,M,N,\Gamma)\sum_{i=0}^{T-1}L^i$ by Lemma~\ref{lem:cov_fnn_fils_cov_tar_dyn_sys_fils}, where $p(D,E,M,N,\Gamma):=D^{1/p}\kappa\sqrt{D+E}MN^{-1/2}+\Gamma$ is an upper bound of approximation error of $\bm{g}$ by a covering FNN (see Lemma~\ref{lem:approx_dyn_sys_by_cov_fnn}).
Therefore, the following worst approximation error holds trivially.
\begin{theorem}[upper bound of the worst approximation error]\label{th:concatenated_fnn_error}
 Assume Assumption~\ref{asm:ds_asm_fin} and $p(D,E,M,N,\Gamma)\le S-P$.
 Then, it holds that
 \begin{align}
  \mathrm{werr}(\mathcal{G},(\bm{F}_{N,M,\Gamma},\bm{x}_{N,M,\Gamma},k_{N,M,\Gamma}))\le p(D,E,M,N,\Gamma)\sum_{i=0}^{T-1}L^i,\label{eq:main_ub}
 \end{align}
 where $\bm{x}_{N,M,\Gamma}:=(\bm{x}_\mathrm{init},\ldots,\bm{x}_\mathrm{init})\in\mathbb{R}^{N_{N,M,\Gamma}D}$ and where $\bm{x}_\mathrm{init}\in\bar{\mathcal{B}}_S$.
\end{theorem}

The upper bound in Theorem~\ref{th:concatenated_fnn_error} can be made arbitrarily small when $N$ is sufficiently large and when $\Gamma$ is sufficiently small.
This means the uniform strong universality of the family of RNN reservoir systems against $\mathcal{G}$.
\qed

\section{Uniform strong universality: Left-infinite inputs}\label{sec:usu_lf}
\subsection{Overview}\label{ssec:usu_lf_overview}
We extend the result of the uniform strong universality to the case with left-infinite input sequences $\{\bm{u}(t)\}_{t\in\mathcal{T}_{-\infty}}\in(\bar{\mathcal{B}}_I)^{-\infty}$, where $\mathcal{T}_{-\infty}:=\{\ldots,-1,0\}$ is the left-infinite set of time indices, and where $\mathcal{B}^{-\infty}:=\cdots\times\mathcal{B}\times\mathcal{B}$ for a set $\mathcal{B}$.

In the left-infinite case, one must pay attention to uniqueness of state sequences of reservoirs and dynamical systems given an input sequence, which is called the echo state property (ESP) \citep{Jaeger:2001:esn}.
For avoiding the approximation problem becoming highly nontrivial, we have to limit the class of target dynamical systems to that with ESP.
Also, even though a target dynamical system has ESP, it is not guaranteed that the RNN reservoir systems constructed via the internal approximation will have ESP (see Theorem~3.1~(iii) in \cite{Grigoryeva_and_Ortega:2018:esn_univ} and Section~\ref{ssec:lim_ex_approach}).
Therefore, we propose use of a cascade structure of an RNN reservoir system for avoiding this problem.

Section~\ref{ssec:dyn_sys_inf} defines target dynamical systems in the case of left-infinite inputs.
Section~\ref{ssec:def_usu_lf} defines the universal strong universality and extends related notions for the case of left-infinite inputs.
Section~\ref{ssec:res_usu_inf} states the result.
Section~\ref{ssec:outline_proof_usu_inf} describes an outline of proof.

\subsection{Assumptions on target dynamical systems}\label{ssec:dyn_sys_inf}
Unlike the case of finite-length inputs, a dynamical system or a reservoir does not necessarily produce a unique state sequence given a left-infinite input sequence:
Consider a dynamical system with a constant input sequence, which
defines an autonomous system.
Then for any left-infinite state sequence $\{\bm{x}(t)\}_{t\in\mathcal{T}_{-\infty}}$
of that system,
an arbitrarily right-shifted sequence $\{\bm{x}(t-m)\}_{t\in\mathcal{T}_{-\infty}}$
with $m\in\mathbb{N}$ is a valid left-infinite state sequence for that system.
It implies that, when a dynamical system has a non-constant state sequence
for a constant input sequence, it does not produce a unique state sequence
for the same constant input sequence. 
We say that a dynamical system or a reservoir has ESP when it produces a unique state sequence for each input sequence $\bm{u}\in(\bar{\mathcal{B}}_I)^{-\infty}$.

We assume that target dynamical systems have ESP for the following reasons.
If RNN reservoir systems for approximating target dynamical systems do not have ESP, the outputs of RNN reservoir systems are not uniquely determined,
which would make evaluation of approximation errors highly nontrivial. 
On the other hand, if we assume RNN reservoir systems to have ESP, such RNN reservoir systems can only approximate target dynamical systems with ESP because the outputs of RNN reservoir systems are unique given an input sequence.
Also, many existing researches about universality of reservoir systems \citep{Grigoryeva_and_Ortega:2018:esn_univ,Grigoryeva_and_Ortega:2018:sas_univ,Gonon_and_Ortega:2021:esn_univ_esp,Gonon_et_al:2021:rc_app_ub,Li_and_Yang:2025:univ_esn_random_weights,Gonon_et_al:2024:inf_res_comp,Cuchiero_et_al:2021:strong_univ} assume the uniqueness of an output given an input sequence for approximation targets, which are certain classes of filters or functionals.

For guaranteeing ESP, we assume target dynamical systems to be uniformly state contracting \citep{Jaeger:2010:esn_note}.
This property is equivalent to ESP under some conditions, which are satisfied in our problem setting (see Proposition~5 in \cite{Jaeger:2010:esn_note}).
We use $p$-norm while the original one considers Euclidean norm.
\begin{definition}[uniform state contracting property]\label{def:usc}
 A dynamical system $\bm{g}:\bar{\mathcal{B}}_S\times\bar{\mathcal{B}}_I\to\bar{\mathcal{B}}_S$ is uniformly state contracting if and only if there exists a null sequence $\{\Delta_t\}_{t=1}^\infty$, $\Delta_t\in(0,\infty)$, $t\in\mathbb{N}_+$, $\lim_{t\to\infty}\Delta_t=0$, such that $\forall T\in\mathbb{N}_+,\forall\bm{x}\in\bar{\mathcal{B}}_S,\forall\bm{x}'\in\bar{\mathcal{B}}_S,\forall\bm{u}\in(\bar{\mathcal{B}}_I)^T$, $\norm{V_{\bm{g},\bm{x}}(\bm{u})_0-V_{\bm{g},\bm{x}'}(\bm{u})_0}_p<\Delta_T$.
\end{definition}

The null sequence represents how a dynamical system contracts its states.
For stating uniform strong universality, there should exist a common null sequence satisfying Definition~\ref{def:usc} for all target dynamical systems in a target class.
\begin{definition}[contracting target dynamical systems]\label{def:contracting_tar_ds}
 Let $\{\Delta_{\mathrm{c},t}\}_{t=1}^\infty$, $\Delta_{\mathrm{c},t}\in(0,\infty)$, $t\in\mathbb{N}_+$ be a fixed null sequence.
 Let $\mathcal{G}_\mathrm{c}\subset\mathcal{G}$ be a subset of the target dynamical systems defined in Assumption~\ref{asm:ds_asm_fin} that are uniformly state contracting with $\{\Delta_{\mathrm{c},t}\}_{t=1}^\infty$.
We call $\mathcal{G}_\mathrm{c}$ a set of contracting target dynamical systems.
\end{definition}

\begin{example}\label{ex:strictly_cont_tar_sys}
 For example, when the target dynamical systems $\mathcal{G}$ defined in Assumption~\ref{asm:ds_asm_fin} satisfy Assumption~A\ref{asm:ds_asm_fin}-\ref{enm:lipschitz_wrt_input} with a Lipschitz constant $L\in(0,1)$, all of the systems in $\mathcal{G}$ are uniformly state contracting with the null sequence $\{2SL_\mathrm{sc}^t\}_{t=1}^\infty$, where $L_\mathrm{sc}\in(L,1)$.
 This can easily be checked via recursive computation based on the Lipschitz continuity.
In the following parts of the paper, $\mathcal{G}_\mathrm{sc}$ denotes the subset of such strictly contracting target dynamical systems.
\end{example}

\subsection{Definition of uniform strong universality}\label{ssec:def_usu_lf}
We define uniform strong universality in the case of left-infinite inputs.
For this purpose, we extend related notions to those for the case of left-infinite inputs.

We define filter \citep{Grigoryeva_and_Ortega:2018:esn_univ} for left-infinite inputs.
We call an operator $U$ that maps a sequence indexed with $\mathcal{T}_{-\infty}$ to another sequence indexed with $\mathcal{T}_{-\infty}$ a left-infinite filter.
Let $\mathcal{A}$ be a set.
For any left-infinite filter $U:\mathcal{A}^{\mathcal{T}_{-\infty}}\to(\mathbb{R}^D)^{\mathcal{T}_{-\infty}}$, let $U(\bm{a})_t\in\mathbb{R}^D$, $t\in\mathcal{T}_{-\infty}$ denote the output of the left-infinite filter at time step $t$ given an input sequence $\bm{a}\in\mathcal{A}^{\mathcal{T}_{-\infty}}$.

As in Definition~\ref{def:dyn_sys_filter}, we can define left-infinite filters $U_{\bm{g}}:(\bar{\mathcal{B}}_I)^{\mathcal{T}_{-\infty}}\to(\mathbb{R}^D)^{\mathcal{T}_{-\infty}}$ and $U_{\bm{g}}^{\bm{h}}:(\bar{\mathcal{B}}_I)^{\mathcal{T}_{-\infty}}\to(\mathbb{R}^{D_\mathrm{out}})^{\mathcal{T}_{-\infty}}$ given a dynamical system $(\bm{g},\bm{h})$ with ESP, that is, a state map $\bm{g}:\mathbb{R}^D\times\bar{\mathcal{B}}_I\to\mathbb{R}^D$ with ESP and a readout $\bm{h}:\mathbb{R}^D\to\mathbb{R}^{D_\mathrm{out}}$.
Note that left-infinite filters for dynamical systems do not need initial states.
We can also define left-infinite filters for reservoir systems in the same manner.

We define the norm of left-infinite filters as in \cite{Grigoryeva_and_Ortega:2018:esn_univ}.
Let $l^{-\infty}(\mathbb{R}^a):=\{\bm{z}\in(\mathbb{R}^a)^{\mathcal{T}_{-\infty}}\mid\sup_{t\in\mathcal{T}_{-\infty}}\|\bm{z}_t\|_p<\infty\}$ be a space of bounded left-infinite sequence.
For the class
$\mathcal{C}((\bar{\mathcal{B}}_I)^{\mathcal{T}_{-\infty}},l^{-\infty}(\mathbb{R}^a))$ of continuous filters whose output is bounded,
we can define the supremum norm as
$\norm{U}_{\mathcal{T}_{-\infty},p}:=\sup_{\bm{u}\in(\bar{\mathcal{B}}_I)^{\mathcal{T}_{-\infty}}}\sup_{t\in\mathcal{T}_{-\infty}}
\|U(\bm{u})_t\|_p$ for
$U\in\mathcal{C}((\bar{\mathcal{B}}_I)^{\mathcal{T}_{-\infty}},l^{-\infty}(\mathbb{R}^a))$.

We can consider the norm of filter for a dynamical system $\bm{g}$ when it has ESP and when the range of the left-infinite state filter $U_{\bm{g}}$ is in $l^{-\infty}(\mathbb{R}^D)$.
In this case, we say that $\bm{g}$ induces a bounded filter.
The same holds for reservoirs and reservoir systems.
Note that the contracting target dynamical systems $\mathcal{G}_\mathrm{c}$ and reservoir systems that we introduce later induce bounded filters.

The left-infinite version of uniform strong universality is defined in terms of the norm of filters.
\begin{definition}[left-infinite version of uniform strong universality]\label{def:us_universality_lf}
 Assume left-infinite inputs.
 Assume that a class $\mathcal{G}$ of dynamical systems that induce bounded filters.
 We say that the family of reservoir systems made with a family $\mathcal{R}\subset\{\mathbb{R}^{N_\mathrm{res}}\times\mathbb{R}^E\to\mathbb{R}^{N_\mathrm{res}},N_\mathrm{res}\in\mathbb{N}_+\}$ of reservoir maps has uniform strong universality for $\mathcal{G}$ if and only if, for all $\varepsilon>0$, there exist $N_\mathrm{res}\in\mathbb{N}_+$, $N_\mathrm{res}$-dimensional reservoir map $\bm{r}\in\mathcal{R}$ inducing a bounded filter and its readout-choosing rule $k\in\{\mathcal{G}\to\mathbb{R}^{D\times N_\mathrm{res}}\}$ such that 
$\mathrm{werr}(\mathcal{G},(\bm{r},k)):=\sup_{\bm{g}\in\mathcal{G}}\norm{U_{\bm{g}}-U_{\bm{r}}^{k(\bm{g})}}_{\mathcal{T}_{-\infty},p}\le\varepsilon$.
 \end{definition}

\subsection{Result}\label{ssec:res_usu_inf}
The uniform strong universality holds as follows.
\begin{theorem}[uniform strong universality]\label{th:usu_fnn_lf}\ 
 When inputs are left-infinite,
 the family of RNN reservoir systems has uniform strong universality for the contracting target dynamical systems $\mathcal{G}_\mathrm{c}$.
\end{theorem}

\subsection{Outline of proof}\label{ssec:outline_proof_usu_inf}
First of all, note that the concatenated RNN reservoir systems constructed in Section~\ref{sec:unif_strong_univ} cannot directly apply to the case of left-infinite input sequences.
This is because it is not guaranteed that they will have ESP.
Therefore, we cascade a concatenated RNN reservoir and make a large RNN reservoir system.
Such a cascaded RNN reservoir system not only satisfies ESP but also inherits the approximation capability of the concatenated RNN.
Note that we are inspired by an RNN reservoir system with ESP in \cite{Gonon_and_Ortega:2021:esn_univ_esp}, which has a structure similar to cascading.

On the basis of the cascading, we derive an upper bound of worst approximation error.
This is done via the triangle inequality.
Firstly, we approximate a contracting target dynamical system $\bm{g}\in\mathcal{G}_\mathrm{c}$ with a cascaded dynamical system of $\bm{g}$.
Then, we further approximate the cascaded dynamical system of $\bm{g}$ with a cascaded version of concatenated RNN reservoir system.
In the following, these two steps are explained in order.

Firstly, a contracting target dynamical system $\bm{g}\in\mathcal{G}_\mathrm{c}$ is approximated with its cascaded version.
As shown in Figure~\ref{fig:dyn_sys_cas}, it simulates the original dynamical system with its latest length-$T$ inputs.
\begin{definition}[cascaded dynamical system]\label{def:cascade}
Let $\bm{x}(t)=(\bm{x}_1(t),\ldots,\bm{x}_T(t))$, where $\bm{x}_i(t)\in\mathbb{R}^D$.
Let $\bm{g}\in\mathcal{G}_\mathrm{c}$ and $\bm{x}_\mathrm{init}\in\bar{\mathcal{B}}_S$.
\begin{align}
\bm{x}(t)&=
\left(
\begin{array}{ccc}
 \bm{x}_1(t)\\
 \bm{x}_2(t)\\
 \vdots\\
 \bm{x}_T(t)
\end{array}
\right)
=c(\bm{g},\bm{x}_\mathrm{init},T)(\bm{x}(t-1),\bm{u}(t))
\\
&:=
\left(
\begin{array}{ccc}
 \bm{g}(\bm{x}_\mathrm{init},\bm{u}(t))\\
 \bm{g}(\bm{x}_1(t-1),\bm{u}(t))\\
 \vdots\\
 \bm{g}(\bm{x}_{T-1}(t-1),\bm{u}(t))
\end{array}
\right)\label{eq:cas}
\\
\bm{y}(t)&=W_{\mathrm{last},T}(I_{D\times D})\bm{x}(t):=[O_{D\times(T-1)D},I_{D\times D}]\bm{x}(t),\label{eq:ro_cas}
\end{align}
where $c(\bm{g},\bm{x}_\mathrm{init},T):\mathbb{R}^{TD}\times\mathbb{R}^E\to\mathbb{R}^{TD}$ is a state map.
We call $c(\bm{g},\bm{x}_\mathrm{init},T)$ an order-$T$ cascade of $\bm{g}$.
We call the dynamical system defined by \eqref{eq:cas} and \eqref{eq:ro_cas} a cascaded dynamical system.
\end{definition}
Note that the order-$T$ cascaded dynamical system of $\bm{g}$ obviously has ESP because the state of the cascade is uniquely determined by the latest length-$T$ inputs.
Note that, here again, we are inspired by another technique in \cite{Gonon_and_Ortega:2021:esn_univ_esp,Matthews:1993:non_fading_mem,Li_and_Yang:2025:univ_esn_random_weights} to approximate a filter by another filter that depends only on latest finite-length inputs.

\def\dynsys(#1,#2){\begin{scope}[shift={(#1,#2)}]
  \node at (0,0.4) (u) {$\bm{u}(t)$};
  \node at (1,0.5) [rectangle,draw] (g) {$\bm{g}$};
  \node at (2.3,0.5) [rectangle,draw] (I) {$I_{D\times D}$};
  \node at (3.6,0.5) (y) {$\bm{y}(t)$};
  
  \draw[->] (u)--(0.78,0.4);
  \draw[->] (g)--(I);
  \draw[->] (I)--(y);
  \draw[<-] (0.78,0.6)--(0.5,0.6)--(0.5,1.0)--(1.5,1.0)--(1.5,0.5);
\end{scope}}

\def\cas(#1,#2){\begin{scope}[shift={(#1,#2)}]
  \node at (0.5,2.4) (x_init) {$\bm{x}_\mathrm{init}$};
  \node at (0.5,1.35) (u) {$\bm{u}(t)$};
  
  \node at (2,2.25) [rectangle,draw] (g1) {$\bm{g}$};
  \node at (2,1.5) [rectangle,draw] (g2) {$\bm{g}$};
  \node at (2,0.85) (vdots) {$\vdots$};
  \node at (2,0) [rectangle,draw] (gT) {$\bm{g}$};
  
  \node at (4.42,1.5) [rectangle,draw] (W) {$W_{\mathrm{last},T}(I_{D\times D})$};
  
  \node at (6.5,1.5) (y) {$\bm{y}(t)$};
  
  \draw[->] (x_init)--(1.78,2.4);
  
  \draw[->] (u)--(1.78,1.35);
  \draw[->] (1.2,1.5)--(1.2,2.1)--(1.78,2.1);
  \draw[->] (1.2,1.5)--(1.2,-0.15)--(1.78,-0.15);
  
  \draw[->] (2.5,2.25)--(2.5,1.875)--(1.5,1.875)--(1.5,1.65)--(1.78,1.65);
  \draw[->] (2.5,1.5)--(2.5,1.125)--(1.5,1.125)--(1.5,1);
  \draw[->] (2.5,0.5)--(2.5,0.375)--(1.5,0.375)--(1.5,0.15)--(1.78,0.15);
  
  \draw[->] (g1)--(2.8,2.25)--(2.8,1.69)--(3.185,1.69);
  \draw[->] (g2)--(W);
  \draw[->] (gT)--(2.8,0)--(2.8,1.31)--(3.185,1.31);
  
  \draw[->] (W)--(y);
\end{scope}}
  
\begin{figure}
\centering
\begin{tikzpicture}
  \dynsys(0,0)
  \cas(4,0)
\end{tikzpicture}
\caption{A target dynamical system and its order-$T$ cascaded dynamical system.}
\label{fig:dyn_sys_cas}
\end{figure}
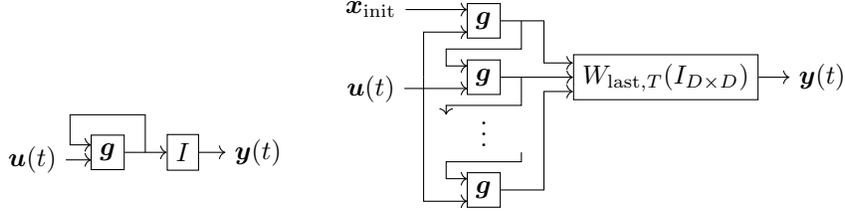

Although one could define a cascade for a dynamical system $\bm{g}_0\in\mathcal{G}$ with strong memory, the resulting cascaded dynamical system would have quite different dynamical properties from $\bm{g}_0$.
One can ensure, however, that the cascade
of a contracting dynamical system $\bm{g}\in\mathcal{G}_\mathrm{c}$ is indeed a good approximation of $\bm{g}$,
as stated in the following lemma.
Recall that $\{\Delta_{\mathrm{c},t}\}_{t=1}^\infty$ represents strength of contraction of the contracting target dynamical systems defined in Definition~\ref{def:contracting_tar_ds}.
Note that the following bound becomes small as $T$ increases because $\{\Delta_{\mathrm{c},t}\}_{t=1}^\infty$ is a null sequence.
\begin{lemma}[finite memory approximation]\label{lem:error_bound_fmf}
For any $\bm{g}\in\mathcal{G}_\mathrm{c}$ and $\bm{x}_\mathrm{init}\in\bar{\mathcal{B}}_S$, one has $\norm{U_{\bm{g}}-U_{c(\bm{g},\bm{x}_\mathrm{init},T)}^{W_{\mathrm{last},T}(I_{D\times D})}}_{\mathcal{T}_{-\infty},p}<\Delta_{\mathrm{c},T}$. 
\end{lemma}

Secondly, we approximate the order-$T$ cascaded dynamical system $c(\bm{g},\bm{x}_{\mathrm{init}},T)$.
For this purpose, we construct an RNN reservoir system whose reservoir map is an order-$T$ cascade of the concatenated FNN $\bm{F}_{N,M,\Gamma}$.
Note that the cascaded RNN reservoir systems have ESP.
In general, given two dynamical systems $\bm{g}_1,\bm{g}_2$ and readouts $W_1,W_2$, one can easily check the inequality 
\begin{align}
 \norm{U_{c(\bm{g}_1,\bm{x}_\mathrm{init},T)}^{W_{\mathrm{last},T}(W_1)}-U_{c(\bm{g}_2,\bm{x}_\mathrm{init},T)}^{W_{\mathrm{last},T}(W_2)}}_{\mathcal{T}_{-\infty},p}\le\norm{V_{\bm{g},\bm{x}_\mathrm{init}}^{W_1}-V_{\bm{g}_2,\bm{x}_\mathrm{init}}^{W_2}}_{\mathcal{T}_T,p}.\label{eq:reduce_to_fin_case}
\end{align}
This means that the approximation error of cascades can be reduced to an approximation error of finite-length filters.
Therefore, by choosing readout appropriately, the approximation error between cascades of $\bm{g}$ and $\bm{F}_{N,M,\Gamma}$ can be bounded from above by that in terms of finite-length filter in Theorem~\ref{th:concatenated_fnn_error}.
Therefore, the following upper bound of worst approximation error holds via the triangle inequality.
\begin{theorem}[upper bound of worst approximation error]\label{th:cas_pl_res_error}
 Assume $p(D,E,M,N,\Gamma)\le S-P$.
 Let $k_{N,M,\Gamma,T}:\mathcal{G}_\mathrm{c}\to\mathbb{R}^{D\times TN_{N,M,\Gamma}D}$, $k_{N,M,\Gamma,T}(\bm{g}):=W_{\mathrm{last},T}(k_{N,M,\Gamma}(\bm{g}))$, where $\bm{g}\in\mathcal{G}_\mathrm{c}$ and where $k_{N,M,\Gamma}:\mathcal{G}\to\mathbb{R}^{D\times N_{N,M,\Gamma}D}$ is the covering RNN selector defined in the paragraph just above Theorem~\ref{th:concatenated_fnn_error}.
 Then, it holds that
 \begin{align}
  \mathrm{werr}(\mathcal{G}_\mathrm{c},(c(\bm{F}_{N,M,\Gamma},\bm{x}_{N,M,\Gamma},T),k_{N,M,\Gamma,T}))
<\Delta_{\mathrm{c},T}+p(D,E,M,N,\Gamma)\sum_{i=0}^{T-1}L^i,\label{eq:werr_cascade_inf}
 \end{align}
 where $\bm{x}_{N,M,\Gamma}:=(\bm{x}_\mathrm{init},\ldots,\bm{x}_\mathrm{init})\in\mathbb{R}^{N_{N,M,\Gamma}D}$ and where $\bm{x}_\mathrm{init}\in\bar{\mathcal{B}}_S$.
\end{theorem}

Note that the cascade of an FNN is again an FNN.
Uniform strong universality of the family of RNN reservoir systems against $\mathcal{G}_\mathrm{c}$ can be stated on the basis of Theorem~\ref{th:cas_pl_res_error} because the upper bound can be made arbitrarily small by choosing a cascade of $\bm{F}_{N,M,\Gamma}$ with sufficiently large $N$ and $T$ and with sufficiently small $\Gamma$.

\section{Discussion}\label{sec:discussion}
\subsection{Scale and error of concatenated RNN reservoir systems}\label{ssec:scale_fnn}
Assume the strictly contracting dynamical systems defined in Example~\ref{ex:strictly_cont_tar_sys} for approximation targets.
The worst approximation error bound in Theorem~\ref{th:cas_pl_res_error} becomes $2SL_\mathrm{sc}^T+p(D,E,M,N,\Gamma)\sum_{i=0}^{T-1}L_\mathrm{sc}^i$ in this case, where $L_\mathrm{sc}<1$.
Suppose that we determine $\Gamma$ and $T$ according to the number $N$ of neurons in the hidden layer of FNNs.
Since $p(D,E,M,N,\Gamma)=D^{1/p}\kappa\sqrt{D+E}MN^{-1/2}+\Gamma$ consists of
the two terms, $D^{1/p}\kappa\sqrt{D+E}MN^{-1/2}$ corresponding to the approximation
error by FNNs and $\Gamma$ arising from the covering,
taking $\Gamma$ proportional to $N^{-1/2}$ makes these two terms comparable.
By regarding the term $\sum_{i=0}^{T-1}L_\mathrm{sc}^i$ as a constant less than $(1-L_\mathrm{sc})^{-1}$, the function approximation term $p(D,E,M,N,\Gamma)\sum_{i=0}^{T-1}L_\mathrm{sc}^i$ becomes $O(N^{-1/2})$.
Furthermore, the bound has yet another term $2SL_\mathrm{sc}^T$ coming from the finite memory approximation,
which can also be made comparable to the function approximation term by letting
$T=\lceil\log_{L_\mathrm{sc}}(N^{-1/2})\rceil$. 
Then, the order, in terms of $N$, of the worst approximation error and of the number of hidden nodes of a cascaded RNN reservoir system can be computed as follows.
Recall that $\bm{x}_{N,M,\Gamma}:=(\bm{x}_\mathrm{init},\ldots,\bm{x}_\mathrm{init})\in\mathbb{R}^{N_{N,M,\Gamma}D}$ and that $k_{N,M,\Gamma,T}$ is the readout-choosing rule defined in Theorem~\ref{th:cas_pl_res_error}.
\begin{proposition}[rates of error and number of hidden nodes]\label{pp:rate_err_and_node_cascade}
For any $N\in\mathbb{N}_+$, let $\Gamma$ be proportional to $N^{-1/2}$ and $T=\ceil{\log_{L_\mathrm{sc}}(N^{-1/2})}$.
Then, the following statements hold.
\begin{enumerate}
 \item[(i)] $\mathrm{werr}(\mathcal{G}_\mathrm{sc},(c(\bm{F}_{N,M,\Gamma},\bm{x}_{N,M,\Gamma},T),k_{N,M,\Gamma,T}))=O(N^{-1/2})$. 
 \item[(ii)] The total number of hidden nodes in the order-$T$ cascade of the concatenated FNN reservoir map is $T\cdot4DNN_{N,M,\Gamma}= O(N^{12D(D+E+2)N+3D+2})$.
\end{enumerate}
\end{proposition}

From this result, we can conclude that the RNN reservoir systems that we have constructed is too large to be practical.
Also, the construction techniques that we have adopted do not provide
any insights on how one can construct good reservoir systems
in practice. 
Therefore, Theorems~\ref{th:concatenated_fnn_error} and \ref{th:cas_pl_res_error}, which derive the worst approximation bounds, should rather be
regarded as an existence theorem.

\subsection{Comparison with existing work}\label{ssec:usu_sigsas}
We compare the uniform strong universality result in this paper and that in existing research \citep{Cuchiero_et_al:2021:strong_univ}.
For target dynamical systems, we consider a set of strictly contracting dynamical systems $\mathcal{G}_\mathrm{sc}'$ that satisfies both of Example~\ref{ex:strictly_cont_tar_sys} and the conditions assumed in \cite{Cuchiero_et_al:2021:strong_univ}.
For example, a particular subset of contracting linear systems can satisfy these conditions.

Theorem~4 in \cite{Cuchiero_et_al:2021:strong_univ} states strong universality of reservoir systems called signature state-affine systems (SigSAS), which can represent truncated Volterra series of filters to be approximated by adjusting linear readout according to each filter to be approximated.
Let $\bm{r}_{\lambda,l,j}$ be the reservoir map, where $\lambda$ is a parameter, where $l$ represents the ranges of time delay of the resulting Volterra series and where $j$ represents the order of the resulting Volterra series.

One can derive uniform strong universality result of SigSAS for $\mathcal{G}_\mathrm{sc}'$.
This is shown via the following worst approximation error bound, which is a corollary of Theorem~4 in \cite{Cuchiero_et_al:2021:strong_univ}.
This corollary is obtained by replacing a term $w_l^U$ in their bound that depends on each target filter $U$ with $2SL_\mathrm{sc}^{l+1}$.
Also note that one has to consider an input space smaller than that of the original bound in order to derive the worst approximation error bound.
\begin{corollary}[worst approximation error bound of SigSAS]\label{cor:lin_sys_usu_sigsas}
 Fix $p=2$, that is, we use Euclidean norm as the norm of the state space of $\mathcal{G}_\mathrm{sc}'$.
 Assume one-dimensional input sequences.
 Let $\mathcal{U}_{I',1}:=(\bar{\mathcal{B}}_{I'})^{-\infty}\cap l_1^{-\infty}(\mathbb{R})$, where $I'\in(0,I)$ and where $l_1^{-\infty}(\mathbb{R}):=\{(\ldots,u_{-1},u_0)\in\mathbb{R}^{-\infty}\mid\sum_{t\in\mathcal{T}_{-\infty}}\abs{u_t}<\infty\}$.
 For any $j,l\in\mathbb{N}_+$ and $0<\lambda<\min\{1,1/\sum_{i=0}^jI^i\}$, there exists a readout-choosing rule $k_{\lambda,l,j}:\mathcal{G}_\mathrm{sc}'\to\mathcal{L}(T^{l+1}(\mathbb{R}^{j+1}),\mathbb{R}^D)$ and it holds that
 \begin{align}
  &\mathrm{werr}(\mathcal{G}_\mathrm{sc}',(\bm{r}_{\lambda,l,j},k_{\lambda,l,j}))
  \\
  &=\sup_{\bm{g}\in\mathcal{G}_\mathrm{sc}'}\sup_{\bm{u}\in\mathcal{U}_{I',1}}\sup_{t\in\mathcal{T}_{-\infty}}\norm{U_{\bm{g}}(\bm{u})_t-U_{\bm{r}_{\lambda,l,j}}^{k_{\lambda,l,j}(\bm{g})}(\bm{u})_t}_2
  \\
  &\le2SL_\mathrm{sc}^{l+1}+S\left(1-\frac{I'}{I}\right)^{-1}\left(\frac{I'}{I}\right)^{j+1},
 \end{align}
 where $T^{l+1}(\mathbb{R}^{j+1})$ is the tensor space of order $(l+1)$ on $\mathbb{R}^{j+1}$ and where $\mathcal{L}(\mathcal{A},\mathcal{B})$ is the set of all of the linear maps from a vector space $\mathcal{A}$ to another vector space $\mathcal{B}$.
\end{corollary}

From Corollary~\ref{cor:lin_sys_usu_sigsas}, we can state the rates of scales and errors of the SigSAS for $\mathcal{G}_\mathrm{sc}'$.
For simplicity and ease of comparison, we fix $j=l$.
\begin{corollary}\label{cor:lin_sys_approx_rate_sigsas}
 In Corollary~\ref{cor:lin_sys_usu_sigsas}, assume further that $N:=j+1=l+1$.
 For any $j\in\mathbb{N}_+$ and $0<\lambda<\min\{1,1/\sum_{i=0}^jI^i\}$, 
 \begin{itemize}
  \item $\mathrm{werr}(\mathcal{G}_\mathrm{sc}',(\bm{r}_{\lambda,l,j},k_{\lambda,l,j}))=O(a^N)$, $a:=\max\{L_\mathrm{sc},I'/I\}<1$.
  \item The dimension of the state is $O(N^N)$.
 \end{itemize}
 Note that $\mathrm{werr}$ assumes that the input space is $\mathcal{U}_{I',1}$, $I'<I$.
\end{corollary}

We compare efficiency of SigSAS's and a cascaded RNN reservoir system on the basis of Proposition~\ref{pp:rate_err_and_node_cascade} and Corollary~\ref{cor:lin_sys_approx_rate_sigsas}.
Note that since SigSAS assumes one-dimensional input sequences, we have to set $E=1$ in Proposition~\ref{pp:rate_err_and_node_cascade} for the comparison.
Also note that Proposition~\ref{pp:rate_err_and_node_cascade} holds for the smaller input space $\mathcal{U}_{I',1}$, $I'<I$, which is assumed in Corollary~\ref{cor:lin_sys_approx_rate_sigsas}.
Therefore, both reservoir systems assume the same target $\mathcal{G}_\mathrm{sc}'$ and input space $\mathcal{U}_{I',1}$.
As a result of comparison, cascaded RNN reservoir systems are inferior in both of the order of the upper bound of worst approximation error and the order of the scale.
This result is in a sense anticipated,
as the Volterra series is thought to be more tailored to
approximation of nonlinear dynamical systems than
neural networks.

\subsection{A case of a monotone sigmoid activation function}\label{ssec:sig_func}
We can also show uniform strong universality for the family of RNN reservoir systems whose activation function is a monotonically increasing sigmoid function that is Lipschitz continuous.
By repeating the argument in Sections~\ref{sec:weak_univ}--\ref{sec:usu_lf} with the FNN approximation bound of Proposition~\ref{pp:prop_2_2_sig} in Section~\ref{ssec:nn_approx_sig}, we can derive a similar upper bound of the worst approximation error and uniform strong universality result.
Note that the Lipschitz continuity is needed for the transformation from \eqref{eq:bf_apply_Lip} to \eqref{eq:af_apply_Lip} in a proof of Lemma~\ref{lem:diff_approx_funcs_ub_wrt_params}.

In order to discuss orders of number of hidden nodes and of approximation error as done in Section~\ref{ssec:scale_fnn}, $\Lambda$ has to be parametrized by $N$.
For example, we can choose the logistic sigmoid function $\sigma_0(x)=1/(1+e^{-x})$ and can choose $\Lambda=N^{1/2}$.
Then, since $\delta(\Lambda)=2\int_{-1}^0\sigma_0(\Lambda x)dx=2[\Lambda^{-1}\log(1+e^{\Lambda x})]_{-1}^0=2\Lambda^{-1}\{\log2-\log(1+e^{-\Lambda})\}$, it holds that $\delta(\Lambda)=O(N^{-1/2})$ and we can repeat the discussion in Section~\ref{ssec:scale_fnn} to derive the orders.

\subsection{Complexity of concatenated recurrent neural network reservoir systems}\label{ssec:complexity}
Complexity of a reservoir system with a fixed reservoir map can be characterized by the dimension of its readout \citep{Yasumoto_and_Tanaka:2025:rc_complexities}.
From this point of view, the complexity of the concatenated RNN reservoir system is seemingly very large because the dimension of the readout is very high.
For example, when $D=1$, the dimension of the readout is $N_{N,M,\Gamma}$, which can be of the exponential order in $N$ as shown in Lemma~\ref{lem:approx_funcs_covering_wrt_map_norm}.
However, the readout does not need to take arbitrary values to achieve the uniform strong universality as we discussed in Section~\ref{sec:unif_strong_univ}.
Therefore, the actual complexity is smaller than expected.
In this subsection, we show that a complexity measure called pseudo-dimension (see, e.g., \cite{Mohri_et_al:2018:fml}) of the concatenated RNN reservoir systems can be much smaller than the dimension of the readout.

For simplicity, suppose a simple situation when one wants to approximate functions from a finite-length input sequence $\bm{u}\in\mathcal{U}:=(\bar{\mathcal{B}}_I)^T$ to the last output $V_{g,x_\mathrm{init}}(\bm{u})_0$ of the finite-length filter of one-dimensional ($D=1$) target dynamical systems $g\in\mathcal{G}$.
According to this situation, consider a concatenated RNN reservoir system with $D=1$ and the following hypothesis sets.
We measure pseudo-dimensions of the hypothesis sets.
Recall that $\bm{F}_{N,M,\Gamma}$ is the concatenated FNN, that $\bm{x}_{N,M,\Gamma}:=(x_\mathrm{init},\ldots,x_\mathrm{init})\in\mathbb{R}^{N_{N,M,\Gamma}}$ is the initial state of $\bm{F}_{N,M,\Gamma}$ and that $\bar{\mathcal{F}}_{N,M,\Gamma}^D$ is a set of covering FNNs, whose elements are concatenated to make $\bm{F}_{N,M,\Gamma}$.
\begin{definition}\label{def:hypothesis_set}
We define the following hypothesis sets.
\begin{itemize}
 \item Set of reservoir functions $\mathcal{H}_\mathrm{all}:=\{h:\mathcal{U}\to\mathbb{R}\mid h(\bm{u}):=V_{\bm{F}_{N,M,\Gamma},\bm{x}_{N,M,\Gamma}}^W(\bm{u})_0,W\in\mathbb{R}^{N_{N,M,\Gamma}}\}$, where the readout $W$ can take arbitrary values. 
 \item Set of restricted reservoir functions $\mathcal{H}_\mathrm{one}:=\{h:\mathcal{U}\to\mathbb{R}\mid h(\bm{u}):=V_{\bm{F}_{N,M,\Gamma},\bm{x}_{N,M,\Gamma}}^{W_i}(\bm{u})_0,i=1,\ldots,N_{N,M,\Gamma}\}$, where $W_i\in\mathbb{R}^{N_{N,M,\Gamma}}$ is a one-hot vector whose $i$th component is 1.
 It is equivalent to 
 $\{h:\mathcal{U}\to\mathbb{R}\mid h(\bm{u}):=V_{\bar{f}_{N,M,\Gamma},x_\mathrm{init}}(\bm{u})_0,\bar{f}_{N,M,\Gamma}\in\bar{\mathcal{F}}_{N,M,\Gamma}^D\}\}$.
\end{itemize}
\end{definition}

Given a hypothesis set $\mathcal{H}$,
its pseudo-dimension $\Pdim(\mathcal{H})$ is defined as follows
 (see, e.g., \cite{Mohri_et_al:2018:fml}).
\begin{definition}[shatter]
Let $U_K:=(\bm{u}_1,\ldots,\bm{u}_K)\in\mathcal{U}^K$.
We say that a hypothesis set $\mathcal{H}$ shatters $U_K$ if and only if there exists $\bm{t}_K=(t_1,\ldots,t_K)\in\mathbb{R}^K$ such that $\{(I_\pm(H(\bm{u}_1)-t_1),\ldots,I_\pm(H(\bm{u}_K)-t_K))\mid h\in\mathcal{H}\}=2^K$, where $I_\pm(a)=1$ if $a\ge0$ and $0$ if $a<0$.
\end{definition}
\begin{definition}[pseudo-dimension]
The pseudo-dimension $\Pdim(\mathcal{H})$ of a hypothesis set $\mathcal{H}$ is defined as the maximum number of inputs that $\mathcal{H}$ can shatter.
\end{definition}

The pseudo-dimension bounds of $\mathcal{H}_\mathrm{all}$ and $\mathcal{H}_\mathrm{one}$ are as follows.
\begin{proposition}\label{pp:pdim_pl_rnns}
It holds that
\begin{align}
 \Pdim(\mathcal{H}_\mathrm{all})&\le N_{N,M,\Gamma},
 \\
 \Pdim(\mathcal{H}_\mathrm{one})&\le\ceil{\log_2N_{N,M,\Gamma}}.
\end{align}
\end{proposition}

From this result, we can state that the actual pseudo-dimension $\Pdim(\mathcal{H}_\mathrm{one})$ can be much smaller than the pseudo-dimension $\Pdim(\mathcal{H}_\mathrm{all})$.

\subsection{Significance of cascades}\label{ssec:lim_ex_approach}
In Section~\ref{sec:usu_lf}, 
we have used cascades to prove the uniform strong universality for the case of left-infinite inputs.
The motivation is to guarantee the ESP of RNN reservoir systems.
Indeed, as the following proposition suggests, the internal approximation cannot guarantee the ESP of RNN reservoir systems, when the FNN approximation theorem is in terms of uniform norm.
\begin{proposition}[denseness of non-ESP dynamical systems]\label{prop:esp_dense_in_ci}
 Let $\mathcal{C}^0(\bar{\mathcal{B}}_{S,I},\mathbb{R}^D):=\mathcal{C}(\bar{\mathcal{B}}_{S,I},\mathbb{R}^D)$.
 Fix $i\in\mathbb{N}_0$.
 Let $\mathcal{G}_\mathrm{esp}\subset\mathcal{C}^i(\bar{\mathcal{B}}_{S,I},\bar{\mathcal{B}}_{S})$ be a set of $C^i$ state maps \eqref{eq:ds} with ESP.
 Let $\mathcal{G}_\mathrm{nesp}:=\mathcal{C}^i(\bar{\mathcal{B}}_{S,I},\mathbb{R}^D)\backslash\mathcal{G}_\mathrm{esp}$ be a set of $C^i$ state maps without ESP.
 Then, $\mathcal{G}_\mathrm{nesp}$ is dense in $(\mathcal{C}^i(\bar{\mathcal{B}}_{S,I},\mathbb{R}^D),\|\cdot\|_{\bar{\mathcal{B}}_{S,I},p})$.
\end{proposition}

The above result suggests that there are non-ESP dynamical systems arbitrarily close to target dynamical systems. 
Therefore, internal approximation in uniform norm, which only guarantees that approximating FNNs are close to the target dynamical systems, is insufficient for guaranteeing the ESP of approximating RNN reservoir systems.
Note that the proposition states the denseness of non-ESP systems in the topology-theoretic sense but not their density in the measure-theretic sense, 
so that we do not know about how likely it is to obtain a non-ESP approximating RNN reservoir system when it achieves a certain level of approximation error.

\section{Conclusion}\label{sec:conclusion}
In this paper, we have discussed the approximation capability of RNN reservoir systems in the uniform strong setting against a class of contracting dynamical systems characterized in terms of the Barron class.
For proving the uniform strong universality, we have introduced the approximation bounds of an FNN for the two types of activation functions stated in the existing researches \citep{Caragea_et_al:2023:barron_unif_improved,Barron:1992:barron_unif,Sreekumar_et_al:2021:barron_unif_sig1,Sreekumar_and_Goldfeld:2022:barron_unif_sig2} with minor revisions or with an alternative proof (Propositions~\ref{pp:prop_2_2} and \ref{pp:prop_2_2_sig}).
We have shown the uniform strong universality of the family of RNN reservoir systems for finite-length inputs (Theorem~\ref{th:usu_fnn}) in a constructive manner on the basis of the parallel concatenation of FNN reservoir maps.
For left-infinite inputs, we have shown the universality by constructing a cascade of the concatenated FNN reservoir map (Theorem~\ref{th:usu_fnn_lf}).

The results in this paper have several weaknesses.
Firstly, one may argue that the assumptions on the target dynamical systems are rather restrictive, as mentioned in Remarks~\ref{rm:lim_dom_prsv} and \ref{rm:comp_assumptions}. 
Secondly, at least from the results in this paper, concatenated RNN reservoir systems or cascaded RNN reservoir systems can be infeasibly large: See the discussion in Section~\ref{ssec:scale_fnn}.
Thirdly, cascaded RNN reservoir systems are less efficient in both scale and error compared with those in existing research \citep{Cuchiero_et_al:2021:strong_univ}.

\section*{Acknowledgements}
We would like to thank Professor Hiroshi Kokubu, Kyoto University,
Japan, for his drawing our attention to the study on
universality of reservoir systems.
We would also like to thank Professors Andrei Caragea
and Felix Voigtlaender, Katholische Universit\"{a}t
Eichst\"{a}tt-Ingolstadt, Germany, for having discussion with us 
on Proposition 2.2 of~\cite{Caragea_et_al:2023:barron_unif_improved}.

\appendix
\section{Proofs of and remarks on approximation upper bounds of feedforward neural networks}
\subsection{Proof of Proposition~\ref{pp:prop_2_2}}\label{aapp:proof_prop_2_2}
The proposition can basically be proved by following the proof of Proposition~2.2 in \cite{Caragea_et_al:2023:barron_unif_improved}, with the following three modifications:
Firstly, we simplify the setting of Proposition~2.2 in \cite{Caragea_et_al:2023:barron_unif_improved} by assuming that the domain of functions includes $\bm{0}$ and by setting what is called a base point to $\bm{0}$.
Note that the setting of the base point does not change the result substantially as remarked in \cite{Caragea_et_al:2023:barron_unif_improved}.
Secondly, we revise the upper bounds of parameters.
Finally, we replace an erroneous equality used in the proof (A.\ Caragea, personal communication, 2024).
This is an integral representation of a function $g$ of the Barron class, which is originating from \cite{Barron:1992:barron_unif} and which is based on an inexact representation of the term $e^{iz}-1$ in the derivation.
We would like to mention that \cite{Cheang:1998:nn_approx} also provides the same corrected integral representation of $g$ as ours, via revising the representation of $e^{iz}-1$.
In addition to using this representation, we furthermore simplify this representation of $g$ so that it has only one step function inside the integral, where the original representation has two step functions.
 We will summarize the revised integral representation of a function $g$ as Lemma~\ref{lem:exp_int_rep} in the following. 
\begin{lemma}\label{lem:exp_int_rep}
 For a non-empty bounded subset $\mathcal{B}$ of $\mathbb{R}^Q$ that includes $\bm{0}$ and for $g\in\mathcal{Z}_{M,\mathcal{B}}$, it holds that
 \begin{align}
  g(\bm{x})-g(\bm{0})&=-2v\int_{\mathbb{R}^Q}\int_0^11_{\bm{\omega}^\top\bm{x}/\norm{\bm{\omega}}_\mathcal{B}-t>0}s(\bm{\omega},t)p(\bm{\omega},t)dtd\bm{\omega},\quad\forall\bm{x}\in\mathcal{B},
 \end{align}
 where 
 \begin{align}
  s(\bm{\omega},t)&:=\mathrm{sgn}(\sin(\norm{\bm{\omega}}_\mathcal{B}t+\theta(\bm{\omega}))),
  \\
  p(\bm{\omega},t)&:=\frac{1}{v}\abs{\sin(\norm{\bm{\omega}}_\mathcal{B}t+\theta(\bm{\omega}))}\norm{\bm{\omega}}_\mathcal{B}\abs{\tilde{g}(\bm{\omega})},
  \\
  v&=\int_{\mathbb{R}^Q}\int_0^1\abs{\sin(\norm{\bm{\omega}}_\mathcal{B}t+\theta(\bm{\omega}))}\norm{\bm{\omega}}_\mathcal{B}\abs{\tilde{g}(\bm{\omega})}dtd\bm{\omega},
\end{align}
where $\theta(\bm{\omega}):=\mathop{\mathrm{arg}}\tilde{g}(\bm{\omega})$.
\end{lemma}
\begin{proof}
  Since $g$ is of the Barron class, it holds that
  \begin{equation}
    \label{eq:Barron_eq}
   g(\bm{x})-g(\bm{0})=\int_{\mathbb{R}^Q}(e^{i\bm{\omega}^\top\bm{x}}-1)\tilde{g}(\bm{\omega})d\bm{\omega},\quad\forall\bm{x}\in\mathcal{B}.
  \end{equation}
  Note that $\tilde{g}(-\bm{\omega})=\overline{\tilde{g}(\bm{\omega})}$ because the left-hand side of the above equation is real-valued.
  We can decompose the integral on the right-hand side into two parts, as
  \begin{align}
   g(\bm{x})-g(\bm{0})=\int_{\mathcal{R}_+(\bm{x})}(e^{i\bm{\omega}^\top\bm{x}}-1)\tilde{g}(\bm{\omega})d\bm{\omega}+\int_{\mathcal{R}_-(\bm{x})}(e^{i\bm{\omega}^\top\bm{x}}-1)\tilde{g}(\bm{\omega})d\bm{\omega},
  \end{align}
  where $\mathcal{R}_\pm(\bm{x}):=\{\bm{\omega}\in\mathbb{R}^Q:\bm{\omega}^\top\bm{x}\gtrless0\}$.

  Recalling the definition $\|\bm{\omega}\|_{\mathcal{B}}
  :=\sup_{\bm{x}\in\mathcal{B}}|\bm{\omega}^\top\bm{x}|$,
  one has $|\bm{\omega}^\top\bm{x}|\le\|\bm{\omega}\|_{\mathcal{B}}=:Y$. 
  As stated in \cite{Cheang:1998:nn_approx}, $e^{iz}-1$ can be represented as follows.
  When $0\le z\le Y$, it holds that 
  \begin{align}
   e^{iz}-1=i\int_0^Y1_{z>u}e^{iu}du.
  \end{align}
  On the other hand, when $-Y\le z\le0$, it holds\footnote{%
  The corresponding equation in~\cite{Barron:1992:barron_unif} reads
  \begin{equation*}
   e^{iz}-1=-i\int_0^Y1_{z<-u}e^{iu}du.
  \end{equation*}
  Note the difference in the sign of the exponent.} that
  \begin{align}
   e^{iz}-1=-i\int_0^Y1_{z<-u}e^{-iu}du.\label{eq:int_rep_corrected}
  \end{align}
  By setting $z=\bm{\omega}^\top\bm{x}$ and $Y=\norm{\bm{\omega}}_\mathcal{B}$, one consequently has
  \begin{align}
   g(\bm{x})-g(\bm{0})&=i\int_{\mathcal{R}_+(\bm{x})}\int_0^{\norm{\bm{\omega}}_\mathcal{B}}1_{\bm{\omega}^\top\bm{x}>u}e^{iu}du\tilde{g}(\bm{\omega})d\bm{\omega}\nonumber\\
   &-i\int_{\mathcal{R}_-(\bm{x})}\int_0^{\norm{\bm{\omega}}_\mathcal{B}}1_{\bm{\omega}^\top\bm{x}<-u}e^{-iu}du\tilde{g}(\bm{\omega})d\bm{\omega}
   \\
   &=i\int_{\mathcal{R}_+(\bm{x})}\int_0^{\norm{\bm{\omega}}_\mathcal{B}}1_{\bm{\omega}^\top\bm{x}>u}e^{iu}du\tilde{g}(\bm{\omega})d\bm{\omega}\nonumber\\
   &-i\int_{\mathcal{R}_+(\bm{x})}\int_0^{\norm{\bm{\omega}}_\mathcal{B}}1_{\bm{\omega}^\top\bm{x}>u}e^{-iu}du\tilde{g}(-\bm{\omega})d\bm{\omega}\label{eq:int_minus_conj_int}
   \\
   &=-2\mathrm{Im}\int_{\mathcal{R}_+(\bm{x})}\int_0^{\norm{\bm{\omega}}_\mathcal{B}}1_{\bm{\omega}^\top\bm{x}>u}e^{iu}du\tilde{g}(\bm{\omega})d\bm{\omega}
   \\
   &=-2\int_{\mathcal{R}_+(\bm{x})}\int_0^{\norm{\bm{\omega}}_\mathcal{B}}1_{\bm{\omega}^\top\bm{x}>u}\sin(u+\theta(\bm{\omega}))du\abs{\tilde{g}(\bm{\omega})}d\bm{\omega}
   \\
   &=-2\int_{\mathbb{R}^Q}\int_0^{\norm{\bm{\omega}}_\mathcal{B}}1_{\bm{\omega}^\top\bm{x}>u}\sin(u+\theta(\bm{\omega}))du\abs{\tilde{g}(\bm{\omega})}d\bm{\omega}.
  \end{align}
  In the third equality, we used the fact that the second term in \eqref{eq:int_minus_conj_int} is the complex conjugate of the first term in \eqref{eq:int_minus_conj_int}.
  The last equality holds because of the indicator function $1_{\bm{\omega}^\top\bm{x}>u}$.
  With the change of variables $u=\norm{\bm{\omega}}_\mathcal{B}t$, the lemma holds.
\end{proof}

Replacing the integral representations and following the original proof
of Proposition 2.2 in~\cite{Caragea_et_al:2023:barron_unif_improved} prove the proposition.
\qed

\subsection{Technical remarks on Proposition~\ref{pp:prop_2_2}}\label{aapp:remark_prop_2_2}
The number $4N$ of hidden nodes of the FNN $f_\rho$ approximating $g$ is decreased by half from $8N$ in the original proposition \citep[Proposition 2.2]{Caragea_et_al:2023:barron_unif_improved}. 
This difference arises from correcting the erroneous integral representation of a function $g$ of the Barron class in Lemma~\ref{lem:exp_int_rep}:
Originally, a function $g$ of the Barron class is represented in terms of integration of two step functions. 
Each step function is then approximated with two ReLU functions, and the total number of ReLU functions needed is related to the number of hidden nodes of the FNN.
However, since a function $g$ of the Barron class is in fact represented as integration of one step function, the number of needed ReLU functions is decreased by half.
Therefore, the number of hidden nodes is also decreased by half.

When we apply this proposition to our problem setting, some parameters of FNNs can be bounded in terms of the norms $\norm{\cdot}_{\bar{\mathcal{B}}_S}$ and $\norm{\cdot}_{\bar{\mathcal{B}}_I}$, which can be regarded as $q$-norms
since $\bar{\mathcal{B}}_S$ and $\bar{\mathcal{B}}_I$ are $p$-norm balls (see Section~V in \cite{Barron:1993:nn_univ}).
Since $q$-norms appear naturally in \eqref{eq:holder_applied_q_norm_appeared} when deriving the upper bound of Lemma~\ref{lem:diff_approx_funcs_ub_wrt_params} as a result of the H\"{o}lder inequality, bounds and coverings in terms of $q$-norms are consistent with our discussion.

\subsection{Proof of Proposition~\ref{pp:prop_2_2_sig}}\label{aapp:proof_prop_2_2_sig}
We prepare a lemma that is needed in the proof.
See, e.g., \cite{Mohri_et_al:2018:fml} for the definitions of shattering and of VC-dimension.
\begin{lemma}\label{lem:vc_dim_of_sig_nn}
Let $\mathcal{B}\subset\mathbb{R}^Q$ and $\sigma:\mathbb{R}\to[0,1]$ be defined as in Proposition~\ref{pp:prop_2_2_sig}.
Let $\Lambda\in(0,\infty)$.
Let $N_{\Lambda,\bm{x}}:(\mathbb{R}^Q\backslash\{\bm{0}\})\times\mathbb{R}\to\mathbb{R}$, $N_{\Lambda,\bm{x}}(\bm{\omega},t):=\sigma(\Lambda(\bm{\omega}^\top\bm{x}/\norm{\bm{\omega}}_\mathcal{B}-t))$.
Then, $\sup_{\lambda\in\mathbb{R}}\VC(\{1_{N_{\Lambda,\bm{x}}-\lambda>0}\mid\bm{x}\in\mathcal{B}\})\le Q+1\le2Q$, where $\VC(\mathcal{H})$ denotes the VC-dimension of a set $\mathcal{H}$ of classifiers.
\end{lemma}
\begin{proof}
 We consider the following functions.
 \begin{align}
 f_{\Lambda,\bm{x},\lambda}(\bm{\omega}^*,t)&:=\sigma(\Lambda(\bm{x}^\top\bm{\omega}^*-t))-\lambda,
 \\
 f_{\Lambda,\bm{x}}(\bm{\omega}^*,t,\eta)&:=\sigma(\Lambda(\bm{x}^\top\bm{\omega}^*-t))-\eta,
 \\
 g_{\Lambda,\bm{x}}(\bm{\omega}^*,t)&:=\sigma(\Lambda(\bm{x}^\top\bm{\omega}^*-t)),
 \\
 h_{\Lambda,\bm{x}}(\bm{\omega}^*,t)&:=\Lambda(\bm{x}^\top\bm{\omega}^*-t),
 \\
 l_{\bm{x},\xi}(\bm{\omega}^*,t)&:=\bm{x}^\top\bm{\omega}^*+\xi t.
 \end{align}
 Note that $f_{\Lambda,\bm{x}}$ is a function of $(\bm{\omega}^*,t,\eta)\in\mathbb{R}^Q\times\mathbb{R}\times\mathbb{R}$ while the others are functions of $(\bm{\omega}^*,t)\in\mathbb{R}^Q\times\mathbb{R}$.
 It holds that
 \begin{align}
  &\VC(\{1_{N_{\Lambda,\bm{x}}-\lambda>0}\mid\bm{x}\in\mathcal{B}\})
  \\
  &\le\VC(\{1_{f_{\Lambda,\bm{x},\lambda}>0}\mid\bm{x}\in\mathcal{B}\})
  \\
  &\le\VC(\{1_{f_{\Lambda,\bm{x}}>0}\mid\bm{x}\in\mathcal{B}\})
  \\
  &=\Pdim(\{g_{\Lambda,\bm{x}}\mid\bm{x}\in\mathcal{B}\})
  \\
  &\le\Pdim(\{h_{\Lambda,\bm{x}}\mid\bm{x}\in\mathcal{B}\})
  \\
  &\le\Pdim(\{l_{\bm{x},\xi}\mid(\bm{x},\xi)\in\mathbb{R}^Q\times\mathbb{R}\})
  \\
  &=\dim(\{l_{\bm{x},\xi}\mid(\bm{x},\xi)\in\mathbb{R}^Q\times\mathbb{R}\})=Q+1.
 \end{align}
 The first inequality holds because the composition with the non-linear transformation $(\bm{\omega},t)\to(\bm{\bm{\omega}}/\norm{\bm{\omega}}_\mathcal{B},t)$ does not increase the VC-dimension of a model.
 The second inequality holds because, if $\{1_{f_{\Lambda,\bm{x},\lambda}>0}\mid\bm{x}\in\mathcal{B}\}$ shatters $\{(\bm{\omega}_i^*,t_i)\}_{i=1}^N$, then $\{1_{f_{\Lambda,\bm{x}}>0}\mid\bm{x}\in\mathcal{B}\}$ shatters $\{(\bm{\omega}_i^*,t_i,\lambda)\}_{i=1}^N$.
The equality in the next line holds trivially from the definitions of VC-dimension and pseudo-dimension as it is explained in equation (11.3) of \cite{Mohri_et_al:2018:fml}.
For proving the third inequality, Theorem~11.3 in \cite{Anthony_and_Bartlett:1999:nnl} is used, which states that composition with non-decreasing function cannot increase pseudo-dimension.
For proving the last line, Theorem~11.7 in \cite{Mohri_et_al:2018:fml} is used.
By taking the supremum in terms of $\lambda$, the desired inequality holds.
\end{proof}

 We prove Proposition~\ref{pp:prop_2_2_sig}.
 We follow the argument of the proof of Proposition~2.2 in \cite{Caragea_et_al:2023:barron_unif_improved} in most part of our proof.
 
 We consider the case of $M=1$ first, and then generalize the result later.
 Furthermore, we assume that $e=g(\bm{0})$.
 Recall that $\abs{g(\bm{0})}\le1$ because $g\in\mathcal{Z}_{1,\mathcal{B}}$.
 Let $g_0(\bm{x}):=g(\bm{x})-g(\bm{0})$, $\forall\bm{x}\in\mathcal{B}$.
 We find the remaining part $f_{\sigma,0}(\bm{x}):=\sum_{i=1}^{2N}a_i\sigma(\bm{b}_i^\top\bm{x}+c_i)$ of the FNN $f_\sigma$ such that $\norm{g_0-f_{\sigma,0}}_{\mathcal{B},\infty}=\norm{g-f_\sigma}_{\mathcal{B},\infty}$ is bounded from above by \eqref{eq:ub_desired_sig} for the case of $M=1$.

In the following, we decompose $g_0$ into two functions $g_+$ and $g_-$.
Then, we approximate $g_+$ and $g_-$ with FNNs $f_{\sigma,+}$ and $f_{\sigma,-}$, respectively.
Finally, we concatenate the two FNNs to construct $f_{\sigma,0}$.

We decompose $g_0$.
Assume the integral representation of $g_0$ shown in Lemma~\ref{lem:exp_int_rep}.
We decompose the integral into two integrals by splitting its domain
according to the sign $s(\bm{\omega},t)$
of $\sin(\norm{\bm{\omega}}_{\mathcal{B}}t+\theta(\bm{\omega}))$.
Let $\Omega:(\mathbb{R}^Q\backslash\{\bm{0}\})\times[0,1]$.
Let 
\begin{align}
 \Gamma_{\bm{x}}(\bm{\omega},t)&:=1_{\bm{\omega}^\top\bm{x}/\norm{\bm{\omega}}_\mathcal{B}-t>0},\ \Gamma_{\bm{x}}:\Omega\to[0,1],
 \\
 V_\pm&:=\int_{\mathbb{R}^Q}\int_0^11_{s(\bm{\omega},t)=\pm1}p(\bm{\omega},t)dtd\bm{\omega},
 \\
 d\mu_\pm&:=\frac{1}{V_\pm}1_{s(\bm{\omega},t)=\pm1}p(\bm{\omega},t)dtd\bm{\omega},
 \\
 g_\pm(\bm{x})&=\int_\Omega\Gamma_{\bm{x}}(\bm{\omega},t)d\mu_\pm(\bm{\omega},t),\ g_\pm:\mathcal{B}\to\mathbb{R}.
\end{align}
Note that $V_\pm\ge0$, $V_++V_-=1$ and $v\le M=1$\footnote{We skip the trivial case $v=0$. Also, if either $V_+$ or $V_-$ is zero, then the following argument holds by dropping $g_+$ or $g_-$, respectively. Finally, for any $\bm{\omega}\in\mathbb{R}^Q\backslash\{\bm{0}\}$, it holds that $\norm{\bm{\omega}}_\mathcal{B}>0$ when $\mathcal{B}$ has non-empty interior \citep{Caragea_et_al:2023:barron_unif_improved}.\label{fn:cases}}.
Then, $g_0$ can be decomposed as $g_0=-2v(V_+g_+-V_-g_-)$.

We approximate $g_+$ and $g_-$ with FNNs $f_{\sigma,+}$ and $f_{\sigma,-}$, respectively.
The construction of the FNNs takes two steps.
Firstly, the step function $\Gamma_{\bm{x}}$ in $g_\pm$ is approximated by $N_{\Lambda,\bm{x}}(\bm{\omega},t):=\sigma(\Lambda(\bm{\omega}^\top\bm{x}/\norm{\bm{\omega}}_\mathcal{B}-t))$.
Let $g_{\pm,\Lambda}(\bm{x}):=\int_\Omega N_{\Lambda,\bm{x}}(\bm{\omega},t)d\mu_\pm(\bm{\omega},t)$.
Then, the error of approximating $g_\pm$ with $g_{\pm,\Lambda}$ is
bounded from above as
\begin{align}
 &\abs{g_\pm(\bm{x})-g_{\pm,\Lambda}(\bm{x})}
 \\
 &=\abs{\int_\Omega(\Gamma_{\bm{x}}(\bm{\omega},t)-N_{\Lambda,\bm{x}}(\bm{\omega},t))d\mu_\pm(\bm{\omega},t)}
 \\
 &\le\int_{\mathbb{R}^Q\backslash\{\bm{0}\}}\int_0^1\abs{\Gamma_{\bm{x}}(\bm{\omega},t)-N_{\Lambda,\bm{x}}(\bm{\omega},t)
 }\frac{1}{V_\pm}1_{s(\bm{\omega},t)=\pm1}p(\bm{\omega},t)dtd\bm{\omega}
 \\
 &\le\int_{\mathbb{R}^Q\backslash\{\bm{0}\}}\int_0^1\abs{\Gamma_{\bm{x}}(\bm{\omega},t)-N_{\Lambda,\bm{x}}(\bm{\omega},t)}\frac{1}{vV_\pm}\norm{\bm{\omega}}_\mathcal{B}\abs{\tilde{g}(\bm{\omega})}dtd\bm{\omega}
 \\
 &=\int_{\mathbb{R}^Q\backslash\{\bm{0}\}}\int_0^1\abs{1_{\bm{\omega}^\top\bm{x}/\norm{\bm{\omega}}_\mathcal{B}-t>0}-\sigma(\Lambda(\bm{\omega}^\top\bm{x}/\norm{\bm{\omega}}_\mathcal{B}-t))}\frac{1}{vV_\pm}\norm{\bm{\omega}}_\mathcal{B}\abs{\tilde{g}(\bm{\omega})}dtd\bm{\omega}
 \\
 &\le\int_{\mathbb{R}^Q\backslash\{\bm{0}\}}\delta(\Lambda)\frac{1}{vV_\pm}\norm{\bm{\omega}}_\mathcal{B}\abs{\tilde{g}(\bm{\omega})}d\bm{\omega}
 \\
 &\le\frac{M\delta(\Lambda)}{vV_\pm}=\frac{\delta(\Lambda)}{vV_\pm},
\end{align}
where we used the fact that approximation error between $\Gamma_{\bm{x}}(\bm{\omega},\cdot)$ and $N_{\Lambda,\bm{x}}(\bm{\omega},\cdot)$ on an interval of length $1$ is less than $\delta(\Lambda)$ because of the monotonicity of $\sigma$.

Secondly, we approximate $g_{\pm,\Lambda}$ by an FNN $f_{\sigma,\pm}$ with $N$ hidden nodes.
By Lemma~\ref{lem:vc_dim_of_sig_nn} and Proposition~A.1 of \cite{Caragea_et_al:2023:barron_unif_improved}, it holds that
\begin{equation}
 \mathbb{E}_{(\bm{\omega}_1,t_1),\ldots,(\bm{\omega}_N,t_N)\overset{\mathrm{i.i.d.}}{\sim}\mu_\pm}\left[\sup_{\bm{x}\in\mathcal{B}}\abs{g_{\pm,\Lambda}(\bm{x})-\frac{1}{N}\sum_{i=1}^NN_{\Lambda,\bm{x}}(\bm{\omega}_i,t_i)}\right]\le\kappa_0\sqrt{2Q/N},\label{eq:ex_sup}
\end{equation}
where $\{(\bm{\omega}_i,t_i)\}_{i=1,\ldots,N}$ are independent and identically-distributed (i.i.d.) samples and $\kappa_0>0$ is a universal constant.
Therefore, there exists a certain realization $\{(\bm{\omega}_{\pm,i},t_{\pm,i})\}_{i=1,\ldots,N}$ such that\footnote{As stated in \cite{Caragea_et_al:2023:barron_unif_improved}, Proposition~A.1 in \cite{Caragea_et_al:2023:barron_unif_improved} gives an upper bound of 
\begin{align}
\sup_{\mathcal{B}_\mathrm{fin}\subset\mathcal{B}}\mathbb{E}_{(\bm{\omega}_1,t_1),\ldots,(\bm{\omega}_N,t_N)\overset{\mathrm{i.i.d.}}{\sim}\mu_\pm}\left[\sup_{\bm{x}\in\mathcal{B}_\mathrm{fin}}\abs{g_{\pm,\Lambda}(\bm{x})-\frac{1}{N}\sum_{i=1}^NN_{\Lambda,\bm{x}}(\bm{\omega}_i,t_i)}\right],\label{eq:ex_sup_alt}
\end{align}
where $\mathcal{B}_\mathrm{fin}$ is a finite subset of $\mathcal{B}$, for avoiding the measurability issue.
However, as stated in the second footnote in \cite{Caragea_et_al:2023:barron_unif_improved}, \eqref{eq:ex_sup_alt} is equivalent to the left-hand side of \eqref{eq:ex_sup} when $g_{\pm,\Lambda}(\bm{x})$ and $\frac{1}{N}\sum_{i=1}^NN_{\Lambda,\bm{x}}(\bm{\omega}_i,t_i)$ are continuous with respect to $\bm{x}$, which is guaranteed by the continuity of $\sigma$ in our proof.}
\begin{align}
 \sup_{\bm{x}\in\mathcal{B}}\abs{g_{\pm,\Lambda}(\bm{x})-\frac{1}{N}\sum_{i=1}^NN_{\Lambda,\bm{x}}(\bm{\omega}_{\pm,i},t_{\pm,i})}\le\kappa_0'\sqrt{Q}N^{-1/2}\quad(\kappa_0':=\kappa_0\sqrt{2}).
\end{align}
The sum of sigmoids $f_{\sigma,\pm}(\bm{x}):=\frac{1}{N}\sum_{i=1}^NN_{\Lambda,\bm{x}}(\bm{\omega}_{\pm,i},t_{\pm,i})=\frac{1}{N}\sum_{i=1}^N\sigma(\Lambda\bm{\omega}_{\pm,i}^\top\bm{x}/\norm{\bm{\omega}_{\pm,i}}_\mathcal{B}-\Lambda t_{\pm,i})$ defines an FNN, with $N$ hidden nodes and with the parameters bounded as $\norm{\Lambda\bm{\omega}_{\pm,i}/\norm{\bm{\omega}_{\pm,i}}_\mathcal{B}}_\mathcal{B}\le\Lambda$, $\abs{\Lambda t_{\pm,i}}\le\Lambda$, $i=1,\ldots,N$.

By the triangle inequality, one has 
\begin{align}
 \norm{g_\pm-f_{\sigma,\pm}}_{\mathcal{B},\infty}\le\frac{\delta(\Lambda)}{vV_\pm}+\kappa_0'\sqrt{Q}N^{-1/2}.
\end{align}
One can easily check that an FNN $f_{\sigma,1}:=-2v(V_+f_{\sigma,+}-V_-f_{\sigma,-})+e$ satisfies \eqref{eq:ub_desired_sig} with $M=1$, and that its parameters satisfy \eqref{eq:coeff_ub_sig} with $M=1$.

Suppose that $M>0$, and $g\in\mathcal{Z}_{M,\mathcal{B}}$.
Then, as discussed above, there exists an FNN $f_{\sigma,1}$ that satisfies \eqref{eq:ub_desired_sig} and \eqref{eq:coeff_ub_sig} for the case of $M=1$ for $g_1:=g/M\in\mathcal{Z}_{1,\mathcal{B}}$.
Let $f_\sigma$ be an FNN constructed by multiplying $a_i$, $i=1,\ldots,2N$ and $e$ of $f_{\sigma,1}$ by $M$.
Then, $f_\sigma$ satisfies \eqref{eq:ub_desired_sig} and \eqref{eq:coeff_ub_sig} for general $M>0$.
\qed

\subsection{Remarks on Proposition~\ref{pp:prop_2_2_sig}}\label{aapp:remark_prop_2_2_sig}
Proposition~\ref{pp:prop_2_2_sig} originates from \cite{Barron:1992:barron_unif}.
One of the results in \cite{Barron:1992:barron_unif} suggests that functions in the Barron class can be approximated in terms of uniform norm by FNNs with a sigmoidal activation function.
However, \cite{Caragea_et_al:2023:barron_unif_improved} suggested that his proof has a gap, and then provided an independent proof of the result in the case of the ReLU activation function as Proposition~2.2 in their paper.
Also, \cite{Sreekumar_et_al:2021:barron_unif_sig1,Sreekumar_and_Goldfeld:2022:barron_unif_sig2} gave refinement of the result of \cite{Barron:1992:barron_unif} in the case of sigmoid activation function, which includes explicit bounds of parameters of FNNs.

On the other hand, we have complemented the result of \cite{Barron:1992:barron_unif} on the basis of the Proposition~2.2 of \cite{Caragea_et_al:2023:barron_unif_improved}.
Compared with the result in \cite{Sreekumar_et_al:2021:barron_unif_sig1,Sreekumar_and_Goldfeld:2022:barron_unif_sig2}, our result improves the rates of the upper bounds of the magnitudes of $\bm{b}_i$ and $c_i$ when activation function is the logistic sigmoid function.
Indeed, when the number of hidden nodes is $2N$, by choosing a parameter $\Lambda=N^{1/2}$ for Proposition~\ref{pp:prop_2_2_sig}, which we have discussed in Section~\ref{ssec:sig_func}, our upper bound of the magnitudes of $\bm{b}_i$ and $c_i$ are $O(N^{1/2})$, while that of the existing work is $O(N^{1/2}\log N)$.

Approximation bounds of FNNs have been extensively studied, in \cite{Barron:1992:barron_unif,Klusowski_and_Barron:2018:approx_relu,Sreekumar_et_al:2021:barron_unif_sig1,Sreekumar_and_Goldfeld:2022:barron_unif_sig2,Belomestny:2023:nn_approx_holder,Jonathan:2023:nn_approx_reluk,Caragea_et_al:2023:barron_unif_improved} to mention just a few.
One may prove approximation bounds in uniform strong setting with FNN approximation bounds other than those which we used in this paper,
provided that the approximation theorem to be used satisfies boundedness of the parameters of FNNs, Lipschitz continuity of activation functions for constructing covering, and an upper bound of approximation error in terms of uniform norm that is independent of each function in a specified function class.

See, e.g., \cite{Caragea_et_al:2023:barron_unif_improved} for review of other function classes, which are also called Barron spaces, and relations between them.

Compared with Proposition~\ref{pp:prop_2_2}, Proposition~\ref{pp:prop_2_2_sig} requires the smaller number $2N$ of hidden nodes while it has the extra term $\delta(\Lambda)$ in the approximation bound.
These differences arise because we approximate the step function in the integral representation of a function $g$ of the Barron class directly with a single sigmoid function in Proposition~\ref{pp:prop_2_2_sig} instead of a pair of ReLU functions in Proposition~\ref{pp:prop_2_2}.

\section{Proofs of Lemmas for proving universality}
\subsection{Proof of Lemma~\ref{lem:filter_nn_diff_ub}}\label{app:pr_lem_filter_nn_diff_ub}
First of all, we prove that $\bm{f}$ also becomes domain-preserving.
We mimick a part of proof of Theorem~4.1 of \cite{Grigoryeva_and_Ortega:2018:esn_univ}.
One has
\begin{align}
  \norm{\bm{f}}_{\bar{\mathcal{B}}_{S,I},p}&\le\abs{\norm{\bm{f}}_{\bar{\mathcal{B}}_{S,I},p}-\norm{\bm{g}}_{\bar{\mathcal{B}}_{S,I},p}}+\norm{\bm{g}}_{\bar{\mathcal{B}}_{S,I},p}
  \nonumber\\
  &\le\norm{\bm{f}-\bm{g}}_{\bar{\mathcal{B}}_{S,I},p}+\norm{\bm{g}}_{\bar{\mathcal{B}}_{S,I},p}
  \nonumber\\
  &\le S-P+P=S,
\end{align}
where the second inequality is due to the (reverse) triangle inequality and where the last inequality is due to the assumption
of the lemma and Assumption A\ref*{asm:ds_asm_fin}-\ref{enm:uniformly_bounded}. 
The above inequality shows that the range of $\bm{f}$ is limited in $\bar{\mathcal{B}}_S$.

Next, we prove the bound  $\norm{V_{\bm{g},\bm{x}_\mathrm{init}}-V_{\bm{f},\bm{x}_\mathrm{init}}}_{\mathcal{T}_T,p}\le\norm{\bm{g}-\bm{f}}_{\bar{\mathcal{B}}_{S,I},p}\sum_{i=0}^{T-1}L^i$.
The proof in the remaining part is essentially the same as the proof of Theorem~3.1~(iii) in \cite{Grigoryeva_and_Ortega:2018:esn_univ}.
Recall Definition~\ref{def:filter_norm} of the norm
of filters: 
\begin{align}
 &\norm{V_{\bm{f},\bm{x}_\mathrm{init}}-V_{\bm{g},\bm{x}_\mathrm{init}}}_{\mathcal{T}_T,p}\label{eq:norm_diff_f_and_cov}
 \\
&=\sup_{\bm{u}\in(\bar{\mathcal{B}}_I)^T}\sup_{t\in\mathcal{T}_T}\norm{V_{\bm{f},\bm{x}_\mathrm{init}}(\bm{u})_t-V_{\bm{g},\bm{x}_\mathrm{init}}(\bm{u})_t}_p.\label{eq:norm_diff_f_and_cov_ub}
\end{align}
We expand the $p$-norm iteratively to a sum of $p$-norms, 
and then bound each $p$-norm.
Consider the following expansion:
\begin{align}
&\norm{V_{\bm{f},\bm{x}_\mathrm{init}}(\bm{u})_t-V_{\bm{g},\bm{x}_\mathrm{init}}(\bm{u})_t}_p
\\
&=\norm{\bm{f}(V_{\bm{f},\bm{x}_\mathrm{init}}(\bm{u})_{t-1},\bm{u}(t))-\bm{g}(V_{\bm{g},\bm{x}_\mathrm{init}}(\bm{u})_{t-1},\bm{u}(t))}_p
\\
&\le\norm{\bm{f}(V_{\bm{f},\bm{x}_\mathrm{init}}(\bm{u})_{t-1},\bm{u}(t))-\bm{g}(V_{\bm{f},\bm{x}_\mathrm{init}}(\bm{u})_{t-1},\bm{u}(t))}_p\nonumber\\
&+\norm{\bm{g}(V_{\bm{f},\bm{x}_\mathrm{init}}(\bm{u})_{t-1},\bm{u}(t))-\bm{g}(V_{\bm{g},\bm{x}_\mathrm{init}}(\bm{u})_{t-1},\bm{u}(t))}_p
\\
&\le\norm{\bm{f}(V_{\bm{f},\bm{x}_\mathrm{init}}(\bm{u})_{t-1},\bm{u}(t))-\bm{g}(V_{\bm{f},\bm{x}_\mathrm{init}}(\bm{u})_{t-1},\bm{u}(t))}_p\nonumber\\
&+ L\norm{V_{\bm{f},\bm{x}_\mathrm{init}}(\bm{u})_{t-1}-V_{\bm{g},\bm{x}_\mathrm{init}}(\bm{u})_{t-1}}_p,\label{eq:before_repeating_ops}
\end{align}
where the first inequality is due to the triangle inequality,
and where the next inequality follows from the Lipschitz condition
in Assumption A\ref*{asm:ds_asm_fin}-\ref{enm:lipschitz_wrt_input}. 
By repeating this expansion, it holds that
\begin{align}
\eqref{eq:before_repeating_ops}&\le\norm{\bm{f}(V_{\bm{f},\bm{x}_\mathrm{init}}(\bm{u})_{t-1},\bm{u}(t))-\bm{g}(V_{\bm{f},\bm{x}_\mathrm{init}}(\bm{u})_{t-1},\bm{u}(t))}_p\nonumber
\\
&+L\norm{\bm{f}(V_{\bm{f},\bm{x}_\mathrm{init}}(\bm{u})_{t-2},\bm{u}(t-1))-\bm{g}(V_{\bm{f},\bm{x}_\mathrm{init}}(\bm{u})_{t-2},\bm{u}(t-1))}_p\nonumber
\\
&\vdots \nonumber
\\
&+L^{T+t-1}\norm{\bm{f}(\bm{x}_\mathrm{init},\bm{u}(-T+1))-\bm{g}(\bm{x}_\mathrm{init},\bm{u}(-T+1))}_p.\label{eq:ub_expanded}
\end{align}

Each $p$-norm in \eqref{eq:ub_expanded} can be bounded
from above by $\norm{\bm{g}-\bm{f}}_{\bar{\mathcal{B}}_{S,I},p}$.
This is because $\bm{g}$ and $\bm{f}$ are maps $\bar{\mathcal{B}}_S\times\bar{\mathcal{B}}_I\to\bar{\mathcal{B}}_S$ by the assumption and because they have the same arguments in each norm.
Therefore, for any $t\in\mathcal{T}_T$ one has 
\begin{align}
\eqref{eq:ub_expanded}&\le\norm{\bm{g}-\bm{f}}_{\bar{\mathcal{B}}_{S,I},p}\sum_{i=0}^{T+t-1}L^i\le\norm{\bm{g}-\bm{f}}_{\bar{\mathcal{B}}_{S,I},p}\sum_{i=0}^{T-1}L^i.
\end{align}
This leads to
\begin{align}
\eqref{eq:norm_diff_f_and_cov_ub}&\le\sup_{\bm{u}\in(\bar{\mathcal{B}}_I)^T}\sup_{t\in\mathcal{T}_T}\norm{\bm{g}-\bm{f}}_{\bar{\mathcal{B}}_{S,I},p}\sum_{i=0}^{T-1}L^i=\norm{\bm{g}-\bm{f}}_{\bar{\mathcal{B}}_{S,I},p}\sum_{i=0}^{T-1}L^i,
\end{align}
proving the lemma. 
\qed

\subsection{Proof of Lemma~\ref{lem:fin_cov_params}}\label{app:proof_lem_fin_cov_params}
We first provide definition of covering numbers, as well as two lemmas related to it. 
\begin{definition}[covering number]\label{def:cov_num}
Given a normed vector space $(\mathcal{V},\|\cdot\|)$ 
and its subset $\mathcal{S}\subset\mathcal{V}$,
the covering number $\mathcal{N}(\mathcal{S},\Gamma,\|\cdot\|)$
with $\Gamma>0$ is defined as the size (i.e., the number of elements)
of the smallest $(\|\cdot\|,\Gamma)$-covering of $\mathcal{S}$. 
\end{definition}

\begin{lemma}[a special case of {\citet[Lemma 14.12]{Anthony_and_Bartlett:1999:nnl}}]
  \label{lemma:cn_scale}
  For a normed vector space $(\mathcal{V},\|\cdot\|)$,
  a set $\mathcal{S}\subset\mathcal{V}$, and $a,\Gamma>0$, 
  one has $\mathcal{N}(a\mathcal{S},\Gamma,\|\cdot\|)=\mathcal{N}(\mathcal{S},\Gamma/a,\|\cdot\|)$.
\end{lemma}
\begin{proof}
  The lemma immediately follows from the fact that 
  for a $(\|\cdot\|,\Gamma)$-covering $\mathcal{C}$ of $a\mathcal{S}$, 
  $a^{-1}\mathcal{C}$ is a $(\|\cdot\|,\Gamma/a)$-covering of $\mathcal{S}$.
\end{proof}

\begin{lemma}
  \label{lemma:cn_intvl}
  One has $\mathcal{N}([-1,1],\Gamma,|\cdot|)\le 1/\Gamma+1$.
\end{lemma}
\begin{proof}
  For $L\in\mathbb{N}_+$,
$\{x_i:=-1+(2i-1)\Gamma,i\in\{1,\ldots,L\}\}$ is a $(|\cdot|,\Gamma)$-covering
of $[-1,1]$ once $L$ satisfies $-1+(2L-1)\Gamma\ge1-\Gamma$, or equivalently,
$L\ge1/\Gamma$.
\end{proof}

Recalling that $\mathcal{Q}_{\mathrm{a},M}=\mathcal{Q}_{\mathrm{d},M}=[-2\sqrt{M},2\sqrt{M}]$ and $\mathcal{Q}_{\mathrm{e},M}=[-M,M]$.
the above two lemmas prove the first, the fourth and the last inequalities
of Lemma~\ref{lem:fin_cov_params}.

In order to prove the second inequality of Lemma~\ref{lem:fin_cov_params},
we introduce the following equation and a lemma.
We observe
\begin{align}
  \label{eq:Gammab_nball}
  \mathcal{Q}_{\mathrm{b},M}
  &=\{\bm{b}\mid\norm{\bm{b}}_{\bar{\mathcal{B}}_S}\le\sqrt{M}\}
  \nonumber\\
  &=\{\bm{b}\mid\norm{\bm{b}}_q\le\sqrt{M}/S\}
  \nonumber\\
  &=\sqrt{M}S^{-1}\bar{\mathcal{B}}_{\mathrm{unit},\|\cdot\|_q},
\end{align}
where the second equality follows from the fact
$\norm{\bm{x}}_{\bar{\mathcal{B}}_S}=S\norm{\bm{x}}_q$ (Section~V in \cite{Barron:1993:nn_univ}).

\begin{lemma}[see, e.g., Example 5.8 in \cite{Wainwright:2019:hds}]
  \label{lemma:cn_nball}
  For a normed vector space $(\mathcal{V},\|\cdot\|_{\mathcal{V}})$
  with dimension $D$, and for any $\Gamma>0$,
  the covering number of the unit-norm ball in $\mathcal{V}$
  is bounded from above as 
  $\mathcal{N}(\bar{\mathcal{B}}_{\mathrm{unit},\|\cdot\|_{\mathcal{V}}},\Gamma,\|\cdot\|_{\mathcal{V}})\le(2/\Gamma+1)^D$.
\end{lemma}

One then has 
\begin{align}
 &\abs{\bar{\mathcal{Q}}_{\mathrm{b},M,\Gamma}}
 \\
 &=\mathcal{N}(\mathcal{Q}_{\mathrm{b},M},\Gamma/(2S\sqrt{M}\cdot20ND^{1/p}),\|\cdot\|_q)
 \\
  &=\mathcal{N}(\sqrt{M}S^{-1}\bar{\mathcal{B}}_{\mathrm{unit},\|\cdot\|_q},\Gamma/(2S\sqrt{M}\cdot20ND^{1/p}),\|\cdot\|_q)
  \quad(\because\eqref{eq:Gammab_nball})
  \\
 &=\mathcal{N}(\bar{\mathcal{B}}_{\mathrm{unit},\|\cdot\|_q},M^{-1/2}S\Gamma/(2S\sqrt{M}\cdot20ND^{1/p}),\|\cdot\|_q)\quad(\because\text{Lemma~\ref{lemma:cn_scale}})
 \\
 &=(4M\cdot20ND^{1/p}/\Gamma+1)^D,\quad(\because\text{Lemma~\ref{lemma:cn_nball}})
\end{align}
proving the second inequality of Lemma~\ref{lem:fin_cov_params}.

We can prove the third inequality of Lemma~\ref{lem:fin_cov_params} in the same manner as the second inequality.
\qed

\subsection{Proof of Lemma~\ref{lem:diff_approx_funcs_ub_wrt_params}}
Let $h_n(\bm{s},\bm{u}):=\rho(\bm{b}_n^\top\bm{s}+\bm{c}_n^\top\bm{u}+d_n)$ and
$h'_n(\bm{s},\bm{u}):=\rho(\bm{b}_n'^\top\bm{s}+\bm{c}_n'^\top\bm{u}+d'_n)$. 
Accordingly, one can write $f_{N,M}(\bm{s},\bm{u})=\sum_{n=1}^{4N}a_nh_n(\bm{s},\bm{u})+e$
and $f_{N,M}'(\bm{s},\bm{u})=\sum_{n=1}^{4N}a_n'h_n'(\bm{s},\bm{u})+e'$.
One then has 
\begin{align}
&\norm{f_{N,M}-f'_{N,M}}_{\bar{\mathcal{B}}_{S,I},\infty}
\\
&=\norm{\sum_{n=1}^{4N}a_nh_n-\sum_{n=1}^{4N}a_n'h_n'+e-e'}_{\bar{\mathcal{B}}_{S,I},\infty}\label{eq:diff_entire}
\\
&\le\norm{\sum_{n=1}^{4N}a_nh_n-\sum_{n=1}^{4N}a_n'h_n}_{\bar{\mathcal{B}}_{S,I},\infty}\nonumber\\
&+\norm{\sum_{n=1}^{4N}a_n'h_n-\sum_{n=1}^{4N}a_n'h_n'}_{\bar{\mathcal{B}}_{S,I},\infty}+\abs{e-e'}.\label{eq:diff_fs}
\end{align}
The first two terms on the right-hand side is further bounded from above.
The first term is bounded from above as 
\begin{align}
\norm{\sum_{n=1}^{4N}a_nh_n-\sum_{n=1}^{4N}a_n'h_n}_{\bar{\mathcal{B}}_{S,I},\infty}&\le\sum_{n=1}^{4N}\abs{a_n-a_n'}\norm{h_n}_{\bar{\mathcal{B}}_{S,I},\infty}.
\end{align}
We further bound $\norm{h_n}_{\bar{\mathcal{B}}_{S,I},\infty}$ from above, as
\begin{align}
 \norm{h_n}_{\bar{\mathcal{B}}_{S,I},\infty}&=\sup_{(\bm{s},\bm{u})\in\bar{\mathcal{B}}_{S,I}}\abs{\rho(\bm{b}_n^\top\bm{s}+\bm{c}_n^\top\bm{u}+d_n)}
 \\
 &\le\sup_{\bm{s}\in\bar{\mathcal{B}}_S}\abs{\bm{b}_n^\top\bm{s}}
 +\sup_{\bm{u}\in\bar{\mathcal{B}}_I}\abs{\bm{c}_n^\top\bm{u}}
 +\abs{d_n}
 \\
 &=\norm{\bm{b}_n}_{\bar{\mathcal{B}}_S}
 +\norm{\bm{c}_n}_{\bar{\mathcal{B}}_I}
 +\abs{d_n}\\
 &\le4\sqrt{M},
\end{align}
where the first inequality follows
from the property of the ReLU function $\rho$
that it is 1-Lipschitz, as well as the triangle inequality.

The second term in~\eqref{eq:diff_fs} is bounded from above as 
\begin{align}
&\norm{\sum_{n=1}^{4N}a_n'h_n-\sum_{n=1}^{4N}a_n'h_n'}_{\bar{\mathcal{B}}_{S,I},\infty}
\\
&\le2\sqrt{M}\sum_{n=1}^{4N}\norm{h_n-h_n'}_{\bar{\mathcal{B}}_{S,I},\infty}\label{eq:bf_apply_Lip}
\\
&\le2\sqrt{M}\sum_{n=1}^{4N}
\sup_{(\bm{s},\bm{u})\in\bar{\mathcal{B}}_{S,I}}  
 \abs{\left(\bm{b}_n-\bm{b}_n'\right)^\top\bm{s}
 +\left(\bm{c}_n-\bm{c}_n'\right)^\top\bm{u}
 +\left(d_n-d_n'\right)}\label{eq:af_apply_Lip}
 \\
&\le2\sqrt{M}\nonumber\\
&\cdot\sum_{n=1}^{4N}\sup_{(\bm{s},\bm{u})\in\bar{\mathcal{B}}_{S,I}}\left\{
\norm{\bm{b}_n-\bm{b}_n'}_q\norm{\bm{s}}_p
+\norm{\bm{c}_n-\bm{c}_n'}_q\norm{\bm{u}}_p
+\abs{d_n-d_n'}\right\}\label{eq:holder_applied_q_norm_appeared}
\\
&\le2\sqrt{M}\sum_{n=1}^{4N}(S\norm{\bm{b}_n-\bm{b}_n'}_q
+I\norm{\bm{c}_n-\bm{c}_n'}_q
+\abs{d_n-d_n'}).
\end{align}
In the above,
the second inequality is due to the fact that $\rho$ is 1-Lipschitz,
the third inequality is obtained via the triangle inequality and the H\"{o}lder inequality,
and the fourth inequality follows from the facts
$\sup_{\bm{s}\in\bar{\mathcal{B}}_S}\norm{\bm{s}}_p=S$
and $\sup_{\bm{u}\in\bar{\mathcal{B}}_I}\norm{\bm{u}}_p=I$.
Therefore, the desired upper bound is obtained.
\qed

\subsection{Proof of Lemma~\ref{lem:approx_rv_funcs_covering_wrt_map_norm}}
Let $\{(a_n,\bm{b}_n,\bm{c}_n,d_n)\}_{n=1}^{4N}$ and $e$ be the parameters of $f_{N,M}\in\mathcal{F}_{N,M}^1$.
Then, for $n=1,\ldots,4N$, there exist $a_n'\in\bar{\mathcal{Q}}_{a,M,\Gamma}$, $\bm{b}_n'\in\bar{\mathcal{Q}}_{b,M,\Gamma}$, $\bm{c}_n'\in\bar{\mathcal{Q}}_{c,M,\Gamma}$, $d_n'\in\bar{\mathcal{Q}}_{d,M,\Gamma}$ that cover $a_n$, $\bm{b}_n$, $\bm{c}_n$, $d_n$, respectively.
Also, there exists $e'\in\bar{\mathcal{Q}}_{e,M,\Gamma}$ that covers $e$.
The FNN with the parameters $\{(a_n',\bm{b}_n',\bm{c}_n',d_n')\}_{n=1}^{4N}$ and $e'$ is an element of $\bar{\mathcal{F}}_{N,M,\Gamma}^1$ by definition.
Let $\bar{f}_{N,M,\Gamma}\in\bar{\mathcal{F}}_{N,M,\Gamma}^1$ be such an FNN.
By \eqref{eq:g_diff_ub_for_cov}, one has 
\begin{align}
&\norm{f_{N,M}-\bar{f}_{N,M,\Gamma}}_{\bar{\mathcal{B}}_{S,I},\infty}
\le4\sum_{n=1}^{4N}\frac{\Gamma}{20ND^{1/p}}+\frac{\Gamma}{5D^{1/p}}
=\frac{\Gamma}{D^{1/p}}.
\end{align}
\qed

\subsection{Proof of Lemma~\ref{lem:approx_funcs_covering_wrt_map_norm}}\label{app:pr_lem_approx_funcs_covering_wrt_map_norm}
 The covering result holds trivially via application of Lemma~\ref{lem:approx_rv_funcs_covering_wrt_map_norm} in a component-wise manner.
$N_{N,M,\Gamma}$ is bounded as follows.
\begin{align}
N_{N,M,\Gamma}:&=|\bar{\mathcal{F}}_{N,M,\Gamma}^D|
=\abs{\bar{\mathcal{F}}_{N,M,\Gamma}^1}^D
\\
&=(\abs{\bar{\mathcal{Q}}_{a,M,\Gamma}}\cdot\abs{\bar{\mathcal{Q}}_{b,M,\Gamma}}\cdot\abs{\bar{\mathcal{Q}}_{c,M,\Gamma}}\cdot\abs{\bar{\mathcal{Q}}_{d,M,\Gamma}})^{4DN}\cdot\abs{\bar{\mathcal{Q}}_{e,M,\Gamma}}^D
\\
&\le(8M\cdot20ND^{1/p}/\Gamma+1)^{4D(D+E+2)N+D}.\label{eq:cov_num_aft_int}
\end{align}
\qed

\subsection{Proof of Lemma~\ref{lem:error_bound_fmf}}
First of all, from the definition of the supremum norm one has 
\begin{align}
 &\norm{U_{\bm{g}}-U_{c(\bm{g},\bm{x}_\mathrm{init},T)}^{W_{\mathrm{last},T}(I_{D\times D})}}_{\mathcal{T}_{-\infty},p}
 \\
&=\sup_{\bm{u}\in(\bar{\mathcal{B}}_I)^{-\infty}}\sup_{t\in\mathcal{T}_{-\infty}}\norm{U_{\bm{g}}(\bm{u})_t-V_{\bm{g},\bm{x}_\mathrm{init}} (\bm{u}_{t-T+1:t})_0}_p,
 \\
&=\sup_{\bm{u}\in(\bar{\mathcal{B}}_I)^{-\infty}}\sup_{t\in\mathcal{T}_{-\infty}}\norm{V_{\bm{g},\bm{x}_{t-T}}(\bm{u}_{t-T+1:t})_0-V_{\bm{g},\bm{x}_\mathrm{init}}(\bm{u}_{t-T+1:t})_0}_p
\end{align}
where $\bm{x}_{t-T}:=U_{\bm{g}}(\bm{u})_{t-T}\in\bar{\mathcal{B}}_S$ 
and where $\bm{u}_{t-T+1:t}:=(u(t-T+1),u(t-T+2),\ldots,u(t-1),u(t))\in(\bar{\mathcal{B}}_I)^T$.
Applying the uniform state contracting property (Definition~\ref{def:usc}), one can show that the term inside supremum is bounded from above by $\Delta_{\mathrm{c},T}$.
\qed

\section{Proofs of Propositions in Discussion}
\subsection{Proof of Proposition~\ref{pp:rate_err_and_node_cascade}}
The discussion just above Proposition~\ref{pp:rate_err_and_node_cascade} immediately shows that the statement (i) holds under the choice $\Gamma\propto N^{-1/2}$ and $T=\lceil\log_{L_\mathrm{sc}}(N^{-1/2})\rceil$.

We prove the statement (ii).
$T=O(N)$ holds trivially.
In the following, we will show that $N_{N,M,\Gamma}=O(N^{12D(D+E+2)N+3D})$,
which immediately proves (ii).

First of all, one has $N_{N,M,\Gamma}\le(160MD^{1/p}N/\Gamma+1)^{4D(D+E+2)N+D}$ by Lemma~\ref{lem:approx_funcs_covering_wrt_map_norm}.
Let $\Gamma=aN^{-1/2}$, $a\in(0,\infty)$.
Let $b:=160MD^{1/p}/a$.
Assume that $N$ is so large that the inequality $N^{3/2}\ge\max\{1/b,2b\}$ holds. 
By Lemma~\ref{lem:approx_funcs_covering_wrt_map_norm},
\begin{align}
 &\ln N_{N,M,\Gamma}
 \\
 &\le\ln\left(1+bN^{3/2}\right)^{4D(D+E+2)N+D}
 \\
&\le\ln\left(2bN^{3/2}\right)^{4D(D+E+2)N+D}
\\
&=(4D(D+E+2)N+D)\left(\ln2b+\ln N^{3/2}\right)
\\
&\le(12D(D+E+2)N+3D)\ln N.
 \end{align}
It shows that one has $N_{N,M,\Gamma}=O(N^{12D(D+E+2)N+3D})$, 
and therefore (ii) holds. 
\qed

\subsection{Proof of Proposition~\ref{pp:pdim_pl_rnns}}
We derive the upper bound of $\Pdim(\mathcal{H}_\mathrm{all})$ by combining a result in \cite{Yasumoto_and_Tanaka:2025:rc_complexities} and Theorem~11.7 in \cite{Mohri_et_al:2018:fml}.
Let $\mathcal{H}_\mathrm{r}:=\{W:\mathbb{R}^{N_{N,M,\Gamma}}\to\mathbb{R}\}$ be a set of linear functions.
By Corollary~17 in \cite{Yasumoto_and_Tanaka:2025:rc_complexities}, $\Pdim(\mathcal{H}_\mathrm{all})\le\Pdim(\mathcal{H}_\mathrm{r})$.
Since the pseudo-dimension of a vector space of real-valued functions
is equal to the dimension of
the vector space~\cite[Theorem~11.7]{Mohri_et_al:2018:fml}, one has 
$\Pdim(\mathcal{H}_\mathrm{r})=N_{N,M,\Gamma}$.
Therefore, $\Pdim(\mathcal{H}_\mathrm{all})\le N_{N,M,\Gamma}$.

We derive the upper bound of $\Pdim(\mathcal{H}_\mathrm{one})$.
Let $\Phi_\mathcal{H}(U_K,\bm{t}_K):=|\{(I_\pm(h(\bm{u}_1)-t_1),\ldots,I_\pm(h(\bm{u}_K)-t_K))\mid h\in\mathcal{H}\}|$.
Let $\bar{\Phi}_\mathcal{H}(K):=\max_{U_K\in\mathcal{U}^K,\bm{t}_K\in\mathbb{R}^K}\Phi_\mathcal{H}(U_K,\bm{t}_K)$.
Since $\mathcal{H}_\mathrm{one}$ is a finite set with $N_{N,M,\Gamma}$ elements, clearly $\bar{\Phi}_{\mathcal{H}_\mathrm{one}}(K)\le N_{N,M,\Gamma}$.
Therefore, 
\begin{align}
\Pdim(\mathcal{H}_\mathrm{one})&:=\max\{K\mid\bar{\Phi}_{\mathcal{H}_\mathrm{one}}(K)=2^K\}
\\
&\le\max\{K\mid N_{N,M,\Gamma}\ge2^K\}
\\
&\le\ceil{\log_2N_{N,M,\Gamma}}.
\end{align}
\qed

\subsection{Proof of Proposition~\ref{prop:esp_dense_in_ci}}
We prove that, for any $\bm{g}\in\mathcal{C}^i(\bar{\mathcal{B}}_{S,I},\mathbb{R}^D)$ and for any $\varepsilon>0$, there exist $\tau>0$ and a function $\bm{g}_\tau\in\mathcal{G}_\mathrm{nesp}$ such that $\norm{\bm{g}_\tau-\bm{g}}_{\bar{\mathcal{B}}_{S,I},p}<\varepsilon$.
If $\bm{g}\in\mathcal{G}_\mathrm{nesp}$, we can trivially take $\bm{g}_\tau=\bm{g}$.
Therefore, in the following, we prove this statement when $\bm{g}\in\mathcal{G}_\mathrm{esp}$.

We firstly show that $\bm{g}$ has a unique fixed point for any constant input.
Secondly, we modify $\bm{g}$ to construct a state map $\bm{g}_\tau\in\mathcal{G}_\mathrm{nesp}$ that has another fixed point whose distance from the original fixed point is $\tau$.
Then, we show that for any $\varepsilon>0$, we can choose sufficiently small $\tau>0$ such that $\norm{\bm{g}_\tau-\bm{g}}_{\bar{\mathcal{B}}_{S,I},p}<\varepsilon$.

We show that, for any $\bm{v}_0\in\bar{\mathcal{B}}_I$ and any $\bm{g}\in\mathcal{G}_\mathrm{esp}$, $\bm{g}(\cdot,\bm{v}_0)$ has a unique fixed point $\bm{x}_0\in\bar{\mathcal{B}}_S$ such that $\bm{g}(\bm{x}_0,\bm{v}_0)=\bm{x}_0$.
Firstly, since $\bm{g}(\cdot,\bm{v}_0)$ is a continuous map from a compact and convex subset of a Euclidean space to itself,
Brouwer's fixed-point theorem (e.g., Theorem~4.5 in \cite{Shapiro:2016:fixed_point_farrago}) ensures that it has at least one fixed point.
Secondly, we show uniqueness of the fixed point of $\bm{g}(\cdot,\bm{v}_0)$ by contradiction.
Suppose that there exist more than one fixed points $\bm{x}_1,\bm{x}_2,\ldots\in\bar{\mathcal{B}}_S$.
Then, for the left-infinte input sequence $\bm{u}:=(\ldots,\bm{v}_0,\bm{v}_0)$, $\bm{g}$ has different state sequences $(\ldots,\bm{x}_i,\bm{x}_i)$, $i=1,2,\ldots$ for $\bm{u}$.
It contradicts ESP.
Therefore, the fixed point is unique.

We modify $\bm{g}$ to construct $\bm{g}_\tau\in\mathcal{G}_\mathrm{nesp}$,
which has at least two fixed points
for a constant input $\bm{v}_0\in\bar{\mathcal{B}}_I$.
Let $\bm{x}_0\in\bar{\mathcal{B}}_S$ be the fixed point of $\bm{g}(\cdot,\bm{v}_0)$.
Take an arbitrary point $\bm{z}_0\in\bar{\mathcal{B}}_S$
with $\bm{z}_0\not=\bm{x}_0$, and let $d:=\|\bm{z}_0-\bm{x}_0\|_p>0$.
The convexity of $\bar{\mathcal{B}}_S$ ensures
existence of $\bm{z}\in\bar{\mathcal{B}}_S$
with $\|\bm{z}-\bm{x}_0\|_p=\tau$
for any $\tau\in[0,d]$.

For $r>0$, let $B_r:=\{(\bm{x},\bm{v})\in\mathbb{R}^D\times\mathbb{R}^E\mid
\|\bm{x}\|_p+\|\bm{v}\|_p\le r\}$ be the product-norm ball
of radius $r$ centered at the origin,
and let $w_\tau:\mathbb{R}^D\times\mathbb{R}^E\to[0,1]$ be
a bump function that takes the value 1 within $B_{\tau/2}$,
the value 0 outside $B_\tau$, and smoothly interpolating values
between the surfaces of these two balls.
Let $\bm{g}_\tau$ be constructed by ``grafting'' 
the constant function $\bm{z}$ onto $\bm{g}$ around $(\bm{z},\bm{v}_0)$, as
\begin{equation}
  \bm{g}_\tau(\bm{x},\bm{v})
  =(1-w_\tau(\bm{x}-\bm{z},\bm{v}-\bm{v}_0))\bm{g}(\bm{x},\bm{v})
  +w_\tau(\bm{x}-\bm{z},\bm{v}-\bm{v}_0)\bm{z}.
\end{equation}
We confirm that both $\bm{x}=\bm{x}_0$ and $\bm{z}$ are fixed points
of $\bm{g}_\tau(\cdot,\bm{v}_0)$.
Since $w_\tau(\bm{x}_0-\bm{z},\mathbf{0})=0$, one has
\begin{align}
  \bm{g}_\tau(\bm{x}_0,\bm{v}_0)
  &=(1-w_\tau(\bm{x}_0-\bm{z},\mathbf{0}))\bm{g}(\bm{x}_0,\bm{v}_0)
  +w_\tau(\bm{x}_0-\bm{z},\mathbf{0})\bm{z}
  \\
  &=\bm{g}(\bm{x}_0,\bm{v}_0)=\bm{x}_0.
\end{align}
One also has
\begin{equation}
  \bm{g}_\tau(\bm{z},\bm{v}_0)
  =(1-w_\tau(\mathbf{0},\mathbf{0}))\bm{g}(\bm{z},\bm{v}_0)
  +w_\tau(\mathbf{0},\mathbf{0})\bm{z}
  =\bm{z}.
\end{equation}
These imply that $\bm{g}_\tau\in\mathcal{G}_\mathrm{nesp}$.

We next show that $\bm{g}_\tau$ can be made arbitrarily close to
$\bm{g}$ by taking $\tau$ close enough to zero.
Since $\bm{g}_\tau(\bm{x},\bm{v})-\bm{g}(\bm{x},\bm{v})
=w_\tau(\bm{x}-\bm{z},\bm{v}-\bm{v}_0)(\bm{z}-\bm{g}(\bm{x},\bm{v}))$,
one can bound the norm $\norm{\bm{g}_\tau-\bm{g}}_{\bar{\mathcal{B}}_{S,I},p}$
from above as
\begin{align}
  \norm{\bm{g}_\tau-\bm{g}}_{\bar{\mathcal{B}}_{S,I},p}
  &=\sup_{(\bm{x},\bm{v})\in\bar{\mathcal{B}}_{S,I},
    \atop(\bm{x}-\bm{z},\bm{v}-\bm{v}_0)\in B_\tau}
  w_\tau(\bm{x}-\bm{z},\bm{v}-\bm{v}_0)\norm{\bm{z}-\bm{g}(\bm{x},\bm{v})}_p
  \nonumber\\
  &\le\norm{\bm{z}-\bm{x}_0}_p
  +\sup_{(\bm{x},\bm{v})\in\bar{\mathcal{B}}_{S,I},
    \atop(\bm{x}-\bm{z},\bm{v}-\bm{v}_0)\in B_\tau}
  \norm{\bm{g}(\bm{x},\bm{v})-\bm{x}_0}_p
  \nonumber\\
  &\le\tau
  +\sup_{(\bm{x},\bm{v})\in\bar{\mathcal{B}}_{S,I},
    \atop(\bm{x}-\bm{x}_0,\bm{v}-\bm{v}_0)\in B_{2\tau}}
  \norm{\bm{g}(\bm{x},\bm{v})-\bm{x}_0}_p,
\end{align}
where the first equality is due to the fact that
the support of $w_\tau$ is $B_\tau$,
where the next inequality is due to the triangle inequality,
and where the last inequality holds because for any $a,b>0$
$(\bm{x},\bm{v})\in B_a$ with $\norm{\bm{b}}_p=b$
implies $(\bm{x}+\bm{b},\bm{v})\in B_{a+b}$,
shown straightforwardly via applying the triangle inequality.

The continuity of $\bm{g}$ implies that, for any $\varepsilon>0$,
there exists $\delta>0$ such that
\begin{equation}
  \sup_{(\bm{x},\bm{v})\in\bar{\mathcal{B}}_{S,I},
    \atop(\bm{x}-\bm{x}_0,\bm{v}-\bm{v}_0)\in B_\delta}
  \norm{\bm{g}(\bm{x},\bm{v})-\bm{x}_0}_p<\varepsilon
\end{equation}
holds.
By letting $\tau\in(0,\min\{\delta/2,\varepsilon,d\}]$,
one has 
\begin{align}
  \norm{\bm{g}_\tau-\bm{g}}_{\bar{\mathcal{B}}_{S,I},p}
  &\le\tau
  +\sup_{(\bm{x},\bm{v})\in\bar{\mathcal{B}}_{S,I},
    \atop(\bm{x}-\bm{x}_0,\bm{v}-\bm{v}_0)\in B_{2\tau}}
  \norm{\bm{g}(\bm{x},\bm{v})-\bm{x}_0}_p
  \nonumber\\
  &\le\varepsilon
  +\sup_{(\bm{x},\bm{v})\in\bar{\mathcal{B}}_{S,I},
    \atop(\bm{x}-\bm{x}_0,\bm{v}-\bm{v}_0)\in B_\delta}
  \norm{\bm{g}(\bm{x},\bm{v})-\bm{x}_0}_p
  \nonumber\\
  &\le2\varepsilon.
\end{align}
Since $\varepsilon$ can be taken arbitrarily close to zero,
the above argument shows that any neighborhood of $\bm{g}\in\mathcal{G}_\mathrm{esp}$ contains a function in $\mathcal{G}_\mathrm{nesp}$,
completing the proof. 
\qed

\section{Glossary of mathematical symbols}\label{app:glossary}
\xentrystretch{0}
\begin{xtabular}{p{3.2cm}p{8.3cm}}
$\|\cdot\|_p$&$l_p$-norm (Section~\ref{ssec:notations})
\\
$\|\cdot\|_{\mathcal{A},p}$&uniform norm of a continuous function on $\mathcal{A}$ (Section~\ref{ssec:notations})
\\
$\norm{\cdot}_{\mathcal{T}_T,p}$&norm of finite-length filter (Definition~\ref{def:filter_norm} in Section~\ref{sssec:norm_filter})
\\
$1_P$&indicator function with a proposition $P$ (Section~\ref{ssec:notations})
\\
$a\mathcal{S}$&space made by scaling $\mathcal{S}$ by $a$ (Section~\ref{ssec:notations})
\\
$\mathcal{B}^{-\infty}$&Cartesian product of left-infinite copies of a set $\mathcal{B}$ (Section~\ref{ssec:usu_lf_overview})
\\
$\bar{\mathcal{B}}_{\mathrm{unit},\|\cdot\|}$&unit ball of a normed vector space (Section~\ref{ssec:notations})
\\
$\bar{\mathcal{B}}_I$&domain of inputs (Section~\ref{sssec:cls_dyn_sys})
\\
$\bar{\mathcal{B}}_S$&domain of states of target dynamical systems (Section~\ref{sssec:cls_dyn_sys})
\\
$\bar{\mathcal{B}}_{S,I}$&domain of target dynamical systems (Section~\ref{sssec:cls_dyn_sys})
\\
$\mathcal{C}$&class of continuous functions
\\
$c(\bm{g},\bm{x}_\mathrm{init},T)$&order-$T$ cascaded dynamical system of $\bm{g}$ (Definition~\ref{def:cascade} in Section~\ref{ssec:outline_proof_usu_inf})
\\
$D\in\mathbb{N}_+$&dimension of state of target dynamical systems (Section~\ref{sssec:dyn_sys})
\\
$D_\mathrm{out}\in\mathbb{N}_+$&dimension of output of target dynamical systems (Section~\ref{sssec:dyn_sys})
\\
$\mathrm{err}$&approximation error of filters (Definition~\ref{def:f_and_res_approx_err} in Section~\ref{sssec:approx_err})
\\
$E\in\mathbb{N}_+$&dimension of inputs (Section~\ref{sssec:dyn_sys})
\\
$\tilde{f}$&Fourier transform of a function $f$ (Section~\ref{ssec:notations})
\\
$\bm{f}_{N,M}\in\mathcal{F}_{N,M}^D$&approximating FNN of $\bm{g}\in\mathcal{G}$ (Lemma~\ref{lem:f_approx_ub} in Section~\ref{ssec:outline_proof_wu})
\\
$\bar{\bm{f}}_{N,M,\Gamma}\in\bar{\mathcal{F}}_{N,M,\Gamma}^D$&a covering FNN (Lemma~\ref{lem:approx_funcs_covering_wrt_map_norm} in Section~\ref{sssec:cov_fnn})
\\
$\bm{F}_{N,M,\Gamma}$&concatenated FNN (Definition~\ref{def:parallel_nns} in Section~\ref{sssec:pl_con_fnn})
\\
$\mathcal{F}$&family of FNN reservoir maps (Section~\ref{ssec:approx_res_sys})
\\
$\mathcal{F}_{N,M}^1$&family of real-valued bounded-parameter FNNs (Definition~\ref{def:cb_approx_fnn} in Section~\ref{ssec:outline_proof_wu})
\\
$\bar{\mathcal{F}}_{N,M,\Gamma}^1$&covering of real-valued bounded-parameter FNNs $\mathcal{F}_{N,M}^1$ (Lemma~\ref{lem:approx_rv_funcs_covering_wrt_map_norm} in Section~\ref{sssec:cov_rv_fnn})
\\
$\mathcal{F}_{N,M}^D$&family of bounded-parameter FNNs (Definition~\ref{def:cb_approx_fnn} in Section~\ref{ssec:outline_proof_wu})
\\
$\bar{\mathcal{F}}_{N,M,\Gamma}^D$&covering of  bounded-parameter FNNs $\mathcal{F}_{N,M}^D$ (Lemma~\ref{lem:approx_funcs_covering_wrt_map_norm} in Section~\ref{sssec:cov_fnn})
\\
$\mathcal{G}$&a set of target dynamical systems (Assumption~\ref{asm:ds_asm_fin} in Section~\ref{sssec:cls_dyn_sys})
\\
$\mathcal{G}_\mathrm{c}$&a set of contracting target dynamical systems (Definition~\ref{def:contracting_tar_ds} in Section~\ref{ssec:dyn_sys_inf})
\\
$\mathcal{G}_\mathrm{sc}$&a set of strictly contracting target dynamical systems (Example~\ref{ex:strictly_cont_tar_sys} in Section~\ref{ssec:dyn_sys_inf})
\\
$I_{D\times D}\in\mathbb{R}^{D\times D}$&identity matrix (Section~\ref{ssec:notations})
\\
$k$&readout-choosing rule (Section~\ref{ssec:unif_str_univ})
\\
$k_{N,M,\Gamma}$&readout-choosing rule for a concatenated RNN reservoir system (Section~\ref{sssec:usu_fin_conc})
\\
$k_{N,M,\Gamma,T}$&readout-choosing rule for a cascaded RNN reservoir system (Theorem~\ref{th:cas_pl_res_error} in Section~\ref{ssec:outline_proof_usu_inf})
\\
$L\in(0,\infty)$&Lipschitz constant of target dynamical systems $\mathcal{G}$ and contracting target dynamical systems $\mathcal{G}_\mathrm{c}$ (Assumption~\ref{asm:ds_asm_fin} in Section~\ref{sssec:cls_dyn_sys})
\\
$L_\mathrm{sc}\in(L,1)$&upper bound of Lipschitz constant of strictly contracting target dynamical systems $\mathcal{G}_\mathrm{sc}$ (Example~\ref{ex:strictly_cont_tar_sys} in Section~\ref{ssec:dyn_sys_inf})
\\
$M\in(0,\infty)$&constant of Barron class (Definition~\ref{def:M_Barron} in Section~\ref{sssec:cls_dyn_sys})
\\
$N\in\mathbb{N}_+$&number proportional to the number of hidden nodes of real-valued FNNs (Proposition~\ref{pp:prop_2_2} in Section~\ref{ssec:nn_approx_relu} or Proposition~\ref{pp:prop_2_2_sig} in Section~\ref{ssec:nn_approx_sig}).
\\
$N_\mathrm{res}\in\mathbb{N}_+$&dimension of reservoir states (Section~\ref{ssec:approx_res_sys})
\\
$N_{N,M,\Gamma}$&number of bounded-parameter FNNs in the covering $\bar{\mathcal{F}}_{N,M,\Gamma}^D$ (Lemma~\ref{lem:approx_funcs_covering_wrt_map_norm} in Section~\ref{sssec:cov_fnn})
\\
$\mathcal{N}(\mathcal{S},\Gamma,\|\cdot\|)$&covering number of a set $\mathcal{S}$ (Definition~\ref{def:cov_num} in Appendix~\ref{app:proof_lem_fin_cov_params})
\\
$O_{D_1\times D_2}\in\mathbb{R}^{D_1\times D_2}$&zero matrix (Section~\ref{ssec:notations})
\\
$p\in[1,\infty]$&degree of $l_p$ space (Section~\ref{ssec:notations})
\\
$P\in(0,S)$&scale of range of target dynamical systems (Assumption~\ref{asm:ds_asm_fin} in Section~\ref{sssec:cls_dyn_sys})
\\
$p(D,E,M,N,\Gamma)$&approximation error of target dynamical systems by covering FNNs (Lemma~\ref{lem:approx_dyn_sys_by_cov_fnn} in Section~\ref{sssec:approx_dyn_sys_by_cov_fnn})
\\
$q\in[1,\infty]$&H\"older conjugate of $p$ (Section~\ref{ssec:notations})
\\
$\mathcal{Q}_{\cdot,M}$&set of parameters of real-valued bounded-parameter FNNs $\mathcal{F}_{N,M}^1$ (Definition~\ref{def:cb_approx_fnn} in Section~\ref{ssec:result_wu})
\\
$\bar{\mathcal{Q}}_{\cdot,M,\Gamma}$&covering of $\mathcal{Q}_{\cdot,M}$ (Definition~\ref{def:cov_param_fnn} in Section~\ref{sssec:cov_param_rv_fnn})
\\
$\bm{r}$&reservoir map (Section~\ref{ssec:approx_res_sys})
\\
$\mathcal{R}$&family of reservoir maps (Definition~\ref{def:weak_univ} in Section~\ref{ssec:def_wu})
\\
$\bm{s}\in\mathbb{R}^{N_\mathrm{res}}$&state of reservoirs (Section~\ref{ssec:approx_res_sys})
\\
$\bm{s}_\mathrm{init}\in\mathbb{R}^{N_\mathrm{res}}$&initial state of reservoirs (Section~\ref{ssec:approx_res_sys})
\\
$\mathcal{T}_T$&set of time indices of length $T$ (Section~\ref{sssec:cls_dyn_sys})
\\
$\mathcal{T}_{-\infty}$&set of left-infinite time indices  (Section~\ref{ssec:usu_lf_overview})
\\
$U$&left-infinite filter (Section~\ref{ssec:def_usu_lf})
\\
$U_{\bm{g}}$&left-infinite filter of a dynamical system $\bm{g}$ (Section~\ref{ssec:def_usu_lf})
\\
$U_{\bm{g}}^{\bm{h}}$&left-infinite filter of a dynamical system $(\bm{g},\bm{h})$ (Section~\ref{ssec:def_usu_lf})
\\
$V$&finite-length filter (Section~\ref{sssec:filter})
\\
$V_{\bm{g},\bm{x}_\mathrm{init}}$&finite-length state filter (Definition~\ref{def:dyn_sys_filter} in Section~\ref{sssec:filter})
\\
$V_{\bm{g},\bm{x}_\mathrm{init}}^{\bm{h}}$&finite-length dynamical system filter (Definition~\ref{def:dyn_sys_filter} in Section~\ref{sssec:filter})
\\
$V_{\bm{r},\bm{s}_\mathrm{init}}$&finite-length reservoir state filter (Definition~\ref{def:res_sys_filter} in Section~\ref{sssec:filter})
\\
$V_{\bm{r},\bm{s}_\mathrm{init}}^W$&finite-length reservoir system filter (Definition~\ref{def:res_sys_filter} in Section~\ref{sssec:filter})
\\
$W$&linear readout of reservoirs
\\
$W_{\mathrm{last},T}(I_{D\times D})$&linear readout for cascaded dynamical systems (Definition~\ref{def:cascade} in Section~\ref{ssec:outline_proof_usu_inf})
\\
$\mathrm{werr}$&worst approximation error (Definition~\ref{def:approx_error} in Section~\ref{ssec:unif_str_univ})
\\
$\bm{x}_\mathrm{init}\in\mathbb{R}^D$&initial state of target dynamical systems (Section~\ref{sssec:dyn_sys})
\\
$\bm{x}_{N,M,\Gamma}\in\mathbb{R}^{N_{N,M,\Gamma}D}$&initial state of a concatenated RNN reservoir $\bm{F}_{N,M,\Gamma}$ (Theorem~\ref{th:concatenated_fnn_error} in Section~\ref{sssec:usu_fin_conc}) 
\\
$\mathcal{Z}_{M,\mathcal{B}}$&Barron class (Definition~\ref{def:M_Barron} in Section~\ref{sssec:cls_dyn_sys})
\\
$\Gamma\in(0,\infty)$&covering radius of bounded parameter FNNs $\mathcal{F}_{N,M}^D$ (Definition~\ref{def:cov_param_fnn} in Section~\ref{sssec:cov_param_rv_fnn} and Lemma~\ref{lem:approx_funcs_covering_wrt_map_norm} in Section~\ref{sssec:cov_fnn})
\\
$\delta(\Lambda)$&approximation error of a step function by a sigmoid function scaled by $\Lambda$ (Proposition~\ref{pp:prop_2_2_sig} in Section~\ref{ssec:nn_approx_sig})
\\
$\Delta_{\mathrm{c},t}$&null sequence of contracting target dynamical systems $\mathcal{G}_\mathrm{c}$ (Definition~\ref{def:contracting_tar_ds} in Section~\ref{ssec:dyn_sys_inf})
\\
$\kappa\in(0,\infty)$&universal constant (Proposition~\ref{pp:prop_2_2} in Section~\ref{ssec:nn_approx_relu})
\\
$\Lambda\in(0,\infty)$&scale of sigmoid function (Proposition~\ref{pp:prop_2_2_sig} in Section~\ref{ssec:nn_approx_sig})
\\
$\rho$&ReLU activation function (Section~\ref{ssec:approx_res_sys})
\\
$\sigma$&monotonically increasing continuous sigmoid function (Section~\ref{ssec:nn_approx_sig})\\
\end{xtabular}

\bibliographystyle{elsarticle-harv}
\bibliography{universality.bib}
\end{document}